\documentclass{article}

\usepackage[preprint]{neurips_2026}
\usepackage[utf8]{inputenc} 
\usepackage[T1]{fontenc}    
\usepackage[english]{babel}
\usepackage{microtype}      
\usepackage{xspace}
\usepackage{minitoc}

\usepackage{amsmath, amssymb, amsthm, amsfonts}
\usepackage{nicefrac}       
\usepackage{algorithm}
\usepackage{algpseudocode}
\usepackage{dsfont}

\usepackage[table]{xcolor}  
\usepackage{booktabs}       
\usepackage{multirow}       
\usepackage{array}          
\usepackage{tabularx}       
\usepackage{ltablex}
\usepackage{arydshln}       
\usepackage{makecell}
\usepackage{colortbl}       
\usepackage{booktabs}
\usepackage{diagbox}
\usepackage{makecell}
\usepackage[table]{xcolor}
\usepackage{arydshln} 

\newcolumntype{C}[1]{>{\centering\arraybackslash}m{#1}}

\usepackage{graphicx}
\usepackage{subcaption}     
\usepackage{adjustbox}
\usepackage{pifont}
\usepackage{fontawesome}
\usepackage{tikz}
\usepackage{wrapfig}

\newcommand{\lightbulbicon}{%
  \begin{tikzpicture}[baseline=-0.5ex]
    \draw[fill=white, draw=insightteal, thick] (0,0) circle (1.5ex);
    \node[scale=0.8, color=insightteal] at (0,0) {\faLightbulbO~};
  \end{tikzpicture}%
}
\newcommand{\boldcircle}[1]{%
    \tikz[baseline=(char.base)]{
        \node[draw=black, line width=0.2mm, shape=circle, inner sep=0.125mm, minimum size=0.5em] (char) {\scriptsize\textbf{#1}};
    }\xspace
}

\definecolor{codegreen}{rgb}{0,0.6,0}
\definecolor{codegray}{rgb}{0.5,0.5,0.5}
\definecolor{codepurple}{rgb}{0.58,0,0.82}
\definecolor{backcolour}{RGB}{245,248,250}
\definecolor{emph}{RGB}{166,88,53}
\definecolor{nightblue}{RGB}{9,49,105}
\definecolor{keywords}{RGB}{207,33,46}
\definecolor{lightpurple}{RGB}{130,81,223}

\definecolor{LightYellow}{rgb}{0.99, 0.99, 0.59}
\definecolor{LightRed}{rgb}{0.97, 0.51, 0.47}
\definecolor{LightBlue}{rgb}{0.99, 0.59, 0.99}
\definecolor{LightGreen}{rgb}{0.59, 0.99, 0.99}
\definecolor{pastelblue}{RGB}{70, 70, 70} 
\definecolor{pastelorange}{RGB}{201, 171, 102} 
\definecolor{pastelgreen}{RGB}{76, 124, 49} 

\definecolor{lighteconomist}{RGB}{252, 233, 237} 
\definecolor{economist}{RGB}{115,00,00} 
\definecolor{customgreen}{RGB}{116, 154, 114}
\definecolor{lightgreen}{RGB}{240, 246, 232}
\definecolor{ForestGreen}{RGB}{34, 139, 34}
\definecolor{insightteal}{RGB}{6, 119, 170}   
\definecolor{insightback}{RGB}{240, 248, 248}  
\definecolor{greydark}{RGB}{90, 90, 90} 
\definecolor{greylight}{RGB}{245, 245, 245}
\definecolor{lightgray}{gray}{0.75}
\definecolor{verylightgray}{gray}{0.95}

\definecolor{corporateblue}{RGB}{0, 80, 155} 
\definecolor{corporatelightblue}{RGB}{230, 240, 250}

\usepackage{listings}
\lstdefinestyle{mystyle}{
    backgroundcolor=\color{backcolour},   
    commentstyle=\color{codegreen},
    keywordstyle=\color{keywords},
    stringstyle=\color{nightblue},
    basicstyle=\ttfamily\footnotesize,
    breakatwhitespace=false,         
    breaklines=true,                 
    captionpos=b,                    
    keepspaces=true,                 
    showspaces=false,                
    showstringspaces=false,
    showtabs=false,                  
    tabsize=2,
    frame=shadowbox,
    emph={Explainer},
    emphstyle={\color{emph}},
    emph={[2]from_pretrained,compute_table},
    emphstyle={[2]\color{lightpurple}},
    linewidth=14.0cm
}

\usepackage{tcolorbox}
\tcbuselibrary{most, breakable, skins} 

\newtcolorbox{customblockquote}{
  colframe=insightteal, colback=insightback, boxrule=0pt,
  left=5pt, right=4pt, top=5pt, bottom=3pt, arc=0pt,
  breakable, enhanced jigsaw, frame hidden,
  before skip=1.2\baselineskip, after skip=0.7\baselineskip,
  overlay={
    \draw[insightteal, line width=2pt] ([yshift=1pt]frame.north west) -- (frame.south west);
    \node[inner sep=0pt] at ([xshift=0pt, yshift=-1.3pt]frame.north west) {\lightbulbicon};
  },
  fontupper=\fontfamily{lmr}\selectfont, boxsep=1pt,
}

\tcbset{
  mybox/.style={
    colback=black!5!white, colframe=cyan!20!white, 
    fonttitle=\bfseries, coltitle=black, boxrule=0.5mm, title={#1}
  }
}

\newtcolorbox{greycustomblock}{
  colframe=greydark, colback=greylight, boxrule=1pt,
  left=2.5pt, right=3pt, top=5pt, bottom=3pt, arc=0pt,
  breakable, enhanced jigsaw, frame hidden,
  before skip=0.2\baselineskip, after skip=0.2\baselineskip,
  overlay={
    \draw[greydark, line width=2pt] ([yshift=-1pt]frame.north west) -- ([yshift=1pt]frame.south west);
  },
  fontupper=\selectfont,
}

\newtcolorbox{bluecustombox}{
  colframe=corporateblue, colback=corporatelightblue, boxrule=1pt,
  left=2.5pt, right=3pt, top=5pt, bottom=3pt, arc=0pt,
  breakable, enhanced jigsaw, frame hidden,
  before skip=0.2\baselineskip, after skip=0.2\baselineskip,
  overlay={
    \draw[corporateblue, line width=2pt] ([yshift=-1pt]frame.north west) -- ([yshift=1pt]frame.south west);
  },
  fontupper=\selectfont,
}

\newcommand{\trainfigroot}{figures/}
\newcommand{\trainimg}[3]{%
    \includegraphics[width=0.265\textwidth]{\trainfigroot/#1/#2_#3_training.pdf}%
}
\newcommand{\trainlegendimg}{%
    \includegraphics[width=0.86\textwidth]{\trainfigroot/shared_training_legend.pdf}%
}

\newcommand{\modeltrainingfigure}[3]{%
\begin{figure*}[h!]
\small
    \centering
    \captionsetup{justification=centering}
    \setlength{\tabcolsep}{3pt}
    \renewcommand{\arraystretch}{1.05}
    \begin{tabular}{>{\centering\arraybackslash}m{0.15\textwidth} >{\centering\arraybackslash}m{0.26\textwidth} >{\centering\arraybackslash}m{0.26\textwidth} >{\centering\arraybackslash}m{0.26\textwidth}}
        \toprule
        & \textbf{(Learned) Reward} & \textbf{Disc. Accuracy} & \textbf{Correctness Accuracy} \\
        \midrule
        \textsc{GSM8K} &
        \trainimg{math}{#1}{reward} &
        \trainimg{math}{#1}{accuracy} &
        \trainimg{math}{#1}{correctness} \\
        \textsc{MedReason} &
        \trainimg{medicine}{#1}{reward} &
        \trainimg{medicine}{#1}{accuracy} &
        \trainimg{medicine}{#1}{correctness} \\
        \textsc{MMLU-Pro} &
        \trainimg{mmlu}{#1}{reward} &
        \trainimg{mmlu}{#1}{accuracy} &
        \trainimg{mmlu}{#1}{correctness} \\
        \bottomrule
    \end{tabular}
    \vspace{0.35em}

    \trainlegendimg
    \caption[Training dynamics for #2]{\textbf{Training dynamics for #2.} #3}
    \label{fig:appendix-training-#1}
\end{figure*}
}

\usepackage{hyperref}
\usepackage{url}

\author{%
  Claudio Fanconi\\
  University of Cambridge\\
  \texttt{caf83@cam.ac.uk} \\
  \And
  Nicolás Astorga\\
  University of Cambridge\\
  \texttt{nja46@cam.ac.uk} 
  \And
  Mihaela van der Schaar\\
  University of Cambridge\\
  \texttt{mv472@cam.ac.uk} 
}
\title{Learning Reasoning Rewards from Expert Demonstrations with Inverse Reinforcement Learning}

\begin{document}

\maketitle

\begin{abstract}
Teaching large language models (LLMs) to reason during post-training typically relies on reinforcement learning with explicit outcome- or process-based reward functions. However, in many real-world settings, obtaining or defining such reward functions is difficult, especially for complex tasks, making learning from expert demonstrations an attractive alternative. The dominant approach, supervised fine-tuning (SFT), trains models to imitate expert reasoning traces directly, but suffers from the general limitations of off-policy learning: performance can be fragile to inference-time deviations from states explicitly covered by the demonstrations. To address this, we propose \textbf{Reasoning Adversarial Inverse Reinforcement Learning (R-AIRL)}. Rather than imitating the expert’s reasoning, R-AIRL infers the underlying process-level reward from the expert Chain-of-Thoughts. Through experiments on GSM8K, MMLU-Pro and MedReason we show that the reasoning reward function learned with R-AIRL can be effectively used throughout the training and inference pipeline: (1) to provide a training signal for \textbf{post-training}, outperforming SFT in most of the considered settings, (2) for \textbf{inference-time reranking}, improving pass@1 by up to 17.4 points, and (3) for \textbf{process-level evaluation}, localising reasoning failures with up to 86.1\% accuracy. Overall, R-AIRL bridges imitation learning and reward-based optimisation, enabling the extraction of meaningful reasoning signals from expert thinking traces. \footnote{We provide the code for our experiments at 
\href{https://github.com/fanconic/expert_reasoning}{https://github.com/fanconic/expert\_reasoning}}
\end{abstract}

\section{Introduction}\label{introduction}
Reasoning post-training for large language models (LLMs) often relies on reinforcement learning (RL) with external outcome verifiers or step-level supervision~\citep{Guo2025, uesato_solving_2022, lightman_lets_2023}. In many practical settings, these signals are difficult to design or expensive to collect, whereas expert demonstrations (from humans or powerful LLMs) are comparatively easy to obtain. This makes supervised fine-tuning (SFT) on expert traces a particularly attractive approach in the demonstration-only regime~\citep{Guo2025}. However, SFT mainly trains models to simply imitate observed reasoning traces rather than to evaluate them or directly extrapolate from them. As a result, when generation deviates from demonstrated paths, the model lacks an explicit objective for comparing alternative reasoning steps or recovering from errors~\citep{setlur_scaling_2025}.

A natural alternative is to optimise reasoning via a \textit{learned} reward. Yet, manually specifying such a reward remains challenging: hand-crafted signals can be brittle, task-specific and difficult to scale. At the same time, expert demonstrations implicitly encode which trajectories are preferred~\citep{abbeel_apprenticeship_2004, ziebart_maximum_nodate}. This suggests learning a reward for reasoning directly from demonstrations, rather than prescribing one externally.

\begin{figure*}[h]
\centering
\includegraphics[width=\linewidth]{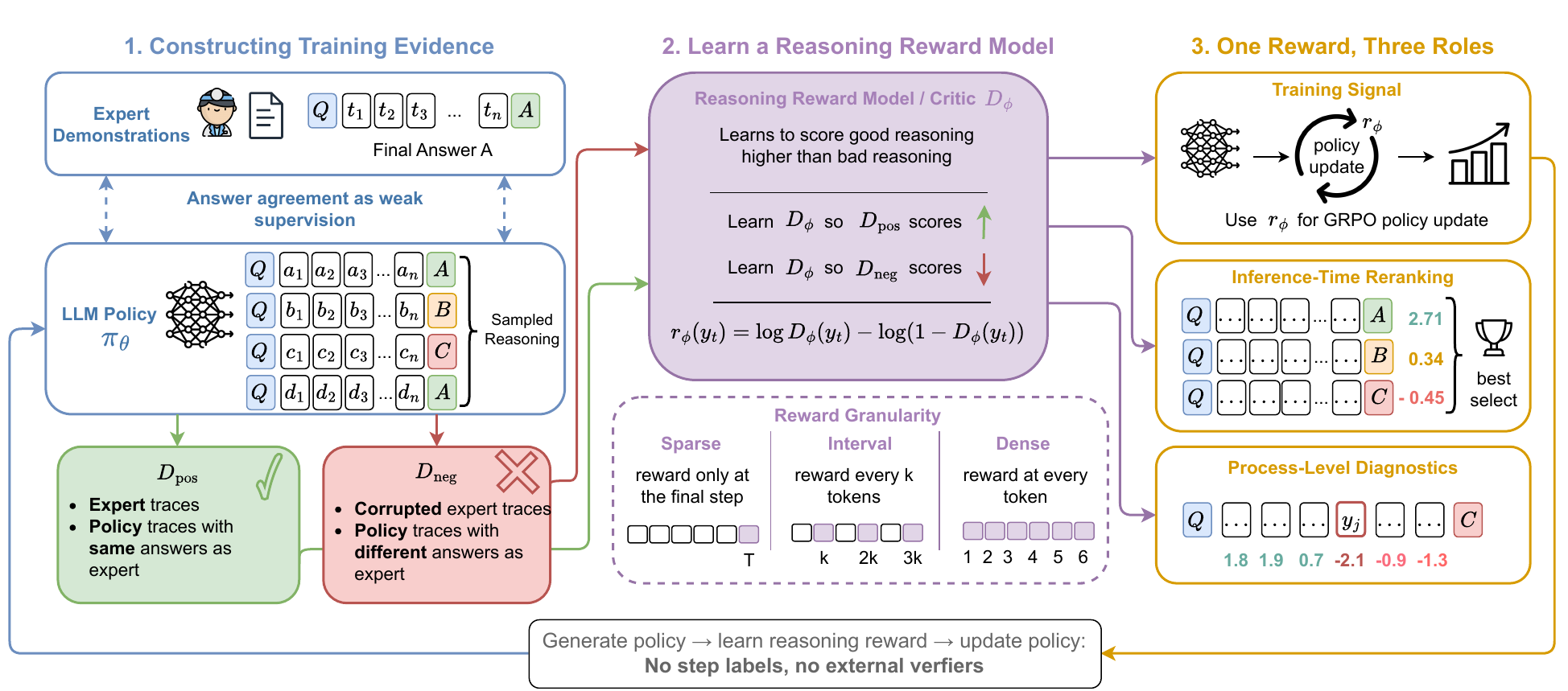}
\caption[Learning and using reasoning rewards from expert demonstrations]{\textbf{R-AIRL: Learning and using reasoning rewards from expert demonstrations.} (i) We construct training evidence from expert and policy traces, (ii) learn a critic-based reward at sparse/interval/dense granularities, and (iii) reuse that reward for GRPO training, inference-time reranking, and process-level error localisation, without step labels or external verifiers.}
\label{fig:figure_1}
\end{figure*}
\newpage
This raises a key question: can we obtain a reward over reasoning traces that is both learnable from demonstrations and useful in practice? We argue that such a reward, in order to be truly useful,  must satisfy three core requirements: it must provide a stable optimisation signal for policy learning, remain useful at inference time for selecting among candidate reasoning paths, and offer process-level feedback for diagnosing where reasoning fails.

\begin{customblockquote}
We formalise three desiderata for a learned reasoning reward:
\begin{enumerate}
    \item \textbf{\boldcircle{D1} Usable training signal.} A stable optimisation signal for on policy-generated trajectories.
    \item \textbf{\boldcircle{D2} Inference-time utility.} Effective tool for ranking the sampled traces under a fixed budget, without retraining.
    \item \textbf{\boldcircle{D3} Process-level feedback.} Step-level scoring function allowing to localise where reasoning first fails.
\end{enumerate}
\end{customblockquote}

To satisfy these desiderata, we formulate reasoning as an inverse reinforcement learning (IRL) problem~\citep{abbeel_apprenticeship_2004, ziebart_maximum_nodate}. Our proposed method, Reasoning Adversarial Inverse Reinforcement Learning (R-AIRL), adapts adversarial IRL~\citep{ho_generative_2016, fu_learning_2018} to the LLM reasoning setting: instead of teaching the model to imitate expert tokens, it learns a reasoning reward from demonstrations and then uses it to optimise the policy on its own sampled trajectories. In addition to providing a competitve training signal, the same reward can also bereused for inference-time reranking (to further boost performance) and fine-grained token/step diagnostics.

\textbf{Contributions.} We make three contributions:
\begin{enumerate}
    \setlength{\itemsep}{0.2em}
    \setlength{\topsep}{0.2em}
    \setlength{\parsep}{0pt}
    \setlength{\partopsep}{0pt}
    \item \textbf{Learning rewards for reasoning.} We adapt adversarial IRL to learn rewards over reasoning trajectories from expert demonstrations at multiple levels of granularity.
    \item \textbf{One reward, multiple roles.} We show that a single learned reward can be reused for policy optimisation, inference-time reranking, and error localisation.
    \item \textbf{Empirical validation across training, inference, and transfer.} Across \textsc{GSM8K}, \textsc{MedReason}, and \textsc{MMLU-Pro}, learned rewards often outperform SFT, improve Best-of-16 reranking by up to 17.4 points, transfer across tasks/backbones in most settings, and localise reasoning errors.
\end{enumerate}

Sections~\ref{sec:first_results},~\ref{sec:exp-d2}, and~\ref{sec:exp-d4} evaluate \boldcircle{D1}--\boldcircle{D3}, respectively; Figure~\ref{fig:figure_1} summarises the full pipeline.

\section{Problem Formalism}
\label{sec:formalism}

For a prompt \(x\sim\mathcal{Q}\), we model an output as a reasoning trajectory \(y=(y_1,\dots,y_T)\) with autoregressive likelihood \(p_\theta(y \mid x)=\prod_{t=1}^{T}\pi_\theta(y_t\mid x,y_{<t})\), where \(y_{<t}\) is the partial reasoning history.

\textbf{Latent process reward.}
The true reward for reasoning is unobserved, so we model a process-level reward \(r_\phi(y_t\mid x,y_{<t})\) and aggregate it as
\(R_\phi(y;m)=\frac{1}{T}\sum_{t=1}^{T} m_t\,r_\phi(y_t\mid x,y_{<t})\), with \(m_t\in\{0,1\}\). This unifies granularities: dense reward uses \(m_t=1\) for all \(t\), while outcome/sparse reward is the special case \(m_t=\mathbb{I}[t=T]\). The \(1/T\) factor makes scores comparable across different trajectory lengths.

\textbf{Inverse Reinforcement Learning (IRL).}
We treat expert demonstrations as samples from \(p_E(y\mid x)\), i.e., from an unknown expert policy that is approximately optimal under a latent reasoning reward. We formulate reward learning as learning a process-level reward that separates expert from policy trajectories, and then update the policy to maximise that learned reward:
\begin{equation}
\label{eq:irl_objective_general}
\max_{\phi}\;\min_{\theta}\;
\mathbb{E}_{y^E \sim p_E}\!\left[R_{\phi}(y^E;m)\right]
-
\mathbb{E}_{y \sim \pi_{\theta}}\!\left[R_{\phi}(y;m)\right].
\end{equation}
The first term learns \(r_\phi\) to prefer expert reasoning; the second drives policy optimisation under that reward. Because the expert term is independent of \(\theta\), minimising Eq.~\eqref{eq:irl_objective_general} w.r.t.\ \(\theta\) is equivalent to maximising expected learned reward on policy trajectories (\boldcircle{D1}). Since \(r_\phi\) is process-level, it yields token-level feedback (\boldcircle{D3}), and the same \(R_\phi\) can be reused for inference-time reranking (\boldcircle{D2}).

\section{Related Work}
\label{sec:related-work}

Using the notation from Section~\ref{sec:formalism}, we organise methods by supervision regime: verifier-based rewards, preference-based reward learning, and demonstration learning; we then compare whether they natively support \boldcircle{D1}--\boldcircle{D3}.

\textbf{Externally specified reward objectives.}
Outcome-supervised reasoning RL optimises the terminal-mask special case
\begin{equation}\label{eq:ver_outcome}
\max_{\theta}\;\mathbb{E}_{y\sim\pi_{\theta}}\!\left[R(y;m^{\mathrm{out}})\right],\qquad
m_t^{\mathrm{out}}=\mathbb{I}[t=T].
\end{equation}
This reward is usually derived from an external verifier~\citep{Guo2025, uesato_solving_2022, yu_dapo_2025, yuyue_vapo_2025}. Process-reward models (PRMs) instead optimise \(\max_{\theta}\mathbb{E}_{y\sim\pi_{\theta}}[R(y;m)]\) with denser masks \(m\) (e.g., full, interval, or step-wise). These approaches can be effective~\citep{uesato_solving_2022, lightman_lets_2023}, and search-based variants can improve exploration~\citep{zelikman_star_2022, yuan_scaling_2023, singh_beyond_2024, hosseini_v-star_2024, Silver2017MasteringTG}. Relative to our setting, they typically provide strong optimisation signals (\boldcircle{D1}) and can support reranking (\boldcircle{D2}) when reward access is available, while \boldcircle{D3} depends on externally specified process supervision. Crucially, however, in R-AIRL we assume no access to an external verifier and instead learn the reasoning from demonstrations.

\textbf{Preference-based reward learning.}
Preference-based RL learns a reward model \(R_{\phi}\), typically outcome-level, from pairwise preferences between a winning response \(y^{w}\) and a losing response \(y^{l}\) for the same prompt \(x\). With \((x,y^{w},y^{l})\sim\mathcal{D}_{\mathrm{pref}}\), a common objective is:
\begin{equation*}
\max_{\phi}\mathbb{E}\!\left[\log \sigma\!\left(R_{\phi}(y^{w};m^{\mathrm{out}})-R_{\phi}(y^{l};m^{\mathrm{out}})\right)\right],\quad \text{ followed by} \quad\max_{\theta}\mathbb{E}_{y\sim\pi_{\theta}}[R_{\phi}(y;m^{\mathrm{out}})].
\end{equation*}
This setting uses preference data rather than verifier-labelled rewards~\citep{christiano_deep_2023, rafailov_direct_2024}. This setting uses preference data rather than external verifier rewards~\citep{christiano_deep_2023, rafailov_direct_2024}. In our framing, preference-based rewards can support \boldcircle{D1} and \boldcircle{D2}, but are typically outcome-level and therefore less direct for \boldcircle{D3}. Again, a key distinction is that in R-AIRL, we assume no access to preference-labelled data and instead learn the reasoning from demonstrations.

\textbf{Demonstration-only learning.}
When supervision is provided as expert trajectories, a common training paradigm is supervised fine-tuning (SFT), i.e., knowledge distillation on teacher/expert traces~\citep{hinton_distilling_2015}, formalised through the following loss function:
\[
\max_{\theta}\;\mathbb{E}_{y^E\sim p_E}\!\left[\frac{1}{T}\sum_t \log \pi_{\theta}(y_t^E)\right].
\]
This often improves reasoning quality~\citep{Guo2025, kang_knowledge-augmented_2023, kujanpaa_efficient_2025, xu_kdrl_2025}, but it does not learn an explicit reward on policy-generated trajectories, so it is limited for \boldcircle{D1} and does not natively provide \boldcircle{D2} reranking or \boldcircle{D3} token-level localisation~\citep{setlur_scaling_2025}. Recent self-distillation work in continual learning remains distillation-based and complementary to reward learning~\citep{shenfeld2026selfdistillation}. Adversarial distillation variants~\citep{ye2025blackboxonpolicydistillationlarge, lee2025unified} infer trajectory-level discriminator rewards from teacher traces as an additional learning signal together with an external verifier:
\[
\footnotesize
\begin{aligned}
\max_{\phi}\min_{\theta}\quad &
\mathbb{E}_{y^E\sim p_E}\!\left[R_{\phi}(y^E;m^{\mathrm{out}})\right]
-\mathbb{E}_{y\sim\pi_{\theta}}\!\left[R_{\phi}(y;m^{\mathrm{out}})\right], \quad\quad
\max_{\theta}
\mathbb{E}_{y\sim\pi_{\theta}}\!\left[R_{\phi}(y;m^{\mathrm{out}})+R(y;m^{\mathrm{out}})\right].
\end{aligned}
\]
R-AIRL differs by learning process-level scoring \(R_{\phi}(y;m)\) from demonstrations and reusing one critic for \boldcircle{D1} training, \boldcircle{D2} reranking, and \boldcircle{D3} token-level localisation.

We summarise this comparison in Table~\ref{tab:related_work}.

\begin{table*}[h]
\scriptsize
\renewcommand{\arraystretch}{1.18}
\setlength{\tabcolsep}{4pt}
\centering
\resizebox{\textwidth}{!}{%
\begin{tabular}{l p{2.2cm} p{6.4cm} c c c c c}
\toprule
\textbf{Method Class} &
\makecell[l]{\textbf{Example}\\\textbf{Works}} &
\textbf{Optimisation Objective} &
\makecell[l]{\textbf{Supervision}\\\textbf{Data}} &
\makecell[l]{\textbf{No External}\\\textbf{Verifier}} &
\textbf{\boldcircle{D1}} &
\textbf{\boldcircle{D2}} &
\textbf{\boldcircle{D3}} \\
\midrule

\rowcolor{black!8}
\multicolumn{8}{l}{\textbf{Verifier-based supervision}} \\
\makecell[l]{Outcome-\\supervised RL}&
\makecell[l]{\citet{Guo2025}\\
\citet{yu_dapo_2025}\\
\citet{yuyue_vapo_2025}}
&
\makecell[l]{$\max_{\theta}\,\mathbb{E}_{y\sim\pi_{\theta}}\!\left[R(y;m^{\mathrm{out}})\right]$}
&
\makecell[c]{Outcome\\verifier}
& \textcolor{gray}{\ding{55}}
& \ding{51} & \ding{51} & \textcolor{gray}{\ding{55}} \\
\cdashline{1-8}[0.4pt/1.8pt]

\makecell[l]{Process-\\supervised RL} &
\makecell[l]{\citet{uesato_solving_2022}\\
\citet{lightman_lets_2023}}
&
\makecell[l]{$\max_{\theta}\,\mathbb{E}_{y\sim\pi_{\theta}}\!\left[R(y;m)\right]$}
&
\makecell[c]{Process\\verifier}
& \textcolor{gray}{\ding{55}}
& \ding{51} & \ding{51} & \ding{51} \\
\specialrule{0.9pt}{0.15em}{0.15em}

\rowcolor{black!8}
\multicolumn{8}{l}{\textbf{Preference supervision}} \\
\makecell[l]{Preference-\\supervised RL} &
\makecell[l]{\citet{christiano_deep_2023}\\
\citet{rafailov_direct_2024}}
&
\makecell[l]{$\max_{\phi}$\\
$\mathbb{E}_{(y^{w},y^{l})\sim\mathcal{D}_{\mathrm{pref}}}
\left[\log \sigma\!\left(R_{\phi}(y^{w};m^{\mathrm{out}})-R_{\phi}(y^{l};m^{\mathrm{out}})\right)\right]$\\
then \(\max_{\theta}\,\mathbb{E}_{y\sim\pi_{\theta}}\!\left[R_{\phi}(y;m^{\mathrm{out}})\right]\)}
&
\makecell[c]{Preference\\pairs}
& \textcolor{gray}{\ding{55}}
& \ding{51} & \ding{51} & \textcolor{gray}{\ding{55}} \\
\specialrule{0.9pt}{0.15em}{0.15em}

\rowcolor{black!8}
\multicolumn{8}{l}{\textbf{Demonstration supervision}} \\

SFT  &
\makecell[l]{\citet{hinton_distilling_2015}\\
\citet{Guo2025}\\
\citet{xu_kdrl_2025}}
&
\makecell[l]{$\max_{\theta}\,\mathbb{E}_{y^E\sim\mathcal{D}_E}\!\left[\frac{1}{T}\sum_t \log \pi_{\theta}(y_t^E)\right]$}
&
\makecell[c]{Demos}
& \ding{51}
& \textcolor{gray}{\ding{55}} & \textcolor{gray}{\ding{55}} & \textcolor{gray}{\ding{55}} \\
\cdashline{1-8}[0.4pt/1.8pt]

\makecell[l]{Adversarial\\distillation} &
\makecell[l]{
\citet{ye2025blackboxonpolicydistillationlarge}\\
\citet{lee2025unified}}
&
\makecell[l]{
$\max_{\phi}\min_{\theta}$\\
$\big(\mathbb{E}_{y^E\sim p_E}[R_{\phi}(y^E;m^{\mathrm{out}})]-\mathbb{E}_{y\sim\pi_{\theta}}[R_{\phi}(y;m^{\mathrm{out}})]\big)$\\
 \(\text{then }\max_{\theta}\mathbb{E}_{y\sim\pi_{\theta}}\!\left[R_{\phi}(y;m^{\mathrm{out}})+R(y;m^{\mathrm{out}})\right]\)}
&
\makecell[c]{Demos\\+ outcome\\verifier}
& \textcolor{gray}{(\ding{55})}
& \ding{51} & (\ding{51})  & \textcolor{gray}{\ding{55}} \\
\cdashline{1-8}[0.4pt/1.8pt]

\textbf{R-AIRL} &
\textbf{this work}
&
\makecell[l]{$\max_{\phi}\min_{\theta}$\\
$\big(\mathbb{E}_{y^E\sim p_E}[R_{\phi}(y^E;m)]-\mathbb{E}_{y\sim\pi_{\theta}}[R_{\phi}(y;m)]\big)$}
&
\makecell[c]{Demos}
& \ding{51}
& \ding{51} & \ding{51} & \ding{51} \\
\bottomrule
\end{tabular}}
\vspace{0.6em}
\caption[Related work comparison]{\textbf{Related work comparison.} Methods are grouped by supervision regime (verifier-based, preference-based, demonstrations), with objectives, supervision sources, and verifier dependence. \boldcircle{D1}: reward-based training signal on policy traces. \boldcircle{D2}: inference-time reranking with the same reward. \boldcircle{D3}: token-level diagnostics for error localisation. Parenthesised checkmarks indicate partial support.}
\label{tab:related_work}
\end{table*}

\section{Method}
\label{sec:method}
Our goal is to learn a reasoning reward from expert demonstrations, so the policy can be optimised on its own rollouts without external verifiers. We build on the adversarial IRL view of GAIL/AIRL~\citep{ho_generative_2016, fu_learning_2018}, but with a different supervision construction: instead of source-based expert-vs-policy labels, we form verifier-free positive/negative trace sets from answer agreement with paired demonstrations and corrupted expert traces, train a token-level critic, and convert its logits into rewards for policy updates.

\subsection{Training Evidence Construction}
A naive split (all expert positive, all policy negative) encourages style discrimination rather than reasoning-quality discrimination~\citep{ye2025blackboxonpolicydistillationlarge,lee2025unified}. We therefore use answer agreement with the paired expert trace as weak, verifier-free supervision. For each prompt, let \(y^E\) be the paired expert trace and let \(\mathcal{O}(y)\) denote the answer tokens from generation \(y\):
\[
\mathcal{D}_{\text{pos}} = \{ y \sim p_E \} \cup \{ y \sim \pi_\theta \mid \mathcal{O}(y) = \mathcal{O}(y^E) \}.
\]
Including answer-consistent policy traces in \(\mathcal{D}_{\text{pos}}\) puts policy text on both sides, reducing source shortcuts and pushing the critic toward reasoning quality. For negatives, we use answer-mismatched policy traces and augment them with corrupted expert traces:
\[
\mathcal{D}_{\text{neg}} = \{ y \sim \pi_\theta \mid \mathcal{O}(y) \neq \mathcal{O}(y^E) \} \cup \{ \mathcal{C}(y^E) \mid y^E \in \mathcal{D}^E \}.
\]
Corruptions preserve expert style while injecting reasoning errors, further discouraging provenance-based separation. This supervision is intentionally weak: answer agreement can miss latent errors or penalise partially correct traces, but it provides enough signal to learn useful rewards from demonstrations alone, without step-level labels or pairwise preferences.

\subsection{Reasoning Critic and Reward Definition}
\label{subsec:disc}
We train a token classifier \(D_\phi(y_t)\in(0,1)\) that assigns each token (and its previous context) a probability of belonging to the positive reasoning class (\(\mathcal{D}_{\text{pos}}\) vs.\ \(\mathcal{D}_{\text{neg}}\)). Following Section~\ref{sec:formalism}, a binary mask \(m_t\in\{0,1\}\) specifies on which tokens supervision is applied along the reasoning trajectory. We evaluate thee granularities: \textit{Sparse Outcome} (\(m_t=\mathbb{I}[t=T]\)), \textit{Interval} (\(m_t=1\) every \(k\) tokens), and \textit{Fully Dense} (\(m_t=1\) for all \(t\)).

Granularity controls credit assignment: coarser masks are typically lower-variance and more stable for optimisation (\boldcircle{D1}), while denser masks provide more detailed process feedback for localisation (\boldcircle{D3}) but can be harder to optimise robustly. This enables a direct sparse-to-dense comparison using a single supervision source (demonstrations only), unlike pipelines that rely on external process labels or verifiers~\citep{Guo2025, uesato_solving_2022, lightman_lets_2023}.

The critic objective is length-normalised binary cross-entropy:
\begin{equation}
\label{eq:discriminator_objective}
\begin{split}
\mathcal{L}_D(\phi) = \mathbb{E}_{y \sim \mathcal{D}_{\text{pos}}}\!\left[\frac{1}{T}\sum_{t=1}^T m_t \log D_\phi(y_t)\right]
-\mathbb{E}_{y \sim \mathcal{D}_{\text{neg}}}\!\left[\frac{1}{T}\sum_{t=1}^T m_t \log \big(1-D_\phi(y_t)\big)\right].
\end{split}
\end{equation}
After each critic update, we convert the probabilities into token rewards:
\begin{equation}
\label{eq:token_reward_def}
v_\phi(y_t)=\log D_\phi(y_t)-\log\!\big(1-D_\phi(y_t)\big).
\end{equation}

\subsection{Policy Optimisation with Learned Rewards}
\label{subsec:policy-learning}
Directly optimising Eq.~\eqref{eq:irl_objective_general} is a bilevel problem, which is computationally expensive for LLMs. We therefore alternate updates: critic steps for \(\phi\), then policy steps for \(\theta\) under the current reward.

We optimise the LLM policy \(\pi_\theta\) with Group Relative Policy Optimisation (GRPO)~\citep{shao_deepseekmath_2024}. At each policy iteration, we freeze \(\theta_{\text{old}}\), sample a group of \(G\) traces \(\{y^{(g)}\}_{g=1}^{G}\sim\pi_{\theta_{\text{old}}}\) for prompt \(x\), and update \(\theta\) while keeping \(\theta_{\text{old}}\) fixed.

\textbf{Reward Densification.}
The reward logit \(v_\phi(y_t)\) is only defined at supervised positions (\(m_t=1\)). For \textit{sparse} and \textit{interval}, we propagate each supervised score backwards to the preceding unsupervised tokens:
\begin{equation*}
t'=\min\{k\ge t\mid m_k=1\},\qquad r_\phi(y_t)=v_\phi(y_{t'}).
\end{equation*}
For \textit{dense}, \(r_\phi(y_t)=v_\phi(y_t)\). Intuitively, the latest supervised token for a segment has the richest context about the reasoning that led to it, so earlier tokens in that segment inherit its score.

\textbf{Advantage Calculation.}
Following \citet{cetin2025reinforcement} and \citet{cui_process_2025}'s approach for dense rewards in GRPO, we compute length-normalised sequence-level reward means
\[
\bar{r}^{(g)}=\frac{1}{T_g}\sum_{t=1}^{T_g} r_\phi(y_t^{(g)}),
\]
then calculate the advantage scores by normalising within each group:
\begin{equation}
\label{eq:advantage_grpo}
A^{(g)}=\frac{\bar{r}^{(g)}-\mu}{\sigma+\epsilon}.
\end{equation}
We assign the sequence-level advantage \(A^{(g)}\) to all tokens in generation \(g\), and optimise the PPO clip loss~\citep{schulman_proximal_2017} on policy-generated traces only (regardless of whether they fall in \(\mathcal{D}_{\text{pos}}\) or \(\mathcal{D}_{\text{neg}}\)).

Additional details on adversarial stability are provided in Appendix~\ref{app:method-details}; the full algorithm is given in Algorithm~\ref{alg:airl_reasoning}.

\section{Experiments}\label{sec:experiments}

We evaluate our learning of reasoning reward models via R-AIRL on \textsc{GSM8K}~\citep{cobbe_training_2021}, \textsc{MedReason}~\citep{wu2025medreasonelicitingfactualmedical}, and \textsc{MMLU-Pro}~\citep{wang2024mmlu}. We choose these benchmarks because they stress complementary reasoning regimes: \textsc{GSM8K} emphasises multi-step arithmetic computation, with demonstrations collected from humans.\textsc{MedReason} emphasises domain-specific clinical reasoning with plausible but incorrect alternatives, with demonstrations generated by LLMs and quality checked by medical doctors. \textsc{MMLU-Pro} emphasises broad, heterogeneous scientific reasoning, and the reasoning demonstrations are generated by LLMs. This diversity is important for assessing whether the learned reward remains useful across different reasoning styles, rather than a single-task format.

Our base policies are open-weight instruction-tuned variants not trained for reasoning: \texttt{Qwen2.5} (7B) \citep{bai_qwen_2023}, \texttt{Llama3} (8B) \citep{touvron_llama_2023}, and \texttt{Qwen3-4B}~\citep{qwen3}. We always instantiate a policy model as a text generator and the reasoning reward model as a token classifier.

Each experimental subsection directly evaluates one desideratum from Section~\ref{introduction}. \boldcircle{D1} (Section~\ref{sec:first_results}) tests training-signal quality in the demonstration-only (no external verifier) regime via final policy after R-AIRL training. \boldcircle{D2} (Section~\ref{sec:exp-d2}) tests inference-time utility via reranking gains under fixed budgets and transfer across tasks/backbones. \boldcircle{D3} (Section~\ref{sec:exp-d4}) tests process-level utility via localisation of the erroneous reasoning step. Implementation details appear in Appendix~\ref{app:implementation-details}.

\subsection{Using the Learned Reward as a Training Signal}\label{sec:first_results}

\begin{greycustomblock}
\textbf{RQ1:} \textit{Can R-AIRL improve reasoning in an LLM policy \(\pi_\theta\) (the generator) when \(\pi_\theta\) is trained only with rewards inferred from expert demonstrations?}
\end{greycustomblock}

\textbf{Experimental setup.} For each backbone, we train \(\pi_\theta\) with R-AIRL using \textit{sparse}, \textit{interval} (\(k=15\)), and \textit{dense} rewards. Following \citet{ye2025blackboxonpolicydistillationlarge,lee2025unified}, we use a 250-step discriminator warm-up, then jointly optimise policy and reward models per dataset. We report held-out pass@1 and compare to SFT under the same demonstration-only assumption (Table~\ref{tab:p1_results_main}).

\begin{table*}[h!]
\scriptsize
\renewcommand{\arraystretch}{1.16}
\setlength{\tabcolsep}{3.8pt}
\centering
\resizebox{\textwidth}{!}{%
\begin{tabular}{l l c c c c c c c c c}
\toprule
\textbf{Method} & \textbf{Granularity} & \multicolumn{3}{c}{\textbf{\textsc{GSM8K}}} & \multicolumn{3}{c}{\textbf{\textsc{MMLU-Pro}}} & \multicolumn{3}{c}{\textbf{\textsc{MedReason}}} \\
\cmidrule(lr){3-5}\cmidrule(lr){6-8}\cmidrule(lr){9-11}
& & \textbf{\tiny\texttt{Qwen2.5-7B}} & \textbf{\tiny\texttt{Llama3.1-8B}} & \textbf{\tiny\texttt{Qwen3-4B}} & \textbf{\tiny\texttt{Qwen2.5-7B}} & \textbf{\tiny\texttt{Llama3.1-8B}} & \textbf{\tiny\texttt{Qwen3-4B}} & \textbf{\tiny\texttt{Qwen2.5-7B}} & \textbf{\tiny\texttt{Llama3.1-8B}} & \textbf{\tiny\texttt{Qwen3-4B}} \\
\midrule
SFT &  & 70.1 {\scriptsize\color{gray}$\pm$ 1.6} & 66.6 {\scriptsize\color{gray}$\pm$ 2.1} & 76.6 {\scriptsize\color{gray}$\pm$ 1.7} & 48.1 {\scriptsize\color{gray}$\pm$ 1.9} & \textbf{47.2} {\scriptsize\color{gray}$\pm$ 1.9} & 53.9 {\scriptsize\color{gray}$\pm$ 2.0} & 65.4 {\scriptsize\color{gray}$\pm$ 2.3} & \textbf{79.5} {\scriptsize\color{gray}$\pm$ 2.0} & 67.4 {\scriptsize\color{gray}$\pm$ 2.3} \\
\cdashline{1-11}[0.5pt/1.8pt]
\multirow{3}{*}{R-AIRL} & \textit{Sparse} & \textbf{85.8} {\scriptsize\color{gray}$\pm$ 1.5} & \textbf{80.6} {\scriptsize\color{gray}$\pm$ 1.6} & \textbf{90.4} {\scriptsize\color{gray}$\pm$ 1.5} & \underline{48.5} {\scriptsize\color{gray}$\pm$ 2.2} & \underline{43.3} {\scriptsize\color{gray}$\pm$ 1.9} & \textbf{55.6} {\scriptsize\color{gray}$\pm$ 2.2} & \textbf{67.1} {\scriptsize\color{gray}$\pm$ 2.5} & \underline{77.6} {\scriptsize\color{gray}$\pm$ 2.0} & 68.9 {\scriptsize\color{gray}$\pm$ 2.6} \\
& \textit{Interval} & \underline{78.8} {\scriptsize\color{gray}$\pm$ 1.6} & \underline{67.6} {\scriptsize\color{gray}$\pm$ 1.9} & 87.8 {\scriptsize\color{gray}$\pm$ 1.6} & \textbf{50.6} {\scriptsize\color{gray}$\pm$ 1.9} & 36.6 {\scriptsize\color{gray}$\pm$ 1.8} & 53.5 {\scriptsize\color{gray}$\pm$ 2.4} & 66.3 {\scriptsize\color{gray}$\pm$ 2.5} & 76.3 {\scriptsize\color{gray}$\pm$ 2.1} & \underline{69.1} {\scriptsize\color{gray}$\pm$ 2.6} \\
& \textit{Dense} & 38.4 {\scriptsize\color{gray}$\pm$ 2.0} & 64.6 {\scriptsize\color{gray}$\pm$ 2.1} & \underline{89.6} {\scriptsize\color{gray}$\pm$ 1.5} & 43.8 {\scriptsize\color{gray}$\pm$ 2.0} & 37.9 {\scriptsize\color{gray}$\pm$ 1.8} & \underline{55.1} {\scriptsize\color{gray}$\pm$ 2.3} & \underline{66.5} {\scriptsize\color{gray}$\pm$ 2.6} & 75.1 {\scriptsize\color{gray}$\pm$ 2.1} & \textbf{69.2} {\scriptsize\color{gray}$\pm$ 2.5} \\
\bottomrule
\end{tabular}}
\vspace{0.6em}
\caption{\textbf{Held-out pass@1 in the demonstration-only setting.} \textbf{Bold}/\underline{underlined} mark best/second-best across SFT and R-AIRL variants. Values are mean \(\pm\) half-width of the bootstrapped 95\% confidence interval.}
\label{tab:p1_results_main}
\end{table*}

\textbf{Training-signal quality.} Table~\ref{tab:p1_results_main} shows learned rewards are competitive but not uniformly better than SFT: selecting the best granularity per backbone--dataset pair, R-AIRL improves over SFT in 7/9 settings. Gains are strongest on \textsc{GSM8K} (3/3 backbones) and positive for 2/3 backbones on \textsc{MedReason}; the main underperforming cases are \texttt{Llama3.1-8B} on \textsc{MMLU-Pro} and \textsc{MedReason}. Granularity controls stability: \textit{sparse} is best in 7/9 pairs, \textit{interval} in 1/9, and \textit{dense} in 1/9. RQ1 targets only \boldcircle{D1}; even where SFT has higher pass@1, it cannot provide the learned reward needed for \boldcircle{D2} reranking and \boldcircle{D3} process localisation (Sections~\ref{sec:exp-d2} and~\ref{sec:exp-d4}).

\begin{figure}[h!]
      \centering
      \begin{subfigure}{0.315\textwidth}
          \centering
          \includegraphics[width=\linewidth]{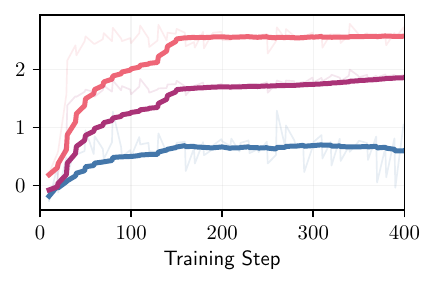}
          \caption{(Learned) Rewards}
          \label{fig:llama8b-rewards-training}
      \end{subfigure}
      \hfill
      \begin{subfigure}{0.315\textwidth}
          \centering
          \includegraphics[width=\linewidth]{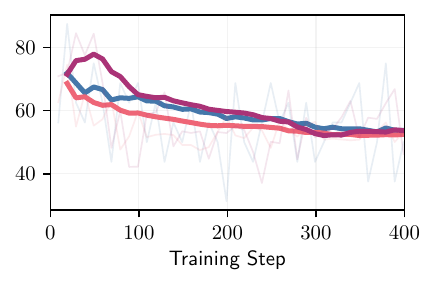}
          \caption{Discriminator Accuracy}
          \label{fig:llama8b-acc-training}
      \end{subfigure}
      \hfill
      \begin{subfigure}{0.315\textwidth}
          \centering
          \includegraphics[width=\linewidth]{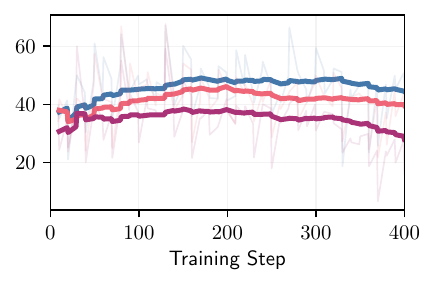}
          \caption{Task Correctness Accuracy}
          \label{fig:llama8b-correctness-training}
      \end{subfigure}

      \includegraphics[trim={0 0 0 6}, clip, width=0.9\textwidth]{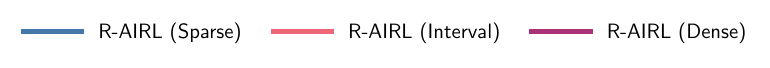}
        \vspace{-1em}
      \caption[Training behaviour of reward, reward-model correctness, and task correctness]{
    \textbf{Training dynamics for \texttt{Qwen2.5-7B} on \textsc{MMLU-Pro}.}
    Left: learned reward; middle: discriminator accuracy; right: held-out pass@1 (diagnostic only; not used for training updates). Reward improvement without pass@1 gains indicates reward--task decoupling.}\label{fig:train-curves-8b}

  \end{figure}

\textbf{Training dynamics.} Figure~\ref{fig:train-curves-8b} (Qwen2.5-7B on \textsc{MMLU-Pro}) should be read jointly: learned reward (left), discriminator accuracy (middle), and held-out pass@1 (right). In R-AIRL, discriminator accuracy approaching \(50\%\) is meaningful only if pass@1 rises; rising reward (\ref{fig:llama8b-rewards-training}) or falling discriminator accuracy (\ref{fig:llama8b-acc-training}) without sustained pass@1 gains (\ref{fig:llama8b-correctness-training}) indicates reward--task decoupling, more common under denser supervision. We attribute \textit{dense} instability to token-level supervision, which amplifies critic misclassifications and yields noisier policy rewards; \textit{sparse} rewards are less sensitive and thus more stable.

For completeness, Appendix Table~\ref{tab:p1_results_supervision_split} reports a comparison against outcome rewards if external verifiers were available. Appendix Figures~\ref{fig:appendix-training-qwen7b}--\ref{fig:appendix-training-qwen4b} show full training curves. Moreover, Appendix~\ref{app:additional-experiments} presents ablations on expert corruptions, group size, clipping, and policy-vs-expert-only separation, as well as on the AIME 2024/2025 datasets.

\begin{customblockquote}
\textbf{Takeaway.} RQ1 is answered positively: in the demonstration-only setting, learned rewards provide a viable training signal, with a granularity trade-off between optimisation stability (sparser rewards) and richer supervision (denser rewards).
\end{customblockquote}

\subsection{Inference-time Assistance via Reward-guided Reranking}
\label{sec:exp-d2}
\vspace{-1em}
\begin{greycustomblock}
\textbf{RQ2:} \textit{Can the learned reasoning reward improve inference by reranking multiple sampled traces under a fixed test-time budget?}
\end{greycustomblock}

\textbf{Experimental setup.} We evaluate inference-time reranking with the learned R-AIRL reward: for each prompt, sample \(N=16\) traces, rank by mean reward, and return the top candidate (Best-of-16). We compare against pass@1 with random selection from the same sample set. Main results are in Table~\ref{tab:reranking_main}, and calibration (AUROC/ECE) is in Appendix Table~\ref{tab:calibration}.

\begin{table*}[h!]
\scriptsize
\renewcommand{\arraystretch}{1.16}
\setlength{\tabcolsep}{3.8pt}
\centering
\resizebox{\textwidth}{!}{%
\begin{tabular}{l l c c c c c c c c c}
\toprule
\textbf{Method} & \textbf{Granularity} & \multicolumn{3}{c}{\textbf{\textsc{GSM8K}}} & \multicolumn{3}{c}{\textbf{\textsc{MMLU-Pro}}} & \multicolumn{3}{c}{\textbf{\textsc{MedReason}}} \\
\cmidrule(lr){3-5} \cmidrule(lr){6-8} \cmidrule(lr){9-11}
& & \textbf{Random} & \textbf{Reward} & $\mathbf{\Delta}$ & \textbf{Random} & \textbf{Reward} & $\mathbf{\Delta}$ & \textbf{Random} & \textbf{Reward} & $\mathbf{\Delta}$ \\
\midrule
\multirow{3}{*}{\texttt{Qwen2.5-7B}} & \textit{Sparse} & 85.8 {\scriptsize\color{gray}$\pm$ 1.5} & 88.8 {\scriptsize\color{gray}$\pm$ 1.7} & \textbf{\textcolor{insightteal}{($\uparrow$ +3.0)}} & 48.5 {\scriptsize\color{gray}$\pm$ 2.2} & 51.5 {\scriptsize\color{gray}$\pm$ 2.5} & \textbf{\textcolor{insightteal}{($\uparrow$ +3.0)}} & 67.1 {\scriptsize\color{gray}$\pm$ 2.5} & 70.1 {\scriptsize\color{gray}$\pm$ 2.9} & \textbf{\textcolor{insightteal}{($\uparrow$ +3.0)}} \\
 & \textit{Interval} & 78.8 {\scriptsize\color{gray}$\pm$ 1.6} & 82.5 {\scriptsize\color{gray}$\pm$ 2.0} & \textbf{\textcolor{insightteal}{($\uparrow$ +3.7)}} & 50.6 {\scriptsize\color{gray}$\pm$ 1.9} & 53.1 {\scriptsize\color{gray}$\pm$ 2.5} & \textbf{\textcolor{insightteal}{($\uparrow$ +2.5)}} & 66.3 {\scriptsize\color{gray}$\pm$ 2.5} & 67.2 {\scriptsize\color{gray}$\pm$ 2.9} & \textbf{\textcolor{insightteal}{($\uparrow$ +0.9)}} \\
 & \textit{Dense} & 38.4 {\scriptsize\color{gray}$\pm$ 2.0} & 55.7 {\scriptsize\color{gray}$\pm$ 2.6} & \textbf{\textcolor{insightteal}{($\uparrow$ +17.4)}} & 43.8 {\scriptsize\color{gray}$\pm$ 2.0} & 46.0 {\scriptsize\color{gray}$\pm$ 2.5} & \textbf{\textcolor{insightteal}{($\uparrow$ +2.2)}} & 66.5 {\scriptsize\color{gray}$\pm$ 2.6} & 66.6 {\scriptsize\color{gray}$\pm$ 2.9} & \textbf{\textcolor{insightteal}{($\uparrow$ +0.1)}} \\
\cdashline{1-11}[0.5pt/1.8pt]
\multirow{3}{*}{\texttt{Llama3.1-8B}} & \textit{Sparse} & 80.6 {\scriptsize\color{gray}$\pm$ 1.6} & 82.1 {\scriptsize\color{gray}$\pm$ 2.1} & \textbf{\textcolor{insightteal}{($\uparrow$ +1.5)}} & 43.3 {\scriptsize\color{gray}$\pm$ 1.9} & 42.3 {\scriptsize\color{gray}$\pm$ 2.4} & \textcolor{purple}{($\downarrow$ -1.0)} & 77.6 {\scriptsize\color{gray}$\pm$ 2.0} & 79.9 {\scriptsize\color{gray}$\pm$ 2.4} & \textbf{\textcolor{insightteal}{($\uparrow$ +2.3)}} \\
 & \textit{Interval} & 67.6 {\scriptsize\color{gray}$\pm$ 1.9} & 78.2 {\scriptsize\color{gray}$\pm$ 2.2} & \textbf{\textcolor{insightteal}{($\uparrow$ +10.7)}} & 36.6 {\scriptsize\color{gray}$\pm$ 1.8} & 38.1 {\scriptsize\color{gray}$\pm$ 2.4} & \textbf{\textcolor{insightteal}{($\uparrow$ +1.6)}} & 76.3 {\scriptsize\color{gray}$\pm$ 2.1} & 81.4 {\scriptsize\color{gray}$\pm$ 2.5} & \textbf{\textcolor{insightteal}{($\uparrow$ +5.1)}} \\
 & \textit{Dense} & 64.6 {\scriptsize\color{gray}$\pm$ 2.1} & 71.1 {\scriptsize\color{gray}$\pm$ 2.4} & \textbf{\textcolor{insightteal}{($\uparrow$ +6.5)}} & 37.9 {\scriptsize\color{gray}$\pm$ 1.8} & 39.7 {\scriptsize\color{gray}$\pm$ 2.5} & \textbf{\textcolor{insightteal}{($\uparrow$ +1.9)}} & 75.1 {\scriptsize\color{gray}$\pm$ 2.1} & 80.9 {\scriptsize\color{gray}$\pm$ 2.4} & \textbf{\textcolor{insightteal}{($\uparrow$ +5.8)}} \\
\cdashline{1-11}[0.5pt/1.8pt]
\multirow{3}{*}{\texttt{Qwen3-4B}} & \textit{Sparse} & 90.4 {\scriptsize\color{gray}$\pm$ 1.5} & 93.0 {\scriptsize\color{gray}$\pm$ 1.4} & \textbf{\textcolor{insightteal}{($\uparrow$ +2.5)}} & 55.6 {\scriptsize\color{gray}$\pm$ 2.2} & 62.1 {\scriptsize\color{gray}$\pm$ 2.4} & \textbf{\textcolor{insightteal}{($\uparrow$ +6.4)}} & 68.9 {\scriptsize\color{gray}$\pm$ 2.6} & 69.9 {\scriptsize\color{gray}$\pm$ 2.9} & \textbf{\textcolor{insightteal}{($\uparrow$ +1.0)}} \\
 & \textit{Interval} & 87.8 {\scriptsize\color{gray}$\pm$ 1.6} & 91.2 {\scriptsize\color{gray}$\pm$ 1.5} & \textbf{\textcolor{insightteal}{($\uparrow$ +3.5)}} & 53.5 {\scriptsize\color{gray}$\pm$ 2.4} & 62.2 {\scriptsize\color{gray}$\pm$ 2.4} & \textbf{\textcolor{insightteal}{($\uparrow$ +8.7)}} & 69.1 {\scriptsize\color{gray}$\pm$ 2.6} & 71.4 {\scriptsize\color{gray}$\pm$ 2.9} & \textbf{\textcolor{insightteal}{($\uparrow$ +2.3)}} \\
 & \textit{Dense} & 89.6 {\scriptsize\color{gray}$\pm$ 1.5} & 91.4 {\scriptsize\color{gray}$\pm$ 1.5} & \textbf{\textcolor{insightteal}{($\uparrow$ +1.8)}} & 55.1 {\scriptsize\color{gray}$\pm$ 2.3} & 59.2 {\scriptsize\color{gray}$\pm$ 2.4} & \textbf{\textcolor{insightteal}{($\uparrow$ +4.1)}} & 69.2 {\scriptsize\color{gray}$\pm$ 2.5} & 69.9 {\scriptsize\color{gray}$\pm$ 2.8} & \textbf{\textcolor{insightteal}{($\uparrow$ +0.7)}} \\
\bottomrule
\end{tabular}}
\vspace{0.6em}
\caption{\textbf{Best-of-16 reranking performance (\%).} Reward-guided reranking is compared against random selection from the same sample set. $\Delta$ reports percentage-point change.}
\label{tab:reranking_main}
\end{table*}

\textbf{Reranking performance.} In Table~\ref{tab:reranking_main}, reward-guided reranking improves pass@1 in 26/27 settings at \(N=16\). Largest gains include +17.4 pp (\texttt{Qwen2.5-7B}+\textit{dense}, \textsc{GSM8K}), +10.7 pp (\texttt{Llama3.1-8B}+\textit{interval}, \textsc{GSM8K}), and +8.7 pp (\texttt{Qwen3-4B}+\textit{interval}, \textsc{MMLU-Pro}). On \textsc{MedReason}, gains are smaller but positive in all 9 settings (+0.1 to +5.8 pp). The only drop is \texttt{Llama3.1-8B}+\textit{sparse} on \textsc{MMLU-Pro} (\(-1.0\) pp).

\textbf{Transfer across tasks and backbones.} Using \textit{interval} reward models trained on different datasets/backbones to rerank natural \texttt{Qwen2.5-7B} SFT samples, Table~\ref{tab:transfer} shows uniformly positive transfer: all 18 source--target pairs improve, including off-diagonal gains up to +12.1 pp (\texttt{Llama3.1-8B}) and +13.0 pp (\texttt{Qwen3-4B}).

\begin{table*}[h!]
\scriptsize
\renewcommand{\arraystretch}{1.16}
\setlength{\tabcolsep}{3.8pt}
\centering
\begin{tabular}{l c c c c c c}
\toprule
\textbf{Policy Dataset} & \multicolumn{3}{c}{\textbf{\texttt{Llama3.1-8B}}} & \multicolumn{3}{c}{\textbf{\texttt{Qwen3-4B}}} \\
\cmidrule(lr){2-4} \cmidrule(lr){5-7}
 & \textbf{\textsc{GSM8K}} & \textbf{\textsc{MMLU-Pro}} & \textbf{\textsc{MedReason}} & \textbf{\textsc{GSM8K}} & \textbf{\textsc{MMLU-Pro}} & \textbf{\textsc{MedReason}} \\
\midrule
\textsc{GSM8K} & \textit{\textcolor{insightteal}{$\uparrow$ +12.1}} & \textcolor{insightteal}{$\uparrow$ +7.7} & \textcolor{insightteal}{$\uparrow$ +12.1} & \textit{\textcolor{insightteal}{$\uparrow$ +13.0}} & \textcolor{insightteal}{$\uparrow$ +8.8} & \textcolor{insightteal}{$\uparrow$ +13.0} \\
\textsc{MMLU-Pro} & \textcolor{insightteal}{$\uparrow$ +4.9} & \textit{\textcolor{insightteal}{$\uparrow$ +4.4}} & \textcolor{insightteal}{$\uparrow$ +4.4} & \textcolor{insightteal}{$\uparrow$ +9.3} & \textit{\textcolor{insightteal}{$\uparrow$ +1.9}} & \textcolor{insightteal}{$\uparrow$ +1.9} \\
\textsc{MedReason} & \textcolor{insightteal}{$\uparrow$ +3.0} & \textcolor{insightteal}{$\uparrow$ +10.1} & \textit{\textcolor{insightteal}{$\uparrow$ +11.5}} & \textcolor{insightteal}{$\uparrow$ +5.5} & \textcolor{insightteal}{$\uparrow$ +5.6} & \textit{\textcolor{insightteal}{$\uparrow$ +5.8}} \\
\bottomrule
\end{tabular}
\vspace{0.6em}
\caption{\textbf{Best-of-16 transfer reranking deltas (\%).} Each cell reports \(\Delta\) (reward-guided minus random reranking on the same sampled generations). Columns group reward-model backbones; rows are policy-task datasets. Diagonal entries (italic) are in-distribution.}
\label{tab:transfer}
\end{table*}

\begin{figure*}[h!]
    \centering
    \begin{subfigure}{0.49\textwidth}
        \centering
        \includegraphics[width=\linewidth]{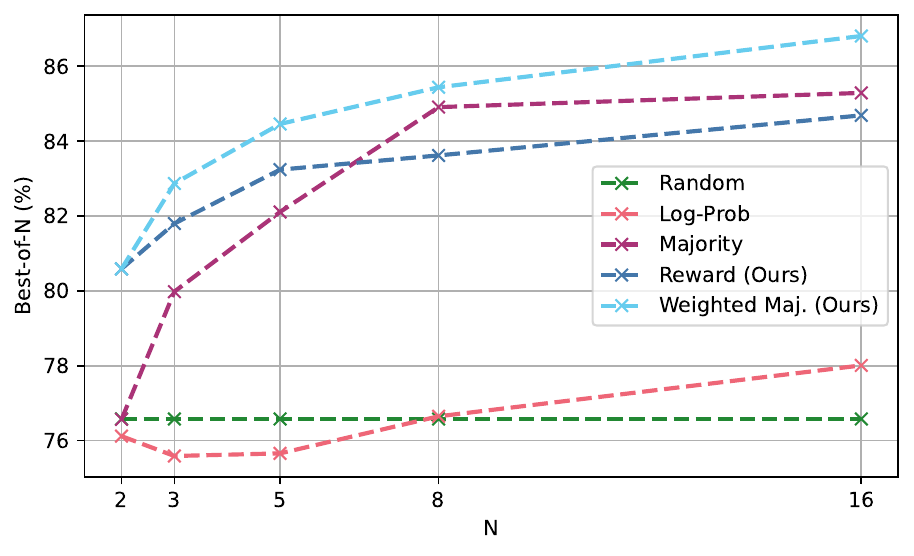}
        \caption{\textbf{Reranking baselines}\\\texttt{Qwen3-4B}, \textit{interval}, \textsc{GSM8K}.}
        \label{fig:baselines_reranking}
    \end{subfigure}\hfill
    \begin{subfigure}{0.49\textwidth}
        \centering
        \includegraphics[width=\linewidth]{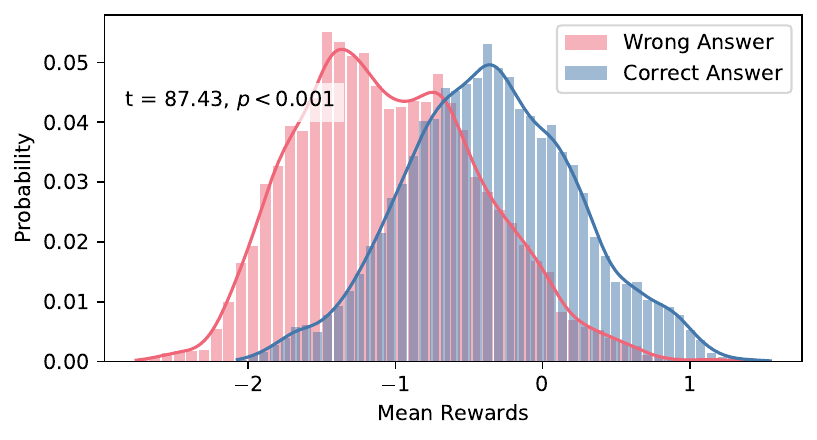}
        \caption{\textbf{Reward separation}\\\texttt{Qwen3-4B}, \textit{interval}, \textsc{MMLU-Pro}.}
        \label{fig:benefit-llama8b}
    \end{subfigure}
    \vspace{0.2em}
    \caption{\textbf{Inference-time behaviour of learned rewards.} \ref{fig:baselines_reranking}: low-budget gains from reward-guided reranking; at higher \(N\), reward-weighted majority voting is strongest. \ref{fig:benefit-llama8b}: correct traces receive higher rewards than incorrect traces (\(t=87.43, p<0.001\)).}
    \label{fig:reranking_analysis}
\end{figure*}

\textbf{Robustness across rerankers and budgets.} Across \(N \in \{2,3,5,8,16\}\), we compare log-probability, reward-guided reranking, majority voting, and reward-weighted majority voting on SFT-generated candidates (so logits remain independent of the reward model). Figure~\ref{fig:baselines_reranking} shows a consistent trend: reward-guided reranking is strongest at low \(N\), majority voting improves with \(N\), and reward-weighted majority voting performs best overall. Appendix Tables~\ref{tab:reranking_baselines_full_N2}--\ref{tab:reranking_baselines_full_N16} show the same pattern.

\textbf{Why reranking works.} Figure~\ref{fig:benefit-llama8b} shows clear reward separation between correct and incorrect traces (\(t=87.43, p<0.001\)) for \texttt{Qwen3-4B}+\textit{interval} on \textsc{MMLU-Pro}, explaining the reranking gains in Table~\ref{tab:reranking_main}. Dataset-wide separation plots are in Appendix Figures~\ref{fig:qwen7b-rewards-separation-math}--\ref{fig:lqwen4b-rewards-separation-medicine}.

\begin{customblockquote}
\textbf{Takeaway.} RQ2 is answered positively: the learned reward is a reusable inference-time ranking signal that transfers across settings and complements consistency-based voting under fixed budgets.
\end{customblockquote}

\subsection{Reasoning Error Localisation}
\label{sec:exp-d4}

\begin{greycustomblock}
\textbf{RQ3:} \textit{Do the learned reasoning rewards provide process-level feedback that localises where reasoning fails?}
\end{greycustomblock}

\begin{figure}[h!]
    \centering
    \frame{\includegraphics[width=0.8\linewidth]{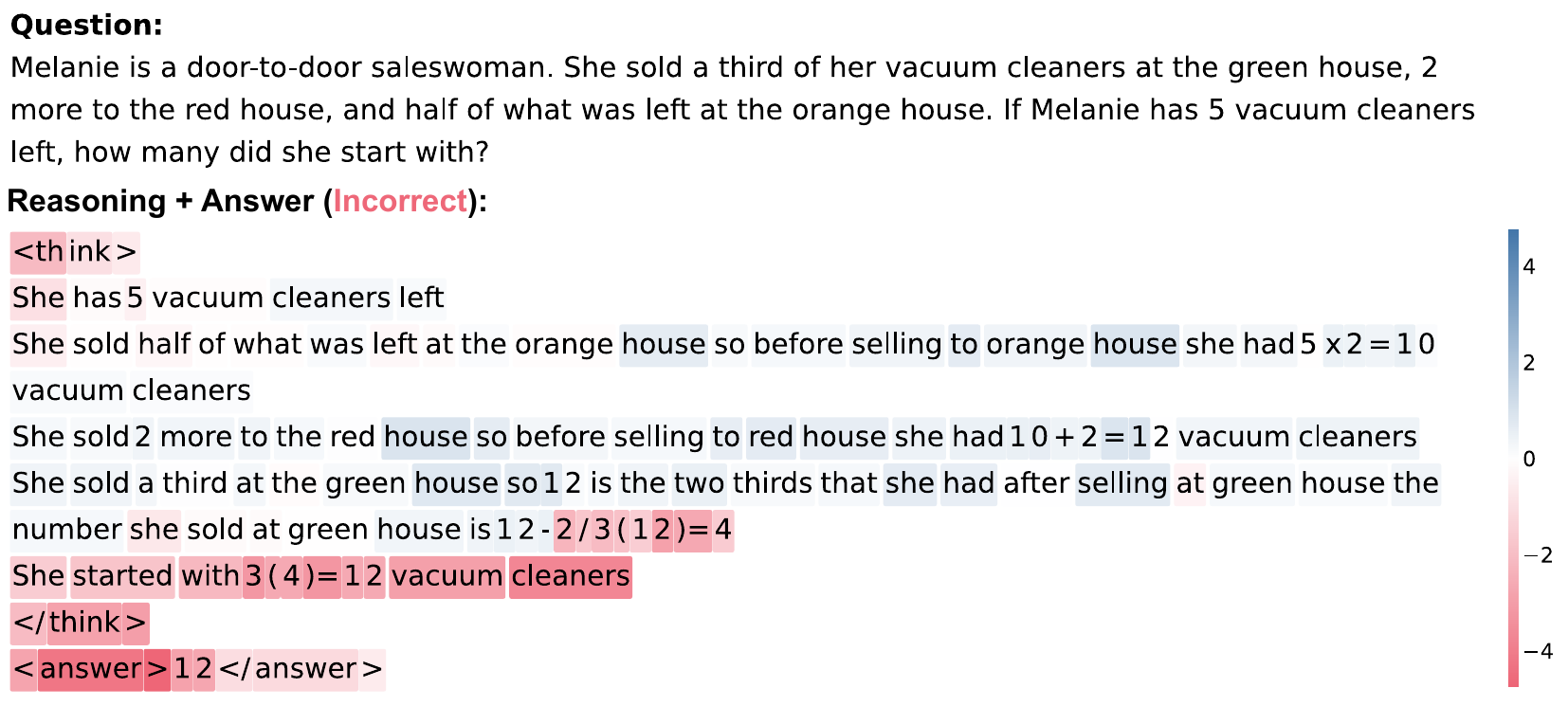}}
    \caption[Error localisation of incorrect reasoning]{\textbf{Error localisation of incorrect reasoning.} Dense reward on incorrect generation for a \textsc{GSM8K} test question, using the \textit{dense} \texttt{Qwen2.5-7B} reasoning reward model. The reward remains positive on valid intermediate steps, then drops sharply at the first erroneous step, followed by propagated penalties on later tokens.}
    \label{fig:qualitative_qwen}
\end{figure}

\textbf{Experimental setup.} To assess process-level supervision, we test whether dense rewards identify where reasoning first fails on \textsc{GSM8K}, and complement this with qualitative paired traces in Figure~\ref{fig:qualitative_qwen}. Starting from correctly answered \texttt{Qwen2.5-7B} SFT traces, we construct controlled examples with a single perturbation in the reasoning chain (operator swaps or number corruption) while keeping the final answer. We score perturbed traces with our reasoning reward models under \textit{dense} and \textit{interval} granularities. For each corrupted trace, let \(r_t\) denote the reward at token \(t\). We estimate the error location by detecting the largest one-step decrease in reward along the trace: \(\Delta_t = \max(0, r_{t-1}-r_t)\), and \(\hat{t}=\arg\max_t \Delta_t\). Intuitively, \(\hat{t}\) is the position where the reward signal drops most sharply. We count a hit if \(\hat{t}\) lies within \(\pm \{1, 7\}\) tokens of the known perturbation index (Hit@1 and Hit@7).

\begin{wraptable}{r}{0.4\linewidth}
\vspace{-0.3em}
\scriptsize
\renewcommand{\arraystretch}{1.16}
\setlength{\tabcolsep}{3.8pt}
\begin{tabular}{l c c c}
\toprule
\textbf{Model} & \multicolumn{2}{c}{\textit{Dense}} & \textit{Interval} \\
& \textbf{Hit@1} & \textbf{Hit@7} & \textbf{Hit@7} \\
\midrule
\texttt{Qwen2.5-7B} &
\makecell[c]{47.6 {\scriptsize\color{gray}$\pm$ 2.8}\\ {\scriptsize\textcolor{insightteal}{($\uparrow$ +36.2)}}} &
\makecell[c]{54.9 {\scriptsize\color{gray}$\pm$ 2.6}\\ {\scriptsize\textcolor{insightteal}{($\uparrow$ +23.6)}}} &
\makecell[c]{65.5 {\scriptsize\color{gray}$\pm$ 2.6}\\ {\scriptsize\textcolor{insightteal}{($\uparrow$ +13.2)}}} \\
\texttt{Llama3.1-8B} &
\makecell[c]{65.8 {\scriptsize\color{gray}$\pm$ 2.8}\\ {\scriptsize\textcolor{insightteal}{($\uparrow$ +53.7)}}} &
\makecell[c]{80.6 {\scriptsize\color{gray}$\pm$ 2.2}\\ {\scriptsize\textcolor{insightteal}{($\uparrow$ +47.5)}}} &
\makecell[c]{86.1 {\scriptsize\color{gray}$\pm$ 2.0}\\ {\scriptsize\textcolor{insightteal}{($\uparrow$ +30.1)}}} \\
\texttt{Qwen3-4B} &
\makecell[c]{62.0 {\scriptsize\color{gray}$\pm$ 2.7}\\ {\scriptsize\textcolor{insightteal}{($\uparrow$ +50.3)}}} &
\makecell[c]{70.3 {\scriptsize\color{gray}$\pm$ 2.5}\\ {\scriptsize\textcolor{insightteal}{($\uparrow$ +38.5)}}} &
\makecell[c]{78.9 {\scriptsize\color{gray}$\pm$ 2.2}\\ {\scriptsize\textcolor{insightteal}{($\uparrow$ +26.0)}}} \\
\bottomrule
\end{tabular}
\vspace{0.2em}
\caption{\textbf{Error Localisation.} Hit@$k$ (\%, mean \(\pm\) 95\% CI half-width). \textcolor{insightteal}{($\uparrow$ +\(\Delta\)pp)} denotes gain over chance. \textit{Interval} uses 15-token buckets, so Hit@1 = Hit@7.}
\label{tab:localisation_hit1at15_pregenerated_full_vs_partial_raw}
\vspace{-2em}
\end{wraptable}


\textbf{Localisation performance.} Table~\ref{tab:localisation_hit1at15_pregenerated_full_vs_partial_raw} shows strong localisation across all backbones, with best scores of 65.5 (\texttt{Qwen2.5-7B}), 86.1 (\texttt{Llama3.1-8B}), and 78.9 (\texttt{Qwen3-4B}). All scores are significantly above the empirical chance, indicating a better-than-random chance of scoring the reward drop within $\pm1$ and $\pm7$ tokens of the failure cases. In Appendix~\ref{app:error_localisation_app}, we demonstrate examples of the diverging reasoning trajectories on correct and corrupted traces after the first mistake.

\textbf{Error localisation on natural traces.} Figure~\ref{fig:qualitative_qwen} shows the similar behaviour qualitatively, on correctly vs incorrectly generated traces: rewards stay positive on locally correct intermediate steps and drop sharply at the first erroneous step (here at \textcolor{purple}{``$12 - 2/3(12) = 4$''}). This suggests the reward is sensitive to where reasoning first diverges. Additional examples on all datasets are in Appendix Figures~\ref{fig:qwen7b-reasoning-math}-\ref{fig:llama8b-reasoning-medicine}.

\vspace{-1em}
\begin{customblockquote}
\textbf{Takeaway.} RQ3 is answered positively: learned process-level rewards provide interpretable diagnostics that localise where reasoning first diverges, beyond final-answer evaluation.
\end{customblockquote}

\section{Limitations}\label{sec:limitations}
Our approach has three main limitations. Reward density trades off with optimisation stability: sparse step-wise signals are generally robust, whereas dense signals are more prone to adversarial mode collapse, non-stationarity, and occasional reward--correctness decoupling. R-AIRL is also computationally heavier due to alternating adversarial updates, and our evaluation is limited to arithmetic, medical, and scientific reasoning rather than open-ended or long-context tasks.

\section{Conclusion}
We presented Reasoning Adversarial Inverse Reinforcement Learning (R-AIRL), a framework for learning reasoning rewards from expert demonstrations without requiring external verifiers. Instead of directly imitating expert trajectories, R-AIRL learns a reusable reward model across the full pipeline. Empirically, this reward is useful in three ways: as a training signal (\boldcircle{D1}), where R-AIRL outperforms SFT in 7/9 demonstration-only settings; as an inference-time reranker (\boldcircle{D2}), where reward-guided reranking improves pass@1 in 26/27 settings at \(N=16\), with gains up to +17.4 points and uniformly positive transfer across all 18 source--target pairs; and as process-level feedback (\boldcircle{D3}), where it localises reasoning errors with Hit@7 up to 86.1\% (30.1\% above chance). Remaining limitations include the trade-off between reward granularity and optimisation stability, which we leave as an important direction for future work.

\newpage
\section*{Acknowledgements and Disclosure of Funding}
We thank our industry collaborators, Yusuke Kano, Jeremy Voisey, and Alison Q Smithard, for their insightful discussions. In addition, we thank Kasia Kobalczyk, Paulius Rauba, and Julianna Piskorz for their valuable feedback. Canon Inc. funds CF's studentship. The W.D. Armstrong Trust Fund and the Cystic Fibrosis Fund support NA's studentship. This work was supported by Microsoft’s Accelerate Foundation Models Academic Research initiative.

\bibliographystyle{unsrtnat}
\bibliography{references_own}

\begin{thebibliography}{44}
\providecommand{\natexlab}[1]{#1}
\providecommand{\url}[1]{\texttt{#1}}
\expandafter\ifx\csname urlstyle\endcsname\relax
  \providecommand{\doi}[1]{doi: #1}\else
  \providecommand{\doi}{doi: \begingroup \urlstyle{rm}\Url}\fi

\bibitem[Guo et~al.(2025)Guo, Yang, Zhang, Song, Wang, Zhu, Xu, Zhang, Ma, Bi, Zhang, Yu, Wu, Wu, Gou, Shao, Li, Gao, Liu, Xue, Wang, Wu, Feng, Lu, Zhao, Deng, Ruan, Dai, Chen, Ji, Li, Lin, Dai, Luo, Hao, Chen, Li, Zhang, Xu, Ding, Gao, Qu, Li, Guo, Li, Chen, Yuan, Tu, Qiu, Li, Cai, Ni, Liang, Chen, Dong, Hu, You, Gao, Guan, Huang, Yu, Wang, Zhang, Zhao, Wang, Zhang, Xu, Xia, Zhang, Zhang, Tang, Zhou, Li, Wang, Li, Tian, Huang, Zhang, Wang, Chen, Du, Ge, Zhang, Pan, Wang, Chen, Jin, Chen, Lu, Zhou, Chen, Ye, Wang, Yu, Zhou, Pan, Li, Zhou, Wu, Yun, Pei, Sun, Wang, Zeng, Liu, Liang, Gao, Yu, Zhang, Xiao, An, Liu, Wang, Chen, Nie, Cheng, Liu, Xie, Liu, Yang, Li, Su, Lin, Li, Jin, Shen, Chen, Sun, Wang, Song, Zhou, Wang, Shan, Li, Wang, Wei, Zhang, Xu, Li, Zhao, Sun, Wang, Yu, Zhang, Shi, Xiong, He, Piao, Wang, Tan, Ma, Liu, Guo, Ou, Wang, Gong, Zou, He, Xiong, Luo, You, Liu, Zhou, Zhu, Huang, Li, Zheng, Zhu, Ma, Tang, Zha, Yan, Ren, Ren, Sha, Fu, Xu, Xie, Zhang, Hao, Ma, Yan, Wu, Gu, Zhu, Liu, Li, Xie, Song,
  Pan, Huang, Xu, Zhang, and Zhang]{Guo2025}
Daya Guo, Dejian Yang, Haowei Zhang, Junxiao Song, Peiyi Wang, Qihao Zhu, Runxin Xu, Ruoyu Zhang, Shirong Ma, Xiao Bi, Xiaokang Zhang, Xingkai Yu, Yu~Wu, Z.~F. Wu, Zhibin Gou, Zhihong Shao, Zhuoshu Li, Ziyi Gao, Aixin Liu, Bing Xue, Bingxuan Wang, Bochao Wu, Bei Feng, Chengda Lu, Chenggang Zhao, Chengqi Deng, Chong Ruan, Damai Dai, Deli Chen, Dongjie Ji, Erhang Li, Fangyun Lin, Fucong Dai, Fuli Luo, Guangbo Hao, Guanting Chen, Guowei Li, H.~Zhang, Hanwei Xu, Honghui Ding, Huazuo Gao, Hui Qu, Hui Li, Jianzhong Guo, Jiashi Li, Jingchang Chen, Jingyang Yuan, Jinhao Tu, Junjie Qiu, Junlong Li, J.~L. Cai, Jiaqi Ni, Jian Liang, Jin Chen, Kai Dong, Kai Hu, Kaichao You, Kaige Gao, Kang Guan, Kexin Huang, Kuai Yu, Lean Wang, Lecong Zhang, Liang Zhao, Litong Wang, Liyue Zhang, Lei Xu, Leyi Xia, Mingchuan Zhang, Minghua Zhang, Minghui Tang, Mingxu Zhou, Meng Li, Miaojun Wang, Mingming Li, Ning Tian, Panpan Huang, Peng Zhang, Qiancheng Wang, Qinyu Chen, Qiushi Du, Ruiqi Ge, Ruisong Zhang, Ruizhe Pan, Runji Wang, R.~J.
  Chen, R.~L. Jin, Ruyi Chen, Shanghao Lu, Shangyan Zhou, Shanhuang Chen, Shengfeng Ye, Shiyu Wang, Shuiping Yu, Shunfeng Zhou, Shuting Pan, S.~S. Li, Shuang Zhou, Shaoqing Wu, Tao Yun, Tian Pei, Tianyu Sun, T.~Wang, Wangding Zeng, Wen Liu, Wenfeng Liang, Wenjun Gao, Wenqin Yu, Wentao Zhang, W.~L. Xiao, Wei An, Xiaodong Liu, Xiaohan Wang, Xiaokang Chen, Xiaotao Nie, Xin Cheng, Xin Liu, Xin Xie, Xingchao Liu, Xinyu Yang, Xinyuan Li, Xuecheng Su, Xuheng Lin, X.~Q. Li, Xiangyue Jin, Xiaojin Shen, Xiaosha Chen, Xiaowen Sun, Xiaoxiang Wang, Xinnan Song, Xinyi Zhou, Xianzu Wang, Xinxia Shan, Y.~K. Li, Y.~Q. Wang, Y.~X. Wei, Yang Zhang, Yanhong Xu, Yao Li, Yao Zhao, Yaofeng Sun, Yaohui Wang, Yi~Yu, Yichao Zhang, Yifan Shi, Yiliang Xiong, Ying He, Yishi Piao, Yisong Wang, Yixuan Tan, Yiyang Ma, Yiyuan Liu, Yongqiang Guo, Yuan Ou, Yuduan Wang, Yue Gong, Yuheng Zou, Yujia He, Yunfan Xiong, Yuxiang Luo, Yuxiang You, Yuxuan Liu, Yuyang Zhou, Y.~X. Zhu, Yanping Huang, Yaohui Li, Yi~Zheng, Yuchen Zhu, Yunxian Ma, Ying
  Tang, Yukun Zha, Yuting Yan, Z.~Z. Ren, Zehui Ren, Zhangli Sha, Zhe Fu, Zhean Xu, Zhenda Xie, Zhengyan Zhang, Zhewen Hao, Zhicheng Ma, Zhigang Yan, Zhiyu Wu, Zihui Gu, Zijia Zhu, Zijun Liu, Zilin Li, Ziwei Xie, Ziyang Song, Zizheng Pan, Zhen Huang, Zhipeng Xu, Zhongyu Zhang, and Zhen Zhang.
\newblock Deepseek-r1 incentivizes reasoning in llms through reinforcement learning.
\newblock \emph{Nature}, 645\penalty0 (8081):\penalty0 633–638, 2025.
\newblock ISSN 1476-4687.
\newblock \doi{10.1038/s41586-025-09422-z}.
\newblock URL \url{http://dx.doi.org/10.1038/s41586-025-09422-z}.

\bibitem[Uesato et~al.(2022)Uesato, Kushman, Kumar, Song, Siegel, Wang, Creswell, Irving, and Higgins]{uesato_solving_2022}
Jonathan Uesato, Nate Kushman, Ramana Kumar, Francis Song, Noah Siegel, Lisa Wang, Antonia Creswell, Geoffrey Irving, and Irina Higgins.
\newblock Solving math word problems with process- and outcome-based feedback, November 2022.
\newblock URL \url{http://arxiv.org/abs/2211.14275}.
\newblock arXiv:2211.14275 [cs].

\bibitem[Lightman et~al.(2023)Lightman, Kosaraju, Burda, Edwards, Baker, Lee, Leike, Schulman, Sutskever, and Cobbe]{lightman_lets_2023}
Hunter Lightman, Vineet Kosaraju, Yura Burda, Harri Edwards, Bowen Baker, Teddy Lee, Jan Leike, John Schulman, Ilya Sutskever, and Karl Cobbe.
\newblock Let's {Verify} {Step} by {Step}, May 2023.
\newblock URL \url{http://arxiv.org/abs/2305.20050}.
\newblock arXiv:2305.20050 [cs].

\bibitem[Setlur et~al.(2025)Setlur, Rajaraman, Levine, and Kumar]{setlur_scaling_2025}
Amrith Setlur, Nived Rajaraman, Sergey Levine, and Aviral Kumar.
\newblock Scaling {Test}-{Time} {Compute} {Without} {Verification} or {RL} is {Suboptimal}, February 2025.
\newblock URL \url{http://arxiv.org/abs/2502.12118}.
\newblock arXiv:2502.12118 [cs].

\bibitem[Abbeel and Ng(2004)]{abbeel_apprenticeship_2004}
Pieter Abbeel and Andrew~Y. Ng.
\newblock Apprenticeship learning via inverse reinforcement learning.
\newblock In \emph{Twenty-first international conference on {Machine} learning - {ICML} '04}, page~1, Banff, Alberta, Canada, 2004. ACM Press.
\newblock \doi{10.1145/1015330.1015430}.
\newblock URL \url{http://portal.acm.org/citation.cfm?doid=1015330.1015430}.

\bibitem[Ziebart et~al.(2004)Ziebart, Maas, Bagnell, and Dey]{ziebart_maximum_nodate}
Brian~D Ziebart, Andrew Maas, J~Andrew Bagnell, and Anind~K Dey.
\newblock Maximum {Entropy} {Inverse} {Reinforcement} {Learning}.
\newblock 2004.

\bibitem[Ho and Ermon(2016)]{ho_generative_2016}
Jonathan Ho and Stefano Ermon.
\newblock Generative {Adversarial} {Imitation} {Learning}, June 2016.
\newblock URL \url{http://arxiv.org/abs/1606.03476}.
\newblock arXiv:1606.03476 [cs].

\bibitem[Fu et~al.(2018)Fu, Luo, and Levine]{fu_learning_2018}
Justin Fu, Katie Luo, and Sergey Levine.
\newblock Learning {Robust} {Rewards} with {Adversarial} {Inverse} {Reinforcement} {Learning}, August 2018.
\newblock URL \url{http://arxiv.org/abs/1710.11248}.
\newblock arXiv:1710.11248 [cs].

\bibitem[Yu et~al.(2025)Yu, Zhang, Zhu, Yuan, Zuo, Yue, Dai, Fan, Liu, Liu, Liu, Lin, Lin, Ma, Sheng, Tong, Zhang, Zhang, Zhang, Zhu, Zhu, Chen, Chen, Wang, Yu, Song, Wei, Zhou, Liu, Ma, Zhang, Yan, Qiao, Wu, and Wang]{yu_dapo_2025}
Qiying Yu, Zheng Zhang, Ruofei Zhu, Yufeng Yuan, Xiaochen Zuo, Yu~Yue, Weinan Dai, Tiantian Fan, Gaohong Liu, Lingjun Liu, Xin Liu, Haibin Lin, Zhiqi Lin, Bole Ma, Guangming Sheng, Yuxuan Tong, Chi Zhang, Mofan Zhang, Wang Zhang, Hang Zhu, Jinhua Zhu, Jiaze Chen, Jiangjie Chen, Chengyi Wang, Hongli Yu, Yuxuan Song, Xiangpeng Wei, Hao Zhou, Jingjing Liu, Wei-Ying Ma, Ya-Qin Zhang, Lin Yan, Mu~Qiao, Yonghui Wu, and Mingxuan Wang.
\newblock {DAPO}: {An} {Open}-{Source} {LLM} {Reinforcement} {Learning} {System} at {Scale}, May 2025.
\newblock URL \url{http://arxiv.org/abs/2503.14476}.
\newblock arXiv:2503.14476 [cs].

\bibitem[YuYue et~al.(2025)YuYue, Yuan, Yu, Zuo, Zhu, Xu, Chen, Wang, Fan, Du, Wei, Liu, Liu, Liu, Lin, Lin, Ma, Zhang, Zhang, Zhang, Zhu, Zhang, Liu, Wang, Wu, and Yan]{yuyue_vapo_2025}
YuYue, Yufeng Yuan, Qiying Yu, Xiaochen Zuo, Ruofei Zhu, Wenyuan Xu, Jiaze Chen, Chengyi Wang, TianTian Fan, Zhengyin Du, Xiangpeng Wei, Gaohong Liu, Juncai Liu, Lingjun Liu, Haibin Lin, Zhiqi Lin, Bole Ma, Chi Zhang, Mofan Zhang, Wang Zhang, Hang Zhu, Ru~Zhang, Xin Liu, Mingxuan Wang, Yonghui Wu, and Lin Yan.
\newblock {VAPO}: {Efficient} and {Reliable} {Reinforcement} {Learning} for {Advanced} {Reasoning} {Tasks}, April 2025.
\newblock URL \url{http://arxiv.org/abs/2504.05118}.
\newblock arXiv:2504.05118 [cs] version: 1.

\bibitem[Zelikman et~al.(2022)Zelikman, Wu, Mu, and Goodman]{zelikman_star_2022}
Eric Zelikman, Yuhuai Wu, Jesse Mu, and Noah~D. Goodman.
\newblock {STaR}: {Bootstrapping} {Reasoning} {With} {Reasoning}, May 2022.
\newblock URL \url{http://arxiv.org/abs/2203.14465}.
\newblock arXiv:2203.14465 [cs].

\bibitem[Yuan et~al.(2023)Yuan, Yuan, Li, Dong, Lu, Tan, Zhou, and Zhou]{yuan_scaling_2023}
Zheng Yuan, Hongyi Yuan, Chengpeng Li, Guanting Dong, Keming Lu, Chuanqi Tan, Chang Zhou, and Jingren Zhou.
\newblock Scaling {Relationship} on {Learning} {Mathematical} {Reasoning} with {Large} {Language} {Models}, September 2023.
\newblock URL \url{http://arxiv.org/abs/2308.01825}.
\newblock arXiv:2308.01825 [cs].

\bibitem[Singh et~al.(2024)Singh, Co-Reyes, Agarwal, Anand, Patil, Garcia, Liu, Harrison, Lee, Xu, Parisi, Kumar, Alemi, Rizkowsky, Nova, Adlam, Bohnet, Elsayed, Sedghi, Mordatch, Simpson, Gur, Snoek, Pennington, Hron, Kenealy, Swersky, Mahajan, Culp, Xiao, Bileschi, Constant, Novak, Liu, Warkentin, Bansal, Dyer, Neyshabur, Sohl-Dickstein, and Fiedel]{singh_beyond_2024}
Avi Singh, John~D. Co-Reyes, Rishabh Agarwal, Ankesh Anand, Piyush Patil, Xavier Garcia, Peter~J. Liu, James Harrison, Jaehoon Lee, Kelvin Xu, Aaron~T. Parisi, Abhishek Kumar, Alexander~A. Alemi, Alex Rizkowsky, Azade Nova, Ben Adlam, Bernd Bohnet, Gamaleldin~Fathy Elsayed, Hanie Sedghi, Igor Mordatch, Isabelle Simpson, Izzeddin Gur, Jasper Snoek, Jeffrey Pennington, Jiri Hron, Kathleen Kenealy, Kevin Swersky, Kshiteej Mahajan, Laura~A. Culp, Lechao Xiao, Maxwell Bileschi, Noah Constant, Roman Novak, Rosanne Liu, Tris Warkentin, Yamini Bansal, Ethan Dyer, Behnam Neyshabur, Jascha Sohl-Dickstein, and Noah Fiedel.
\newblock Beyond {Human} {Data}: {Scaling} {Self}-{Training} for {Problem}-{Solving} with {Language} {Models}.
\newblock \emph{Transactions on Machine Learning Research}, January 2024.
\newblock ISSN 2835-8856.
\newblock URL \url{https://openreview.net/forum?id=lNAyUngGFK}.

\bibitem[Hosseini et~al.(2024)Hosseini, Yuan, Malkin, Courville, Sordoni, and Agarwal]{hosseini_v-star_2024}
Arian Hosseini, Xingdi Yuan, Nikolay Malkin, Aaron Courville, Alessandro Sordoni, and Rishabh Agarwal.
\newblock V-{STaR}: {Training} {Verifiers} for {Self}-{Taught} {Reasoners}.
\newblock August 2024.
\newblock URL \url{https://openreview.net/forum?id=stmqBSW2dV#discussion}.

\bibitem[Silver et~al.(2017)Silver, Schrittwieser, Simonyan, Antonoglou, Huang, Guez, Hubert, baker, Lai, Bolton, Chen, Lillicrap, Hui, Sifre, van~den Driessche, Graepel, and Hassabis]{Silver2017MasteringTG}
David Silver, Julian Schrittwieser, Karen Simonyan, Ioannis Antonoglou, Aja Huang, Arthur Guez, Thomas Hubert, Lucas baker, Matthew Lai, Adrian Bolton, Yutian Chen, Timothy~P. Lillicrap, Fan Hui, L.~Sifre, George van~den Driessche, Thore Graepel, and Demis Hassabis.
\newblock Mastering the game of go without human knowledge.
\newblock \emph{Nature}, 550:\penalty0 354--359, 2017.
\newblock URL \url{https://api.semanticscholar.org/CorpusID:205261034}.

\bibitem[Christiano et~al.(2023)Christiano, Leike, Brown, Martic, Legg, and Amodei]{christiano_deep_2023}
Paul Christiano, Jan Leike, Tom~B. Brown, Miljan Martic, Shane Legg, and Dario Amodei.
\newblock Deep reinforcement learning from human preferences, February 2023.
\newblock URL \url{http://arxiv.org/abs/1706.03741}.
\newblock arXiv:1706.03741.

\bibitem[Rafailov et~al.(2024)Rafailov, Sharma, Mitchell, Ermon, Manning, and Finn]{rafailov_direct_2024}
Rafael Rafailov, Archit Sharma, Eric Mitchell, Stefano Ermon, Christopher~D. Manning, and Chelsea Finn.
\newblock Direct {Preference} {Optimization}: {Your} {Language} {Model} is {Secretly} a {Reward} {Model}, July 2024.
\newblock URL \url{http://arxiv.org/abs/2305.18290}.
\newblock arXiv:2305.18290.

\bibitem[Hinton et~al.(2015)Hinton, Vinyals, and Dean]{hinton_distilling_2015}
Geoffrey Hinton, Oriol Vinyals, and Jeff Dean.
\newblock Distilling the {Knowledge} in a {Neural} {Network}, March 2015.
\newblock URL \url{http://arxiv.org/abs/1503.02531}.
\newblock arXiv:1503.02531 [stat].

\bibitem[Kang et~al.(2023)Kang, Lee, Baek, Kawaguchi, and Hwang]{kang_knowledge-augmented_2023}
Minki Kang, Seanie Lee, Jinheon Baek, Kenji Kawaguchi, and Sung~Ju Hwang.
\newblock Knowledge-{Augmented} {Reasoning} {Distillation} for {Small} {Language} {Models} in {Knowledge}-{Intensive} {Tasks}.
\newblock In A.~Oh, T.~Naumann, A.~Globerson, K.~Saenko, M.~Hardt, and S.~Levine, editors, \emph{Advances in {Neural} {Information} {Processing} {Systems}}, volume~36, pages 48573--48602. Curran Associates, Inc., 2023.
\newblock URL \url{https://proceedings.neurips.cc/paper_files/paper/2023/file/97faedc90260eae5c400f92d5831c3d7-Paper-Conference.pdf}.

\bibitem[Kujanpää et~al.(2025)Kujanpää, Marttinen, Valpola, and Ilin]{kujanpaa_efficient_2025}
Kalle Kujanpää, Pekka Marttinen, Harri Valpola, and Alexander Ilin.
\newblock Efficient {Knowledge} {Injection} in {LLMs} via {Self}-{Distillation}, August 2025.
\newblock URL \url{http://arxiv.org/abs/2412.14964}.
\newblock arXiv:2412.14964 [cs].

\bibitem[Xu et~al.(2025)Xu, Zhu, Deng, Li, Hou, Wang, Shang, Xu, and Mi]{xu_kdrl_2025}
Hongling Xu, Qi~Zhu, Heyuan Deng, Jinpeng Li, Lu~Hou, Yasheng Wang, Lifeng Shang, Ruifeng Xu, and Fei Mi.
\newblock {KDRL}: {Post}-{Training} {Reasoning} {LLMs} via {Unified} {Knowledge} {Distillation} and {Reinforcement} {Learning}, June 2025.
\newblock URL \url{http://arxiv.org/abs/2506.02208}.
\newblock arXiv:2506.02208 [cs].

\bibitem[Shenfeld et~al.(2026)Shenfeld, Damani, H{\"u}botter, and Agrawal]{shenfeld2026selfdistillation}
Idan Shenfeld, Mehul Damani, Jonas H{\"u}botter, and Pulkit Agrawal.
\newblock Self-distillation enables continual learning.
\newblock In \emph{ICLR 2026 Workshop on Lifelong Agents: Learning, Aligning, Evolving}, 2026.
\newblock URL \url{https://openreview.net/forum?id=HlWA3V6iKF}.

\bibitem[Ye et~al.(2025)Ye, Dong, Chi, Wu, Huang, and Wei]{ye2025blackboxonpolicydistillationlarge}
Tianzhu Ye, Li~Dong, Zewen Chi, Xun Wu, Shaohan Huang, and Furu Wei.
\newblock Black-box on-policy distillation of large language models.
\newblock \emph{arXiv preprint arXiv:2511.10643}, 2025.
\newblock URL \url{https://arxiv.org/abs/2511.10643}.

\bibitem[Lee et~al.(2025)Lee, Hachiuma, Ro, Wang, and Wu]{lee2025unified}
Byung-Kwan Lee, Ryo Hachiuma, Yong~Man Ro, Yu-Chiang~Frank Wang, and Yueh-Hua Wu.
\newblock Unified reinforcement and imitation learning for vision-language models.
\newblock In \emph{The Thirty-ninth Annual Conference on Neural Information Processing Systems}, 2025.
\newblock URL \url{https://openreview.net/forum?id=7wEvjzkNXg}.

\bibitem[Shao et~al.(2024)Shao, Wang, Zhu, Xu, Song, Bi, Zhang, Zhang, Li, Wu, and Guo]{shao_deepseekmath_2024}
Zhihong Shao, Peiyi Wang, Qihao Zhu, Runxin Xu, Junxiao Song, Xiao Bi, Haowei Zhang, Mingchuan Zhang, Y.~K. Li, Y.~Wu, and Daya Guo.
\newblock {DeepSeekMath}: {Pushing} the {Limits} of {Mathematical} {Reasoning} in {Open} {Language} {Models}, April 2024.
\newblock URL \url{http://arxiv.org/abs/2402.03300}.
\newblock arXiv:2402.03300 [cs].

\bibitem[Cetin et~al.(2025)Cetin, Zhao, and Tang]{cetin2025reinforcement}
Edoardo Cetin, Tianyu Zhao, and Yujin Tang.
\newblock Reinforcement learning teachers of test time scaling.
\newblock In \emph{The Thirty-ninth Annual Conference on Neural Information Processing Systems}, 2025.
\newblock URL \url{https://openreview.net/forum?id=tebG8q5EeK}.

\bibitem[Cui et~al.(2025)Cui, Yuan, Wang, Wang, Li, He, Fan, Yu, Xu, Chen, Yuan, Chen, Zhang, Lv, Wang, Yao, Han, Peng, Cheng, Liu, Sun, Zhou, and Ding]{cui_process_2025}
Ganqu Cui, Lifan Yuan, Zefan Wang, Hanbin Wang, Wendi Li, Bingxiang He, Yuchen Fan, Tianyu Yu, Qixin Xu, Weize Chen, Jiarui Yuan, Huayu Chen, Kaiyan Zhang, Xingtai Lv, Shuo Wang, Yuan Yao, Xu~Han, Hao Peng, Yu~Cheng, Zhiyuan Liu, Maosong Sun, Bowen Zhou, and Ning Ding.
\newblock Process {Reinforcement} through {Implicit} {Rewards}, February 2025.
\newblock URL \url{http://arxiv.org/abs/2502.01456}.
\newblock arXiv:2502.01456 [cs].

\bibitem[Schulman et~al.(2017)Schulman, Wolski, Dhariwal, Radford, and Klimov]{schulman_proximal_2017}
John Schulman, Filip Wolski, Prafulla Dhariwal, Alec Radford, and Oleg Klimov.
\newblock Proximal {Policy} {Optimization} {Algorithms}, August 2017.
\newblock URL \url{http://arxiv.org/abs/1707.06347}.
\newblock arXiv:1707.06347 [cs].

\bibitem[Cobbe et~al.(2021)Cobbe, Kosaraju, Bavarian, Chen, Jun, Kaiser, Plappert, Tworek, Hilton, Nakano, Hesse, and Schulman]{cobbe_training_2021}
Karl Cobbe, Vineet Kosaraju, Mohammad Bavarian, Mark Chen, Heewoo Jun, Lukasz Kaiser, Matthias Plappert, Jerry Tworek, Jacob Hilton, Reiichiro Nakano, Christopher Hesse, and John Schulman.
\newblock Training {Verifiers} to {Solve} {Math} {Word} {Problems}, November 2021.
\newblock URL \url{http://arxiv.org/abs/2110.14168}.
\newblock arXiv:2110.14168 [cs].

\bibitem[Wu et~al.(2025)Wu, Deng, Li, Liu, Mi, Peng, Xu, Liu, Cho, Choi, Cao, Ren, Li, Li, and Zhou]{wu2025medreasonelicitingfactualmedical}
Juncheng Wu, Wenlong Deng, Xingxuan Li, Sheng Liu, Taomian Mi, Yifan Peng, Ziyang Xu, Yi~Liu, Hyunjin Cho, Chang-In Choi, Yihan Cao, Hui Ren, Xiang Li, Xiaoxiao Li, and Yuyin Zhou.
\newblock Medreason: Eliciting factual medical reasoning steps in llms via knowledge graphs, 2025.
\newblock URL \url{https://arxiv.org/abs/2504.00993}.

\bibitem[Wang et~al.(2024)Wang, Ma, Zhang, Ni, Chandra, Guo, Ren, Arulraj, He, Jiang, et~al.]{wang2024mmlu}
Yubo Wang, Xueguang Ma, Ge~Zhang, Yuansheng Ni, Abhranil Chandra, Shiguang Guo, Weiming Ren, Aaran Arulraj, Xuan He, Ziyan Jiang, et~al.
\newblock Mmlu-pro: A more robust and challenging multi-task language understanding benchmark.
\newblock \emph{arXiv preprint arXiv:2406.01574}, 2024.

\bibitem[Bai et~al.(2023)Bai, Bai, Chu, Cui, Dang, Deng, Fan, Ge, Han, Huang, Hui, Ji, Li, Lin, Lin, Liu, Liu, Lu, Lu, Ma, Men, Ren, Ren, Tan, Tan, Tu, Wang, Wang, Wang, Wu, Xu, Xu, Yang, Yang, Yang, Yang, Yao, Yu, Yuan, Yuan, Zhang, Zhang, Zhang, Zhang, Zhou, Zhou, Zhou, and Zhu]{bai_qwen_2023}
Jinze Bai, Shuai Bai, Yunfei Chu, Zeyu Cui, Kai Dang, Xiaodong Deng, Yang Fan, Wenbin Ge, Yu~Han, Fei Huang, Binyuan Hui, Luo Ji, Mei Li, Junyang Lin, Runji Lin, Dayiheng Liu, Gao Liu, Chengqiang Lu, Keming Lu, Jianxin Ma, Rui Men, Xingzhang Ren, Xuancheng Ren, Chuanqi Tan, Sinan Tan, Jianhong Tu, Peng Wang, Shijie Wang, Wei Wang, Shengguang Wu, Benfeng Xu, Jin Xu, An~Yang, Hao Yang, Jian Yang, Shusheng Yang, Yang Yao, Bowen Yu, Hongyi Yuan, Zheng Yuan, Jianwei Zhang, Xingxuan Zhang, Yichang Zhang, Zhenru Zhang, Chang Zhou, Jingren Zhou, Xiaohuan Zhou, and Tianhang Zhu.
\newblock Qwen {Technical} {Report}, September 2023.
\newblock URL \url{http://arxiv.org/abs/2309.16609}.
\newblock arXiv:2309.16609 [cs].

\bibitem[Touvron et~al.(2023)Touvron, Lavril, Izacard, Martinet, Lachaux, Lacroix, Rozière, Goyal, Hambro, Azhar, Rodriguez, Joulin, Grave, and Lample]{touvron_llama_2023}
Hugo Touvron, Thibaut Lavril, Gautier Izacard, Xavier Martinet, Marie-Anne Lachaux, Timothée Lacroix, Baptiste Rozière, Naman Goyal, Eric Hambro, Faisal Azhar, Aurelien Rodriguez, Armand Joulin, Edouard Grave, and Guillaume Lample.
\newblock {LLaMA}: {Open} and {Efficient} {Foundation} {Language} {Models}, February 2023.
\newblock URL \url{http://arxiv.org/abs/2302.13971}.
\newblock arXiv:2302.13971 [cs].

\bibitem[Yang et~al.(2025)Yang, Li, Yang, Zhang, Hui, Zheng, Yu, Gao, Huang, Lv, Zheng, Liu, Zhou, Huang, Hu, Ge, Wei, Lin, Tang, Yang, Tu, Zhang, Yang, Yang, Zhou, Zhou, Lin, Dang, Bao, Yang, Yu, Deng, Li, Xue, Li, Zhang, Wang, Zhu, Men, Gao, Liu, Luo, Li, Tang, Yin, Ren, Wang, Zhang, Ren, Fan, Su, Zhang, Zhang, Wan, Liu, Wang, Cui, Zhang, Zhou, and Qiu]{qwen3}
An~Yang, Anfeng Li, Baosong Yang, Beichen Zhang, Binyuan Hui, Bo~Zheng, Bowen Yu, Chang Gao, Chengen Huang, Chenxu Lv, Chujie Zheng, Dayiheng Liu, Fan Zhou, Fei Huang, Feng Hu, Hao Ge, Haoran Wei, Huan Lin, Jialong Tang, Jian Yang, Jianhong Tu, Jianwei Zhang, Jianxin Yang, Jiaxi Yang, Jing Zhou, Jingren Zhou, Junyang Lin, Kai Dang, Keqin Bao, Kexin Yang, Le~Yu, Lianghao Deng, Mei Li, Mingfeng Xue, Mingze Li, Pei Zhang, Peng Wang, Qin Zhu, Rui Men, Ruize Gao, Shixuan Liu, Shuang Luo, Tianhao Li, Tianyi Tang, Wenbiao Yin, Xingzhang Ren, Xinyu Wang, Xinyu Zhang, Xuancheng Ren, Yang Fan, Yang Su, Yichang Zhang, Yinger Zhang, Yu~Wan, Yuqiong Liu, Zekun Wang, Zeyu Cui, Zhenru Zhang, Zhipeng Zhou, and Zihan Qiu.
\newblock Qwen3 technical report.
\newblock \emph{arXiv preprint arXiv:2505.09388}, 2025.

\bibitem[Heusel et~al.(2018)Heusel, Ramsauer, Unterthiner, Nessler, and Hochreiter]{heusel2018ganstrainedtimescaleupdate}
Martin Heusel, Hubert Ramsauer, Thomas Unterthiner, Bernhard Nessler, and Sepp Hochreiter.
\newblock Gans trained by a two time-scale update rule converge to a local nash equilibrium, 2018.
\newblock URL \url{https://arxiv.org/abs/1706.08500}.

\bibitem[Arjovsky et~al.(2017)Arjovsky, Chintala, and Bottou]{arjovsky_wasserstein_2017}
Martin Arjovsky, Soumith Chintala, and Léon Bottou.
\newblock Wasserstein {GAN}, December 2017.
\newblock URL \url{http://arxiv.org/abs/1701.07875}.
\newblock arXiv:1701.07875 [stat].

\bibitem[Jin et~al.(2021)Jin, Pan, Oufattole, Weng, Fang, and Szolovits]{jin2021disease}
Di~Jin, Eileen Pan, Nassim Oufattole, Wei-Hung Weng, Hanyi Fang, and Peter Szolovits.
\newblock What disease does this patient have? a large-scale open domain question answering dataset from medical exams.
\newblock \emph{Applied Sciences}, 11\penalty0 (14):\penalty0 6421, 2021.

\bibitem[Pal et~al.(2022)Pal, Umapathi, and Sankarasubbu]{pmlr-v174-pal22a}
Ankit Pal, Logesh~Kumar Umapathi, and Malaikannan Sankarasubbu.
\newblock Medmcqa: A large-scale multi-subject multi-choice dataset for medical domain question answering.
\newblock In Gerardo Flores, George~H Chen, Tom Pollard, Joyce~C Ho, and Tristan Naumann, editors, \emph{Proceedings of the Conference on Health, Inference, and Learning}, volume 174 of \emph{Proceedings of Machine Learning Research}, pages 248--260. PMLR, 07--08 Apr 2022.
\newblock URL \url{https://proceedings.mlr.press/v174/pal22a.html}.

\bibitem[Van~Rossum and Drake~Jr(1995)]{python}
Guido Van~Rossum and Fred~L Drake~Jr.
\newblock \emph{Python reference manual}.
\newblock Centrum voor Wiskunde en Informatica Amsterdam, 1995.

\bibitem[Paszke et~al.(2017)Paszke, Gross, Chintala, Chanan, Yang, DeVito, Lin, Desmaison, Antiga, and Lerer]{pytorch}
Adam Paszke, Sam Gross, Soumith Chintala, Gregory Chanan, Edward Yang, Zachary DeVito, Zeming Lin, Alban Desmaison, Luca Antiga, and Adam Lerer.
\newblock Automatic differentiation in pytorch.
\newblock 2017.

\bibitem[Wolf et~al.(2020)Wolf, Debut, Sanh, Chaumond, Delangue, Moi, Cistac, Rault, Louf, Funtowicz, Davison, Shleifer, Platen, Ma, Jernite, Plu, Xu, Scao, Gugger, Drame, Lhoest, and Rush]{wolf_huggingfaces_2020}
Thomas Wolf, Lysandre Debut, Victor Sanh, Julien Chaumond, Clement Delangue, Anthony Moi, Pierric Cistac, Tim Rault, Rémi Louf, Morgan Funtowicz, Joe Davison, Sam Shleifer, Patrick~von Platen, Clara Ma, Yacine Jernite, Julien Plu, Canwen Xu, Teven~Le Scao, Sylvain Gugger, Mariama Drame, Quentin Lhoest, and Alexander~M. Rush.
\newblock {HuggingFace}'s {Transformers}: {State}-of-the-art {Natural} {Language} {Processing}, July 2020.
\newblock URL \url{http://arxiv.org/abs/1910.03771}.
\newblock arXiv:1910.03771 [cs].

\bibitem[Daniel~Han and team(2023)]{unsloth}
Michael~Han Daniel~Han and Unsloth team.
\newblock Unsloth, 2023.
\newblock URL \url{http://github.com/unslothai/unsloth}.

\bibitem[Loshchilov and Hutter(2019)]{loshchilov2018decoupled}
Ilya Loshchilov and Frank Hutter.
\newblock Decoupled weight decay regularization.
\newblock In \emph{International Conference on Learning Representations}, 2019.
\newblock URL \url{https://openreview.net/forum?id=Bkg6RiCqY7}.

\bibitem[Hu et~al.(2021)Hu, Shen, Wallis, Allen-Zhu, Li, Wang, Wang, and Chen]{hu2021loralowrankadaptationlarge}
Edward~J. Hu, Yelong Shen, Phillip Wallis, Zeyuan Allen-Zhu, Yuanzhi Li, Shean Wang, Lu~Wang, and Weizhu Chen.
\newblock Lora: Low-rank adaptation of large language models, 2021.
\newblock URL \url{https://arxiv.org/abs/2106.09685}.

\end{thebibliography}

\appendix
\onecolumn
\newpage

\section{Additional Method Details}\label{app:method-details}

\subsection{Training Stability.}
A known limitation of adversarial training is that a single discriminator update per policy step leaves $D_\phi$ under-trained relative to $\pi_\theta$, making the reward signal noisy and susceptible to reward hacking \citep{heusel2018ganstrainedtimescaleupdate, arjovsky_wasserstein_2017, ye2025blackboxonpolicydistillationlarge}. To mitigate this, we use three complementary stabilisation mechanisms. First, we maintain replay buffers $\mathcal{B}_{\text{pos}}$ and $\mathcal{B}_{\text{neg}}$ that accumulate positive and negative traces across iterations; each discriminator update samples mini-batches from these buffers rather than solely from current policy rollouts, broadening the training distribution and reducing non-stationarity. Second, we perform $N_{\text{disc}}$ discriminator gradient steps per policy update, allowing $D_\phi$ to track the evolving policy more closely before the next GRPO step. Third, during policy learning, we clip token rewards:
\begin{equation}
\label{eq:reward_clipping}
r_\phi(y_t)\leftarrow \text{clip}\!\left(r_\phi(y_t),-\beta,\beta\right),\quad \text{where}\quad \beta>0
\end{equation}
where \(\beta>0\). This bounds extreme logit-derived rewards and prevents unstable updates. Together, these mechanisms better calibrate the reward signal and reduce mode-collapse-like behaviour in adversarial training.

\subsection{Algorithm}
\begin{algorithm}[h!]
\caption{Reasoning Adversarial Inverse Reinforcement Learning (R-AIRL)} \label{alg:airl_reasoning}
\begin{algorithmic}[1]
\Require Expert traces $\mathcal{D}_E$; Iterations $N_{\text{step}}$; warm-up steps $N_{\text{warm}}$; discriminator updates per step $N_{\text{disc}}$
\State Initialise policy $\pi_{\theta}$, discriminator $D_{\phi}$, and replay buffers $\mathcal{B}_{\text{pos}} \gets \emptyset$, $\mathcal{B}_{\text{neg}} \gets \emptyset$

\State \textbf{Warm-up Phase:}
\For{$j \gets 1$ to $N_{\text{warm}}$}
    \State Sample prompts $x \sim \mathcal{Q}$ and generate $\mathcal{D}_{P} \sim \pi_{\theta_{\text{init}}}(\cdot \mid x)$
    \State $\mathcal{D}_{\text{pos}} \gets \mathcal{D}_E \cup \{ y \in \mathcal{D}_P \mid \mathcal{O}(y) = \mathcal{O}(y^E) \}$
    \State $\mathcal{D}_{\text{neg}} \gets \{ y \in \mathcal{D}_P \mid \mathcal{O}(y) \neq \mathcal{O}(y^E) \} \cup \{ \mathcal{C}(y^E) \mid y^E \in \mathcal{D}_E\}$
    \State Update $D_{\phi}$ to minimise Eq.~\eqref{eq:discriminator_objective} using ${\mathcal{D}}_{\text{pos}}$ and ${\mathcal{D}}_{\text{neg}}$
\EndFor

\State \textbf{Adversarial Training Phase:}
\For{$i \gets 1$ to $N_{\text{step}}$}
    \State Sample prompts $x \sim \mathcal{Q}$
    \State Generate group $\mathcal{D}_{P} \gets \{y^{(g)}\}_{g=1}^{G} \sim \pi_{\theta}(\cdot \mid x)$
    \State Construct training sets based on answer correctness:
    \State \quad $\mathcal{D}_{\text{pos}} \gets \mathcal{D}_E \cup \{ y \in \mathcal{D}_P \mid \mathcal{O}(y) = \mathcal{O}(y^E) \}$
    \State \quad $\mathcal{D}_{\text{neg}} \gets \{ y \in \mathcal{D}_P \mid \mathcal{O}(y) \neq \mathcal{O}(y^E) \} \cup \{ \mathcal{C}(y^E) \mid y^E \in \mathcal{D}_E\}$
    \State \quad $\mathcal{B}_{\text{pos}} \gets \mathcal{B}_{\text{pos}} \cup \mathcal{D}_{\text{pos}}; \quad \mathcal{B}_{\text{neg}} \gets \mathcal{B}_{\text{neg}} \cup \mathcal{D}_{\text{neg}}$ \Comment{Update replay buffers}
    
    \State \textbf{Reward Model Update:}
    \For{$k \gets 1$ to $N_{\text{disc}}$}
        \State Sample mini-batches $\tilde{\mathcal{D}}_{\text{pos}} \sim \mathcal{B}_{\text{pos}}$ and $\tilde{\mathcal{D}}_{\text{neg}} \sim \mathcal{B}_{\text{neg}}$
        \State Update $D_{\phi}$ to minimise Eq.~\eqref{eq:discriminator_objective} using $\tilde{\mathcal{D}}_{\text{pos}}$ and $\tilde{\mathcal{D}}_{\text{neg}}$
    \EndFor
    
    \State \textbf{Policy Update:}
    \State Get dense rewards $r_{\phi}(y_t^{(g)})$ via backfilling logits (Eq.~\ref{eq:token_reward_def} and Eq.~\ref{eq:reward_clipping})
    \State Compute advantages $A^{(g)}_{t}$ via group standardisation (Eq.~\ref{eq:advantage_grpo})
    \State Optimise $\pi_{\theta}$ using GRPO loss on policy generations
\EndFor
\end{algorithmic}
\end{algorithm}

\newpage

\section{Implementation Details}\label{app:implementation-details}
We evaluate the proposed expert reasoning approach on \textsc{GSM8K}~\citep{cobbe_training_2021}, a benchmark for grade school math problems that provides final answers and human-written demonstrations. Moreover, to demonstrate the effectiveness of the proposed method in extracting a dense reasoning reward model, we performed our experiments on \textsc{MedReason}~\citep{wu2025medreasonelicitingfactualmedical}, and more specifically on the \textsc{MedQA}~\citep{jin2021disease} subset, consisting of questions from the US medical board exam, and \textsc{the MedMCQA}~\citep{pmlr-v174-pal22a} subset, comprising questions from the entrance exam from the Indian medical school curriculum. The dataset used in our experiments comprises approximately 7'000 questions for training and 1'000 for evaluation. In addition, \cite{wu2025medreasonelicitingfactualmedical} provides quality filtered medical reasoning traces constructed by strong language models (ChatGPT), which can be used for supervised fine-tuning or, in our case, adversarial inverse RL. The \textsc{MMLU-Pro} dataset \citep{wang2024mmlu} is an advanced version of the original Massive Multi-Task Language Understanding (MMLU) dataset, comprising questions that are particularly difficult to reason about across domains such as maths, biology, health, business, law, etc. Moreover, the answer options have been expanded from 4 options to 10 options to increase difficulty. This dataset provides additional insight into more challenging and open-domain reasoning tasks.
\\\\
Unless otherwise noted, we use open-weight, instruction-tuned models as base policies and train a learned reward function via adversarial inverse reinforcement learning. To obtain a dense signal, we implement the discriminator as a token classifier that shares the backbone with a language model and replaces the language modelling head with a single linear layer that outputs one logit per token. The code for all our experiments can be found in \href{https://github.com/fanconic/expert_reasoning}{https://github.com/fanconic/expert\_reasoning}.

All experiments are implemented in Python~\citep{python} with PyTorch~\citep{pytorch} and Hugging Face Transformers~\citep{wolf_huggingfaces_2020}. We accelerate training and evaluation with \textsc{Unsloth}~\citep{unsloth}. Unless stated otherwise, we use a starting learning rate of \(1\times10^{-5}\) for the reasoning discriminator and \(5\times10^{-6}\) for the generator. The reward reasoning model is warmed up for \(250 \) optimisation steps, and we train for \(400\) adversarial optimisation steps with a batch size of \(16\), generating \(G=8\) samples per prompt, accumulated over eight gradient steps (actual batch size + 128). During the adversarial update, we perform one policy-model update and three reasoning-reward model updates. To prevent data drift, we implemented a replay buffer that keeps track of the 50 most recent batches and samples a balanced combination of them when feeding the reward model. Both the discriminator and policy optimisers use a cosine-annealing learning rate schedule. The discriminator optimiser has a warm-up sequence of $250$ steps (same as the discriminator warm-up period), while the policy optimiser is warmed up for $50$ steps. We use a quantised ADAMW~\citep{loshchilov2018decoupled} optimiser.\label{app:training-hparams}
\\\\
\textbf{Data and preprocessing.}\label{app:data-preprocessing}
Prompts consist of the problem text with a short system instruction that requests step-by-step reasoning. Demonstrations are formatted as \texttt{<think> ... </think>} followed by \texttt{<answer> ... </answer>} format and provide the system instructions below. Tokenisation uses the native tokeniser of each backbone. For evaluation, we decode with temperature \(T=0.5\) and \texttt{top\_p} \(=0.95\). We validate every run after 100 steps, save the model with the best validation accuracy on correctness, for all baselines and models, and use these weights to evaluate on the test set.

\begin{tcolorbox}[mybox={System Prompt for \textsc{GSM8K} and \textsc{MMLU-Pro}}]
\footnotesize
\begin{verbatim}
A conversation between User and Assistant. The user asks a question,
and the  Assistant solves it. The assistant first thinks about the 
reasoning process in the mind and then provides the user with the 
answer. The reasoning process and answer are enclosed within 
<think> </think> and <answer> </answer> tags, respectively, i.e., 
<think> reasoning process here </think><answer> answer here </answer>
\end{verbatim}
\end{tcolorbox}

\begin{tcolorbox}[mybox={System Prompt for \textsc{MedReason}}]
\footnotesize
\begin{verbatim}
You are an expert medical AI. You must analyze clinical scenarios and adhere strictly 
to this output format:

1. Enclose your step-by-step reasoning within `<think>` and `</think>` tags. 
   Keep your reasoning strictly <500 words.
2. Immediately after, output your final conclusion within `<answer>` and `</answer>` 
   tags.    The content must be exactly in the format: `<LETTER>. <ANSWER_TEXT>`, where:
   - `<LETTER>` is one of A, B, C, D
   - `<ANSWER_TEXT>` is the exact option text corresponding to that letter

Example:
<think>
[Your step-by-step clinical reasoning]
</think>
<answer>
C. Hyperthyroidism
</answer>
"""
\end{verbatim}
\end{tcolorbox}

\textbf{Inference time scoring.}\label{app:inference-scoring}
At inference time, we draw \(N=16\) samples per prompt, compute the mean polynomially discounted ($\gamma=0.95$) reward over the reasoning tokens for each sample, and rerank by this score. We introduce this discount to reward tokens towards the end of a sequence, as they are closer to completion.
\\\\
\textbf{Perturbations.}\label{app:perturbations}
To improve robustness and reduce reliance on surface form, we introduce targeted perturbations during discriminator training for both expert and policy traces. For \textsc{GSM8K} we apply the following random operations: (i) flip arithmetic operator signs, (ii) corrupt numeric literals by small random offsets, and (iii) swap the final answer with an earlier intermediate number. Perturbed traces are labelled as non-expert. For \textsc{MedReason}, we use ChatGPT-5.2 to generate incorrect reasoning traces that lead to incorrect results, by providing in-context the correct reasoning trace and the correct answer and instructing it to corrupt it. We do the same thing for \textsc{MMLU-Pro}. We provide the prompt below for corruption:

\textbf{Compute.}\label{app:compute}
Experiments are conducted on 4 A100-class GPUs using mixed-precision training. We use gradient accumulation to match effective batch sizes across the backbones. All models operate in 4-bit mode, as provided by UNSLOTH~\citep{unsloth}, to improve training speed and memory efficiency.  All experiments are run on a single GPU instance to enable parallelisation.

\subsection{Policy and Reward Model}\label{app:policy-reward-model}
Policies are initialised from instruction-tuned checkpoints and trained with the learned reward signal. The following policy backbones are used:
\begin{itemize}
    \item \texttt{Llama3.1-8B-Instruct}
    \item \texttt{Llama3.2-3B-Instruct}
    \item \texttt{Qwen2.5-3B-Instruct}
    \item \texttt{Qwen2.5-7B-Instruct}
    \item \texttt{Qwen3-4B-Instruct-2507}
\end{itemize}
We instantiate the same reasoning reward models as the policy models. All models are turned into LoRA~\cite{hu2021loralowrankadaptationlarge} adapters ($r=256, \alpha=512$) and PEFT is used.

\begin{tcolorbox}[mybox={System Prompt for Corrupted Dataset}]
\footnotesize
\begin{verbatim}
"""
You are an expert educator creating adversarial training examples.
Below is a question, the correct reasoning, and the correct answer.

Your task:
1. Identify the incorrect options (distractors) provided in the question.
2. For EACH incorrect option, write a plausible reasoning trace that leads to 
   that specific wrong answer.
3. The reasoning should sound logical but must conclude with the wrong answer.
4. Match the formatting of the correct reasoning provided.

Format your output as sequential blocks:

<block>
<think>
[Reasoning for Distractor]
</think>
<answer>
[Distractor Label and Text]
</answer>
</block>

---
Question: 
{question}

Correct Reasoning (do not use):
{correct_reasoning}

Correct Answer (do not use):
{correct_answer}
"""
\end{verbatim}
\end{tcolorbox}

\subsection{Future Work}
Our findings suggest that learned reasoning reward models can effectively serve as both a training signal and an interpretable inference-time assistant. Addressing the stability-granularity trade-off is critical; this involves improving the optimisation dynamics at the discriminator–policy interface to prevent mode collapse in dense settings, potentially through alternative objectives such as the Wasserstein GAN~\citep{arjovsky_wasserstein_2017}. Moreover, the dense reward token-level interpretability opens the door to active test-time intervention, such as reward-guided decoding, early-exit mechanisms, and iterative self-revision strategies.

\subsection{Statement about the Use of Large Language Models}\label{llm-statement}
We utilised large language models to assist with drafting and editing the manuscript and to accelerate implementation by generating boilerplate code and providing debugging suggestions. LLMs were not involved in the design of the methods, the study design, or the interpretation of the results. All outputs were reviewed and verified by the authors, who take full responsibility for the content.

\newpage


\section{Additional Experiments}\label{app:additional-experiments}
\subsection{Ablation Reward Clipping}

\begin{greycustomblock}
\textbf{RQ-A1:} \textit{How sensitive is R-AIRL to reward clipping bounds once the stabilised adversarial training recipe is used?}
\end{greycustomblock}

\textbf{Experimental setup.} We ablate the clipping interval \(\beta=(\mathrm{LB},\mathrm{UB})\) for \texttt{Llama3.2-3B} with step-wise rewards on \textsc{GSM8K}, while keeping the rest of the optimisation recipe fixed (KL penalty, replay buffer, and three discriminator updates per policy update). We report the held-out Pass@1, the reranking gain (\(\Delta\)Rerank), AUROC, and ECE.

\begin{table}[h!]
\centering
{%
\begin{tabular}{l cccc}
\toprule
\textbf{$\beta$ (LB, UB)} & \textbf{Pass@1} & \textbf{$\Delta$ Rerank} & \textbf{AUROC $\uparrow$} & \textbf{ECE $\downarrow$} \\
\midrule
(0, 2) & 54 \tiny\textcolor{gray}{[52, 56]} & \textbf{+14} & 86 \tiny\textcolor{gray}{[85, 86]} & 21 \tiny\textcolor{gray}{[20, 22]} \\
(-1, 1) & 57 \tiny\textcolor{gray}{[55, 59]} & \textbf{+11} & 84 \tiny\textcolor{gray}{[83, 84]} & 15 \tiny\textcolor{gray}{[14, 15]} \\
(-3, 3) & 54 \tiny\textcolor{gray}{[52, 56]} & \textbf{+14} & 89 \tiny\textcolor{gray}{[88, 89]} & 13 \tiny\textcolor{gray}{[12, 13]} \\
(-5, 5) & 58 \tiny\textcolor{gray}{[56, 60]} & \textbf{+11} & 83 \tiny\textcolor{gray}{[82, 83]} & 20 \tiny\textcolor{gray}{[20, 21]} \\
(-10, 10) & 52 \tiny\textcolor{gray}{[51, 54]} & \textbf{+17} & 87 \tiny\textcolor{gray}{[87, 88]} & 18 \tiny\textcolor{gray}{[17, 18]} \\
($-\infty$, $\infty$) & 58 \tiny\textcolor{gray}{[56, 60]} & \textbf{+12} & 87 \tiny\textcolor{gray}{[87, 88]} & 12 \tiny\textcolor{gray}{[12, 13]} \\
\bottomrule
\end{tabular}%
}\vspace{1em}
\caption[Ablation on Reward Clipping ($\beta$)]{\textbf{Ablation on reward clipping ($\beta$)} for \texttt{Llama3.2-3B} (Step-wise) on GSM8K. All metrics are shown in full percentages (\%). Intervals indicate 95\% confidence intervals.}
\label{tab:beta_ablation}
\end{table}

\begin{customblockquote}
\textbf{Takeaway.} Under our stabilised adversarial recipe for R-AIRL, reward clipping is a secondary design choice rather than a primary driver of performance.
\end{customblockquote}

\subsection{Ablation Groups}

\begin{greycustomblock}
\textbf{RQ-A2:} \textit{How sensitive is performance to the number of sampled generations per prompt (\(g\)) during policy optimisation and reranking?}
\end{greycustomblock}

\textbf{Experimental setup.} We vary the group size \(g \in \{4,8,16\}\) for \texttt{Llama3.2-3B} with step-wise rewards on \textsc{GSM8K}. As above, we report Pass@1, \(\Delta\) Rerank, AUROC, and ECE to capture both downstream accuracy and reward quality.

\begin{table}[h!]
\centering
{%
\begin{tabular}{l cccc}
\toprule
\textbf{Gen. ($g$)} & \textbf{Pass@1} & \textbf{$\Delta$ Rerank} & \textbf{AUROC $\uparrow$} & \textbf{ECE $\downarrow$} \\
\midrule
4 & 53 \tiny\textcolor{gray}{[52, 55]} & \textbf{+18} & 89 \tiny\textcolor{gray}{[89, 90]} & 13 \tiny\textcolor{gray}{[13, 14]} \\
8 & 58 \tiny\textcolor{gray}{[56, 60]} & \textbf{+11} & 83 \tiny\textcolor{gray}{[82, 83]} & 20 \tiny\textcolor{gray}{[20, 21]} \\
16 & 57 \tiny\textcolor{gray}{[56, 59]} & \textbf{+13} & 86 \tiny\textcolor{gray}{[86, 87]} & 14 \tiny\textcolor{gray}{[13, 14]} \\
\bottomrule
\end{tabular}%
}\vspace{0.5em}
\caption[Ablation on Number of Generations ($G$)]{\textbf{Ablation on number of generations ($G$)} for \texttt{Llama3.2-3B} (Interval) on GSM8K. All metrics are in full percentages (\%). Intervals indicate 95\% confidence intervals.}
\end{table}

\begin{customblockquote}
\textbf{Takeaway.} The method is not brittle to the exact group size, suggesting that its gains are not tied to a narrowly tuned sampling configuration.
\end{customblockquote}

\newpage
\subsection{AIME}

\begin{greycustomblock}
\textbf{RQ-A3:} \textit{Do demonstration-only learned rewards remain competitive on harder olympiad-style math benchmarks?}
\end{greycustomblock}

\textbf{Experimental setup.} We evaluate pass@1 on \textsc{AIME 2024} and \textsc{AIME 2025} for the \texttt{Qwen3-4B}, comparing SFT with R-AIRL variants across reward granularities (sparse, interval, dense).

\begin{table}[h!]
\centering
\scriptsize
\begin{tabular}{l c c}
\toprule
\textbf{Method} & \textbf{\textsc{AIME 2024}} & \textbf{\textsc{AIME 2025}} \\
\midrule
\multicolumn{3}{l}{\textbf{\texttt{Qwen3-4B}}} \\
\hspace{1em}SFT & \textbf{3.45} & 3.12 \\
\hspace{1em}Ours (\textit{Sparse}) & \underline{3.43} & \underline{3.82} \\
\hspace{1em}Ours (\textit{Interval}) & 3.31 & \textbf{4.30} \\
\hspace{1em}Ours (\textit{Dense}) & 3.00 & 2.17 \\
\bottomrule
\end{tabular}\vspace{0.5em}
\caption[Reward-weighted Pass@1 Performance (\%)]{\textbf{Reward-weighted Pass@1 Performance (\%).} \textbf{Bold} indicates the best performance compared between SFT and our methods.}
\label{tab:p1_results_aime}
\end{table}

\vspace{-2em}

\begin{customblockquote}
\textbf{Takeaway.} On harder olympiad-style benchmarks, the main message is competitiveness rather than uniform dominance, with stronger backbones benefiting more from reward-based training.
\end{customblockquote}

\subsection{Ablation Corruptions}

\begin{greycustomblock}
\textbf{RQ-A4:} \textit{Do answer agreement and targeted corruptions reduce style shortcuts and improve reasoning-discriminative rewards?}
\end{greycustomblock}

\begin{figure}[h!]
    \begin{subfigure}[t]{0.32\textwidth}
        \vspace{0pt}
        \centering
        \includegraphics[width=\linewidth]{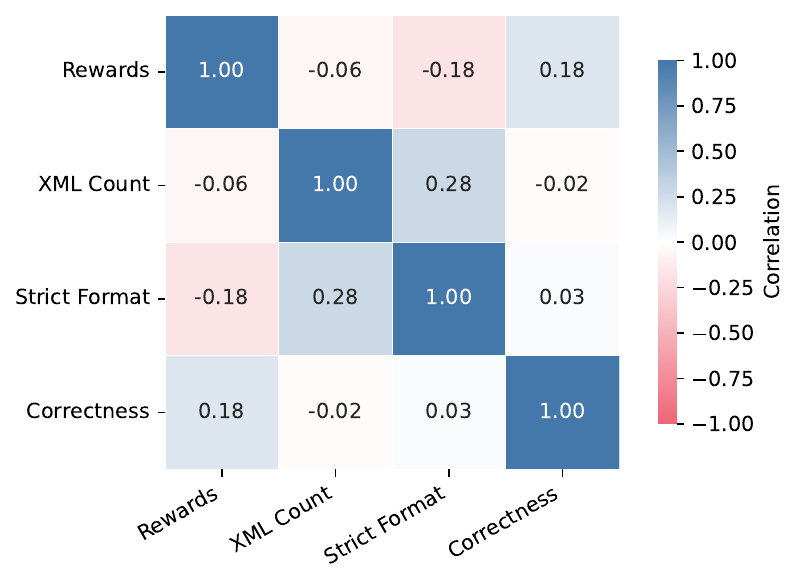}
        \caption{No Answer Agreement}
        \label{fig:qwen3b-corruption-1}
    \end{subfigure}
    \hfill
    \begin{subfigure}[t]{0.32\textwidth}
        \vspace{0pt}
        \centering
        \includegraphics[width=\linewidth]{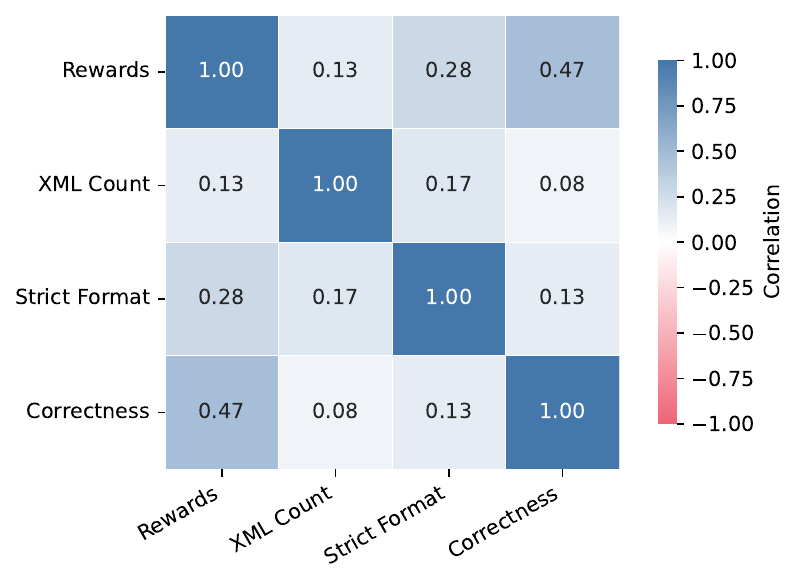}
        \caption{Answer Agreement only}
        \label{fig:qwen3b-corruption-2}
    \end{subfigure}
    \hfill
    \begin{subfigure}[t]{0.32\textwidth}
        \vspace{0pt}
        \centering
        \includegraphics[width=\linewidth]{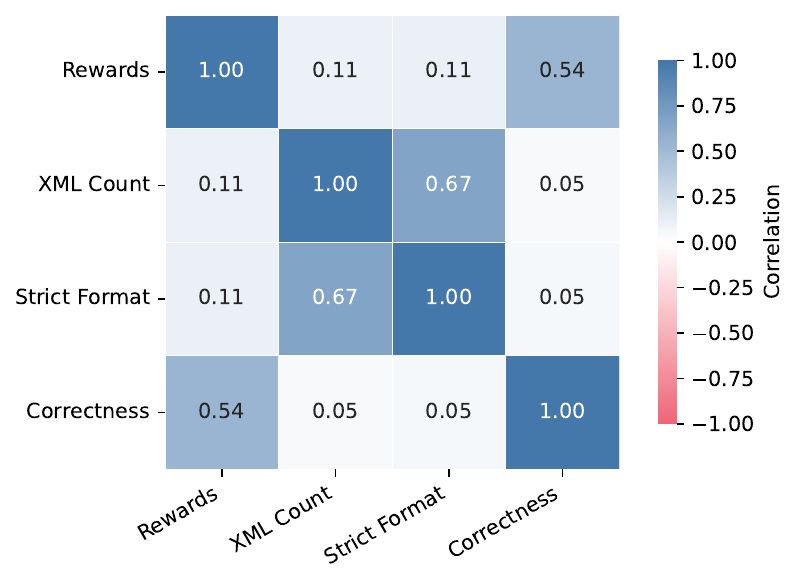}
        \caption{Answer Agr. + Full Corruption}
        \label{fig:qwen3b-corruption-3}
    \end{subfigure}

    \caption[Correlation of rewards with correctness and format]{\textbf{Correlation of rewards with correctness and format.} \texttt{Qwen2.5-3B} on \textsc{GSM8K} under increasing corruption (no answer agreement \(\rightarrow\) answer agreement \(\rightarrow\) answer agreement + full corruption). Correctness correlation increases (0.18 \(\rightarrow\) 0.47 \(\rightarrow\) 0.54), while format correlation is reduced in the fully corrupted setting (0.28 \(\rightarrow\) 0.11).}
\label{fig:qwen3b-corruption}
\end{figure}

\textbf{Experimental setup.} We run a progressive ablation on \texttt{Qwen2.5-3B} (interval) on \textsc{GSM8K}: no answer agreement, answer agreement as soft supervision only, then answer agreement plus number corruption, number swapping, or symbol corruption. We report Pass@1, \(\Delta\) Rerank, AUROC, and ECE (Table~\ref{tab:corruption_ablation}), and analyse reward correlation with correctness versus format cues (Figure~\ref{fig:qwen3b-corruption}).

\begin{table}[h]
\centering
\scriptsize
{%
\begin{tabular}{>{\raggedright\arraybackslash}p{5cm} cccc}
\toprule

\textbf{Corruption Type} & \textbf{Pass@1} & \textbf{$\Delta$ Rerank} & \textbf{AUROC $\uparrow$} & \textbf{ECE $\downarrow$} \\
\midrule
No Answer Agreement & 31 \tiny\textcolor{gray}{[29, 32]} & \textbf{+4} & 62 \tiny\textcolor{gray}{[61, 62]} & 8 \tiny\textcolor{gray}{[7, 8]} \\
Answer Agreement & 56 \tiny\textcolor{gray}{[55, 58]} & \textbf{+8} & 77 \tiny\textcolor{gray}{[76, 78]} & 22 \tiny\textcolor{gray}{[22, 23]} \\
 + Number corruption & 58 \tiny\textcolor{gray}{[56, 59]} & \textbf{+10} & 78 \tiny\textcolor{gray}{[77, 78]} & 34 \tiny\textcolor{gray}{[34, 35]} \\
 + Number swapping & 56 \tiny\textcolor{gray}{[54, 58]} & \textbf{+15} & 84 \tiny\textcolor{gray}{[83, 84]} & 27 \tiny\textcolor{gray}{[27, 28]} \\
 + Symbol corruption & 56 \tiny\textcolor{gray}{[55, 58]} & \textbf{+18} & 85 \tiny\textcolor{gray}{[84, 85]} & 31 \tiny\textcolor{gray}{[31, 32]} \\
\bottomrule
\end{tabular}%
}\vspace{0.5em}
\caption[Ablation on Reasoning Corruption]{\textbf{Ablation on Reasoning Corruption} for \texttt{Qwen2.5-3B} (interval) on GSM8K. All metrics are in full percentages (\%). Intervals indicate 95\% confidence intervals.}
\label{tab:corruption_ablation}
\end{table}
\vspace{-1em}

\begin{customblockquote}
\textbf{Takeaway.} Answer agreement is key to learning rewards that track reasoning quality rather than superficial style cues. Corruptions can make this further robust.
\end{customblockquote}

\newpage


\section{Additional Results}\label{app:additional-results}
\subsection{Complete Performance Table}

\begin{table*}[h!]
\scriptsize
\renewcommand{\arraystretch}{1.18}
\setlength{\tabcolsep}{3.8pt}
\centering
\resizebox{\textwidth}{!}{%
\begin{tabular}{l l c c c c c c c c c}
\toprule
\textbf{Method} & \textbf{Granularity} & \multicolumn{3}{c}{\textbf{\textsc{GSM8K}}} & \multicolumn{3}{c}{\textbf{\textsc{MMLU-Pro}}} & \multicolumn{3}{c}{\textbf{\textsc{MedReason}}} \\
\cmidrule(lr){3-5}\cmidrule(lr){6-8}\cmidrule(lr){9-11}
& & \textbf{\tiny\texttt{Qwen2.5-7B}} & \textbf{\tiny\texttt{Llama3.1-8B}} & \textbf{\tiny\texttt{Qwen3-4B}} & \textbf{\tiny\texttt{Qwen2.5-7B}} & \textbf{\tiny\texttt{Llama3.1-8B}} & \textbf{\tiny\texttt{Qwen3-4B}} & \textbf{\tiny\texttt{Qwen2.5-7B}} & \textbf{\tiny\texttt{Llama3.1-8B}} & \textbf{\tiny\texttt{Qwen3-4B}} \\
\midrule
\rowcolor{black!8}
\multicolumn{11}{l}{\textbf{External verifier}} \\
Outcome Reward & -- & 89.0 {\scriptsize\color{gray}$\pm$ 1.5} & 83.3 {\scriptsize\color{gray}$\pm$ 1.6} & 91.6 {\scriptsize\color{gray}$\pm$ 1.3} & 53.5 {\scriptsize\color{gray}$\pm$ 2.0} & 48.4 {\scriptsize\color{gray}$\pm$ 2.0} & 57.1 {\scriptsize\color{gray}$\pm$ 2.2} & 69.2 {\scriptsize\color{gray}$\pm$ 2.4} & 82.8 {\scriptsize\color{gray}$\pm$ 2.1} & 71.8 {\scriptsize\color{gray}$\pm$ 2.5} \\
\specialrule{0.9pt}{0.15em}{0.15em}
\rowcolor{black!8}
\multicolumn{11}{l}{\textbf{Demonstrations only}} \\
SFT & -- & 70.1 {\scriptsize\color{gray}$\pm$ 1.6} & 66.6 {\scriptsize\color{gray}$\pm$ 2.1} & 76.6 {\scriptsize\color{gray}$\pm$ 1.7} & 48.1 {\scriptsize\color{gray}$\pm$ 1.9} & \textbf{47.2} {\scriptsize\color{gray}$\pm$ 1.9} & 53.9 {\scriptsize\color{gray}$\pm$ 2.0} & 65.4 {\scriptsize\color{gray}$\pm$ 2.3} & \textbf{79.5} {\scriptsize\color{gray}$\pm$ 2.0} & 67.4 {\scriptsize\color{gray}$\pm$ 2.3} \\
\cdashline{1-11}[0.5pt/1.8pt]
\multirow{3}{*}{R-AIRL} & \textit{Sparse} & \textbf{85.8} {\scriptsize\color{gray}$\pm$ 1.5} & \textbf{80.6} {\scriptsize\color{gray}$\pm$ 1.6} & \textbf{90.4} {\scriptsize\color{gray}$\pm$ 1.5} & \underline{48.5} {\scriptsize\color{gray}$\pm$ 2.2} & \underline{43.3} {\scriptsize\color{gray}$\pm$ 1.9} & \textbf{55.6} {\scriptsize\color{gray}$\pm$ 2.2} & \underline{67.1} {\scriptsize\color{gray}$\pm$ 2.5} & \underline{77.6} {\scriptsize\color{gray}$\pm$ 2.0} & 68.9 {\scriptsize\color{gray}$\pm$ 2.6} \\
& \textit{Interval} & \underline{78.8} {\scriptsize\color{gray}$\pm$ 1.6} & \underline{67.6} {\scriptsize\color{gray}$\pm$ 1.9} & 87.8 {\scriptsize\color{gray}$\pm$ 1.6} & \textbf{50.6} {\scriptsize\color{gray}$\pm$ 1.9} & 36.6 {\scriptsize\color{gray}$\pm$ 1.8} & 53.5 {\scriptsize\color{gray}$\pm$ 2.4} & 66.3 {\scriptsize\color{gray}$\pm$ 2.5} & 76.3 {\scriptsize\color{gray}$\pm$ 2.1} & \underline{69.1} {\scriptsize\color{gray}$\pm$ 2.6} \\
& \textit{Dense} & 38.4 {\scriptsize\color{gray}$\pm$ 2.0} & 64.6 {\scriptsize\color{gray}$\pm$ 2.1} & \underline{89.6} {\scriptsize\color{gray}$\pm$ 1.5} & 43.8 {\scriptsize\color{gray}$\pm$ 2.0} & 37.9 {\scriptsize\color{gray}$\pm$ 1.8} & \underline{55.1} {\scriptsize\color{gray}$\pm$ 2.3} & \textbf{67.6} {\scriptsize\color{gray}$\pm$ 2.4} & 76.7 {\scriptsize\color{gray}$\pm$ 2.0} & \textbf{69.3} {\scriptsize\color{gray}$\pm$ 2.6} \\
\bottomrule
\end{tabular}}
\vspace{0.6em}
\caption{\textbf{Held-out pass@1 by supervision regime.} Methods are separated into \textit{External verifier} (Outcome Reward) and \textit{Demonstrations only} (SFT and R-AIRL variants). \textbf{Bold} indicates the best performance and \underline{underlined} the second best among demonstration-only methods. Values are reported as mean $\pm$ half-width of the 95\% confidence interval bootstrapped over the test set.}
\label{tab:p1_results_supervision_split}
\end{table*}
\subsection{Training Behaviour}\label{app:training-behaviour}

\modeltrainingfigure{qwen7b}{\texttt{Qwen2.5-7B}}{}
\modeltrainingfigure{llama8b}{\texttt{Llama3.1-8B}}{}
\modeltrainingfigure{qwen4b}{\texttt{Qwen3-4B}}{}

\clearpage
\subsection{Reranking Calibration}

\begin{table*}[h!]
\centering\scriptsize
\resizebox{\textwidth}{!}{%
\begin{tabular}{ll cc cc cc}
\toprule
& & \multicolumn{2}{c}{\textbf{\textsc{GSM8K}}} & \multicolumn{2}{c}{\textbf{\textsc{MedReason}}} & \multicolumn{2}{c}{\textbf{\textsc{MMLU-Pro}}} \\
\cmidrule(lr){3-4} \cmidrule(lr){5-6} \cmidrule(lr){7-8}
\textbf{Backbone} & \textbf{Method}  & AUROC (\%) $\uparrow$ & ECE (\%) $\downarrow$  & AUROC (\%) $\uparrow$ & ECE (\%) $\downarrow$  & AUROC (\%) $\uparrow$ & ECE (\%) $\downarrow$  \\
\midrule
\multirow{3}{*}{\textbf{\texttt{Qwen2.5-7B}}} & \textit{Sparse} & 69.9 \tiny\textcolor{gray}{[68.8, 71.0]} & 8.3 \tiny\textcolor{gray}{[7.9, 8.8]} & 76.8 \tiny\textcolor{gray}{[76.0, 77.6]} & 15.8 \tiny\textcolor{gray}{[15.1, 16.4]} & 70.9 \tiny\textcolor{gray}{[70.3, 71.5]} & 17.2 \tiny\textcolor{gray}{[16.6, 17.7]} \\
 & \textit{Interval} & 70.1 \tiny\textcolor{gray}{[69.1, 71.0]} & 7.5 \tiny\textcolor{gray}{[7.0, 8.0]} & 71.0 \tiny\textcolor{gray}{[70.2, 71.8]} & 9.7 \tiny\textcolor{gray}{[9.0, 10.4]} & 68.0 \tiny\textcolor{gray}{[67.3, 68.7]} & 35.6 \tiny\textcolor{gray}{[35.0, 36.2]} \\
 & \textit{Dense} & 71.4 \tiny\textcolor{gray}{[70.7, 72.1]} & 45.2 \tiny\textcolor{gray}{[44.6, 45.9]} & 70.4 \tiny\textcolor{gray}{[69.5, 71.2]} & 15.5 \tiny\textcolor{gray}{[14.8, 16.2]} & 60.2 \tiny\textcolor{gray}{[59.5, 60.9]} & 41.3 \tiny\textcolor{gray}{[40.7, 41.9]} \\
\midrule
\multirow{3}{*}{\textbf{\texttt{Llama3.1-8B}}} & \textit{Sparse} & 75.0 \tiny\textcolor{gray}{[74.1, 75.9]} & 2.4 \tiny\textcolor{gray}{[1.9, 2.9]} & 82.2 \tiny\textcolor{gray}{[81.5, 83.0]} & 14.1 \tiny\textcolor{gray}{[13.5, 14.6]} & 68.7 \tiny\textcolor{gray}{[68.0, 69.3]} & 9.8 \tiny\textcolor{gray}{[9.2, 10.3]} \\
 & \textit{Interval} & 81.8 \tiny\textcolor{gray}{[81.2, 82.5]} & 8.7 \tiny\textcolor{gray}{[8.2, 9.2]} & 85.1 \tiny\textcolor{gray}{[84.5, 85.7]} & 6.5 \tiny\textcolor{gray}{[6.0, 7.1]} & 70.8 \tiny\textcolor{gray}{[70.1, 71.5]} & 39.7 \tiny\textcolor{gray}{[39.2, 40.2]} \\
 & \textit{Dense} & 75.9 \tiny\textcolor{gray}{[75.1, 76.6]} & 27.2 \tiny\textcolor{gray}{[26.6, 27.8]} & 86.0 \tiny\textcolor{gray}{[85.4, 86.6]} & 10.9 \tiny\textcolor{gray}{[10.3, 11.5]} & 60.2 \tiny\textcolor{gray}{[59.4, 60.9]} & 50.9 \tiny\textcolor{gray}{[50.4, 51.5]} \\
\midrule
\multirow{3}{*}{\textbf{\texttt{Qwen3-4B}}} & \textit{Sparse} & 84.1 \tiny\textcolor{gray}{[83.0, 85.1]} & 6.8 \tiny\textcolor{gray}{[6.5, 7.1]} & 73.9 \tiny\textcolor{gray}{[73.1, 74.7]} & 11.0 \tiny\textcolor{gray}{[10.4, 11.7]} & 86.8 \tiny\textcolor{gray}{[86.3, 87.2]} & 9.9 \tiny\textcolor{gray}{[9.4, 10.3]} \\
 & \textit{Interval} & 78.3 \tiny\textcolor{gray}{[77.1, 79.5]} & 13.5 \tiny\textcolor{gray}{[13.2, 13.9]} & 74.8 \tiny\textcolor{gray}{[73.9, 75.6]} & 3.2 \tiny\textcolor{gray}{[2.6, 3.9]} & 84.2 \tiny\textcolor{gray}{[83.7, 84.7]} & 21.5 \tiny\textcolor{gray}{[21.0, 22.0]} \\
 & \textit{Dense} & 78.3 \tiny\textcolor{gray}{[77.2, 79.4]} & 7.3 \tiny\textcolor{gray}{[7.0, 7.7]} & 73.3 \tiny\textcolor{gray}{[72.5, 74.1]} & 8.1 \tiny\textcolor{gray}{[7.4, 8.8]} & 75.0 \tiny\textcolor{gray}{[74.4, 75.7]} & 18.6 \tiny\textcolor{gray}{[18.0, 19.2]} \\
\bottomrule
\end{tabular}}
\caption{\textbf{Critic Calibration Metrics.} All values are reported as percentages (\%). AUROC indicates ranking ability; ECE measures calibration error (lower is better).}\label{tab:calibration}
\end{table*}

\subsection{Reranking Performance Ablation}

\begin{table}[h!]
\centering
\scriptsize
\resizebox{\textwidth}{!}{%
\begin{tabular}{ll ccccc ccccc}
\toprule
& & \multicolumn{5}{c}{\textbf{\textsc{GSM8K}}} & \multicolumn{5}{c}{\textbf{\textsc{MMLU-Pro}}} \\
\cmidrule(lr){3-7} \cmidrule(lr){8-12}
\textbf{Backbone} & \textbf{Method} & Random & $\Delta$ Logp. & $\Delta$ Maj. & $\Delta$ Rew. & $\Delta$ W.Maj. & Random & $\Delta$ Logp. & $\Delta$ Maj. & $\Delta$ Rew. & $\Delta$ W.Maj. \\
\midrule
\multirow{4}{*}{\textbf{\texttt{Qwen2.5-4B}}} & \textit{Sparse} & 76.6\% & \textcolor{purple}{($\downarrow$ -0.5)} & (0.0) & \textbf{\textcolor{insightteal}{($\uparrow$ +3.2)}} & \textbf{\textcolor{insightteal}{($\uparrow$ +3.2)}} & 53.9\% & \textcolor{insightteal}{($\uparrow$ +2.7)} & \textcolor{insightteal}{($\uparrow$ +0.5)} & \textbf{\textcolor{insightteal}{($\uparrow$ +4.5)}} & \textbf{\textcolor{insightteal}{($\uparrow$ +4.5)}} \\
 & \textit{Interval} & 76.6\% & \textcolor{purple}{($\downarrow$ -0.5)} & (0.0) & \textbf{\textcolor{insightteal}{($\uparrow$ +4.0)}} & \textbf{\textcolor{insightteal}{($\uparrow$ +4.0)}} & 53.9\% & \textbf{\textcolor{insightteal}{($\uparrow$ +2.7)}} & \textcolor{insightteal}{($\uparrow$ +0.5)} & \textcolor{insightteal}{($\uparrow$ +0.9)} & \textcolor{insightteal}{($\uparrow$ +0.9)} \\
 & \textit{Dense} & 76.6\% & \textcolor{purple}{($\downarrow$ -0.5)} & (0.0) & \textbf{\textcolor{insightteal}{($\uparrow$ +3.6)}} & \textbf{\textcolor{insightteal}{($\uparrow$ +3.6)}} & 53.9\% & \textbf{\textcolor{insightteal}{($\uparrow$ +2.7)}} & \textcolor{insightteal}{($\uparrow$ +0.5)} & \textcolor{insightteal}{($\uparrow$ +1.9)} & \textcolor{insightteal}{($\uparrow$ +1.9)} \\
\midrule
\multirow{4}{*}{\textbf{\texttt{Qwen2.5-7B}}} & \textit{Sparse} & 70.1\% & \textcolor{insightteal}{($\uparrow$ +0.4)} & \textcolor{insightteal}{($\uparrow$ +0.3)} & \textbf{\textcolor{insightteal}{($\uparrow$ +6.0)}} & \textbf{\textcolor{insightteal}{($\uparrow$ +6.0)}} & 48.1\% & \textcolor{insightteal}{($\uparrow$ +0.9)} & \textcolor{purple}{($\downarrow$ -0.6)} & \textbf{\textcolor{insightteal}{($\uparrow$ +3.5)}} & \textbf{\textcolor{insightteal}{($\uparrow$ +3.5)}} \\
 & \textit{Interval} & 70.1\% & \textcolor{insightteal}{($\uparrow$ +0.4)} & \textcolor{insightteal}{($\uparrow$ +0.3)} & \textbf{\textcolor{insightteal}{($\uparrow$ +5.4)}} & \textbf{\textcolor{insightteal}{($\uparrow$ +5.4)}} & 48.1\% & \textcolor{insightteal}{($\uparrow$ +0.9)} & \textcolor{purple}{($\downarrow$ -0.6)} & \textbf{\textcolor{insightteal}{($\uparrow$ +1.1)}} & \textbf{\textcolor{insightteal}{($\uparrow$ +1.1)}} \\
 & \textit{Dense} & 70.1\% & \textcolor{insightteal}{($\uparrow$ +0.4)} & \textcolor{insightteal}{($\uparrow$ +0.3)} & \textbf{\textcolor{insightteal}{($\uparrow$ +6.4)}} & \textbf{\textcolor{insightteal}{($\uparrow$ +6.4)}} & 48.1\% & \textbf{\textcolor{insightteal}{($\uparrow$ +0.9)}} & \textcolor{purple}{($\downarrow$ -0.6)} & \textcolor{purple}{($\downarrow$ -0.1)} & \textcolor{purple}{($\downarrow$ -0.1)} \\
\midrule
\multirow{4}{*}{\textbf{\texttt{Llama3.1-8B}}} & \textit{Sparse} & 66.6\% & \textcolor{insightteal}{($\uparrow$ +1.4)} & \textcolor{insightteal}{($\uparrow$ +0.2)} & \textbf{\textcolor{insightteal}{($\uparrow$ +2.3)}} & \textbf{\textcolor{insightteal}{($\uparrow$ +2.3)}} & 47.2\% & \textcolor{insightteal}{($\uparrow$ +1.7)} & \textcolor{insightteal}{($\uparrow$ +0.6)} & \textbf{\textcolor{insightteal}{($\uparrow$ +2.8)}} & \textbf{\textcolor{insightteal}{($\uparrow$ +2.8)}} \\
 & \textit{Interval} & 66.6\% & \textcolor{insightteal}{($\uparrow$ +1.4)} & \textcolor{insightteal}{($\uparrow$ +0.2)} & \textbf{\textcolor{insightteal}{($\uparrow$ +3.1)}} & \textbf{\textcolor{insightteal}{($\uparrow$ +3.1)}} & 47.2\% & \textcolor{insightteal}{($\uparrow$ +1.7)} & \textcolor{insightteal}{($\uparrow$ +0.6)} & \textbf{\textcolor{insightteal}{($\uparrow$ +3.7)}} & \textbf{\textcolor{insightteal}{($\uparrow$ +3.7)}} \\
 & \textit{Dense} & 66.6\% & \textcolor{insightteal}{($\uparrow$ +1.4)} & \textcolor{insightteal}{($\uparrow$ +0.2)} & \textbf{\textcolor{insightteal}{($\uparrow$ +3.0)}} & \textbf{\textcolor{insightteal}{($\uparrow$ +3.0)}} & 47.2\% & \textcolor{insightteal}{($\uparrow$ +1.7)} & \textcolor{insightteal}{($\uparrow$ +0.6)} & \textbf{\textcolor{insightteal}{($\uparrow$ +2.0)}} & \textbf{\textcolor{insightteal}{($\uparrow$ +2.0)}} \\
\bottomrule
\end{tabular}%
}\vspace{0.5em}
\caption{\textbf{Best-of-2 Reranking Performance \& Baselines (\%).} Random is SFT pass@1. Deltas are percentage-point changes for each reranker. \textcolor{insightteal}{Blue arrows (\,$\uparrow$\,)} indicate gains and \textcolor{purple}{purple arrows (\,$\downarrow$\,)} indicate drops.}
\label{tab:reranking_baselines_full_N2}
\end{table}

\begin{table}[h!]
\centering
\scriptsize
\resizebox{\textwidth}{!}{%
\begin{tabular}{ll ccccc ccccc}
\toprule
& & \multicolumn{5}{c}{\textbf{\textsc{GSM8K}}} & \multicolumn{5}{c}{\textbf{\textsc{MMLU-Pro}}} \\
\cmidrule(lr){3-7} \cmidrule(lr){8-12}
\textbf{Backbone} & \textbf{Method} & Random & $\Delta$ Logp. & $\Delta$ Maj. & $\Delta$ Rew. & $\Delta$ W.Maj. & Random & $\Delta$ Logp. & $\Delta$ Maj. & $\Delta$ Rew. & $\Delta$ W.Maj. \\
\midrule
\multirow{4}{*}{\textbf{\texttt{Qwen2.5-4B}}} & \textit{Sparse} & 76.6\% & \textcolor{purple}{($\downarrow$ -1.0)} & \textcolor{insightteal}{($\uparrow$ +3.4)} & \textcolor{insightteal}{($\uparrow$ +5.4)} & \textbf{\textcolor{insightteal}{($\uparrow$ +6.5)}} & 53.9\% & \textcolor{insightteal}{($\uparrow$ +3.7)} & \textcolor{insightteal}{($\uparrow$ +3.5)} & \textbf{\textcolor{insightteal}{($\uparrow$ +6.5)}} & \textcolor{insightteal}{($\uparrow$ +6.4)} \\
 & \textit{Interval} & 76.6\% & \textcolor{purple}{($\downarrow$ -1.0)} & \textcolor{insightteal}{($\uparrow$ +3.4)} & \textcolor{insightteal}{($\uparrow$ +5.2)} & \textbf{\textcolor{insightteal}{($\uparrow$ +6.3)}} & 53.9\% & \textbf{\textcolor{insightteal}{($\uparrow$ +3.7)}} & \textcolor{insightteal}{($\uparrow$ +3.5)} & \textcolor{insightteal}{($\uparrow$ +2.1)} & \textcolor{insightteal}{($\uparrow$ +3.3)} \\
 & \textit{Dense} & 76.6\% & \textcolor{purple}{($\downarrow$ -1.0)} & \textcolor{insightteal}{($\uparrow$ +3.4)} & \textcolor{insightteal}{($\uparrow$ +5.2)} & \textbf{\textcolor{insightteal}{($\uparrow$ +6.1)}} & 53.9\% & \textcolor{insightteal}{($\uparrow$ +3.7)} & \textcolor{insightteal}{($\uparrow$ +3.5)} & \textcolor{insightteal}{($\uparrow$ +3.7)} & \textbf{\textcolor{insightteal}{($\uparrow$ +4.3)}} \\
\midrule
\multirow{4}{*}{\textbf{\texttt{Qwen2.5-7B}}} & \textit{Sparse} & 70.1\% & \textcolor{insightteal}{($\uparrow$ +1.7)} & \textcolor{insightteal}{($\uparrow$ +7.5)} & \textcolor{insightteal}{($\uparrow$ +8.1)} & \textbf{\textcolor{insightteal}{($\uparrow$ +9.2)}} & 48.1\% & \textcolor{insightteal}{($\uparrow$ +1.9)} & \textcolor{insightteal}{($\uparrow$ +2.1)} & \textcolor{insightteal}{($\uparrow$ +4.3)} & \textbf{\textcolor{insightteal}{($\uparrow$ +4.5)}} \\
 & \textit{Interval} & 70.1\% & \textcolor{insightteal}{($\uparrow$ +1.7)} & \textcolor{insightteal}{($\uparrow$ +7.5)} & \textcolor{insightteal}{($\uparrow$ +6.4)} & \textbf{\textcolor{insightteal}{($\uparrow$ +8.5)}} & 48.1\% & \textcolor{insightteal}{($\uparrow$ +1.9)} & \textcolor{insightteal}{($\uparrow$ +2.1)} & \textcolor{insightteal}{($\uparrow$ +2.4)} & \textbf{\textcolor{insightteal}{($\uparrow$ +3.4)}} \\
 & \textit{Dense} & 70.1\% & \textcolor{insightteal}{($\uparrow$ +1.7)} & \textcolor{insightteal}{($\uparrow$ +7.5)} & \textcolor{insightteal}{($\uparrow$ +9.4)} & \textbf{\textcolor{insightteal}{($\uparrow$ +10.1)}} & 48.1\% & \textcolor{insightteal}{($\uparrow$ +1.9)} & \textcolor{insightteal}{($\uparrow$ +2.1)} & \textcolor{insightteal}{($\uparrow$ +0.2)} & \textbf{\textcolor{insightteal}{($\uparrow$ +2.5)}} \\
\midrule
\multirow{4}{*}{\textbf{\texttt{Llama3.1-8B}}} & \textit{Sparse} & 66.6\% & \textcolor{insightteal}{($\uparrow$ +2.3)} & \textcolor{insightteal}{($\uparrow$ +1.8)} & \textcolor{insightteal}{($\uparrow$ +2.2)} & \textbf{\textcolor{insightteal}{($\uparrow$ +3.4)}} & 47.2\% & \textcolor{insightteal}{($\uparrow$ +2.0)} & \textbf{\textcolor{insightteal}{($\uparrow$ +4.0)}} & \textcolor{insightteal}{($\uparrow$ +2.1)} & \textcolor{insightteal}{($\uparrow$ +3.2)} \\
 & \textit{Interval} & 66.6\% & \textcolor{insightteal}{($\uparrow$ +2.3)} & \textcolor{insightteal}{($\uparrow$ +1.8)} & \textcolor{insightteal}{($\uparrow$ +3.9)} & \textbf{\textcolor{insightteal}{($\uparrow$ +4.4)}} & 47.2\% & \textcolor{insightteal}{($\uparrow$ +2.0)} & \textcolor{insightteal}{($\uparrow$ +4.0)} & \textcolor{insightteal}{($\uparrow$ +4.0)} & \textbf{\textcolor{insightteal}{($\uparrow$ +4.8)}} \\
 & \textit{Dense} & 66.6\% & \textcolor{insightteal}{($\uparrow$ +2.3)} & \textcolor{insightteal}{($\uparrow$ +1.8)} & \textbf{\textcolor{insightteal}{($\uparrow$ +3.7)}} & \textcolor{insightteal}{($\uparrow$ +3.6)} & 47.2\% & \textcolor{insightteal}{($\uparrow$ +2.0)} & \textcolor{insightteal}{($\uparrow$ +4.0)} & \textcolor{insightteal}{($\uparrow$ +2.8)} & \textbf{\textcolor{insightteal}{($\uparrow$ +4.4)}} \\
\bottomrule
\end{tabular}%
}\vspace{0.5em}
\caption{\textbf{Best-of-3 Reranking Performance \& Baselines (\%).} Random is SFT pass@1. Deltas are percentage-point changes for each reranker. \textcolor{insightteal}{Blue arrows (\,$\uparrow$\,)} indicate gains and \textcolor{purple}{purple arrows (\,$\downarrow$\,)} indicate drops.}
\label{tab:reranking_baselines_full_N3}
\end{table}

\begin{table}[h!]
\centering
\scriptsize
\resizebox{\textwidth}{!}{%
\begin{tabular}{ll ccccc ccccc}
\toprule
& & \multicolumn{5}{c}{\textbf{\textsc{GSM8K}}} & \multicolumn{5}{c}{\textbf{\textsc{MMLU-Pro}}} \\
\cmidrule(lr){3-7} \cmidrule(lr){8-12}
\textbf{Backbone} & \textbf{Method} & Random & $\Delta$ Logp. & $\Delta$ Maj. & $\Delta$ Rew. & $\Delta$ W.Maj. & Random & $\Delta$ Logp. & $\Delta$ Maj. & $\Delta$ Rew. & $\Delta$ W.Maj. \\
\midrule
\multirow{4}{*}{\textbf{\texttt{Qwen2.5-4B}}} & \textit{Sparse} & 76.6\% & \textcolor{purple}{($\downarrow$ -0.9)} & \textcolor{insightteal}{($\uparrow$ +5.5)} & \textcolor{insightteal}{($\uparrow$ +6.8)} & \textbf{\textcolor{insightteal}{($\uparrow$ +7.9)}} & 53.9\% & \textcolor{insightteal}{($\uparrow$ +2.8)} & \textcolor{insightteal}{($\uparrow$ +5.7)} & \textcolor{insightteal}{($\uparrow$ +7.5)} & \textbf{\textcolor{insightteal}{($\uparrow$ +8.1)}} \\
 & \textit{Interval} & 76.6\% & \textcolor{purple}{($\downarrow$ -0.9)} & \textcolor{insightteal}{($\uparrow$ +5.5)} & \textcolor{insightteal}{($\uparrow$ +6.7)} & \textbf{\textcolor{insightteal}{($\uparrow$ +7.9)}} & 53.9\% & \textcolor{insightteal}{($\uparrow$ +2.8)} & \textbf{\textcolor{insightteal}{($\uparrow$ +5.7)}} & \textcolor{insightteal}{($\uparrow$ +2.1)} & \textcolor{insightteal}{($\uparrow$ +4.8)} \\
 & \textit{Dense} & 76.6\% & \textcolor{purple}{($\downarrow$ -0.9)} & \textcolor{insightteal}{($\uparrow$ +5.5)} & \textcolor{insightteal}{($\uparrow$ +6.4)} & \textbf{\textcolor{insightteal}{($\uparrow$ +8.1)}} & 53.9\% & \textcolor{insightteal}{($\uparrow$ +2.8)} & \textbf{\textcolor{insightteal}{($\uparrow$ +5.7)}} & \textcolor{insightteal}{($\uparrow$ +3.5)} & \textcolor{insightteal}{($\uparrow$ +5.2)} \\
\midrule
\multirow{4}{*}{\textbf{\texttt{Qwen2.5-7B}}} & \textit{Sparse} & 70.1\% & \textcolor{insightteal}{($\uparrow$ +2.5)} & \textcolor{insightteal}{($\uparrow$ +10.9)} & \textcolor{insightteal}{($\uparrow$ +9.9)} & \textbf{\textcolor{insightteal}{($\uparrow$ +12.3)}} & 48.1\% & \textcolor{insightteal}{($\uparrow$ +3.5)} & \textcolor{insightteal}{($\uparrow$ +4.9)} & \textcolor{insightteal}{($\uparrow$ +5.9)} & \textbf{\textcolor{insightteal}{($\uparrow$ +6.5)}} \\
 & \textit{Interval} & 70.1\% & \textcolor{insightteal}{($\uparrow$ +2.5)} & \textcolor{insightteal}{($\uparrow$ +10.9)} & \textcolor{insightteal}{($\uparrow$ +6.8)} & \textbf{\textcolor{insightteal}{($\uparrow$ +11.7)}} & 48.1\% & \textcolor{insightteal}{($\uparrow$ +3.5)} & \textbf{\textcolor{insightteal}{($\uparrow$ +4.9)}} & \textcolor{insightteal}{($\uparrow$ +2.1)} & \textcolor{insightteal}{($\uparrow$ +4.8)} \\
 & \textit{Dense} & 70.1\% & \textcolor{insightteal}{($\uparrow$ +2.5)} & \textcolor{insightteal}{($\uparrow$ +10.9)} & \textcolor{insightteal}{($\uparrow$ +10.2)} & \textbf{\textcolor{insightteal}{($\uparrow$ +12.4)}} & 48.1\% & \textcolor{insightteal}{($\uparrow$ +3.5)} & \textcolor{insightteal}{($\uparrow$ +4.9)} & \textcolor{insightteal}{($\uparrow$ +1.0)} & \textbf{\textcolor{insightteal}{($\uparrow$ +5.3)}} \\
\midrule
\multirow{4}{*}{\textbf{\texttt{Llama3.1-8B}}} & \textit{Sparse} & 66.6\% & \textcolor{insightteal}{($\uparrow$ +2.5)} & \textcolor{insightteal}{($\uparrow$ +3.0)} & \textcolor{insightteal}{($\uparrow$ +2.2)} & \textbf{\textcolor{insightteal}{($\uparrow$ +3.7)}} & 47.2\% & \textcolor{insightteal}{($\uparrow$ +3.0)} & \textbf{\textcolor{insightteal}{($\uparrow$ +6.0)}} & \textcolor{insightteal}{($\uparrow$ +2.0)} & \textcolor{insightteal}{($\uparrow$ +5.9)} \\
 & \textit{Interval} & 66.6\% & \textcolor{insightteal}{($\uparrow$ +2.5)} & \textcolor{insightteal}{($\uparrow$ +3.0)} & \textcolor{insightteal}{($\uparrow$ +4.5)} & \textbf{\textcolor{insightteal}{($\uparrow$ +4.6)}} & 47.2\% & \textcolor{insightteal}{($\uparrow$ +3.0)} & \textcolor{insightteal}{($\uparrow$ +6.0)} & \textcolor{insightteal}{($\uparrow$ +4.0)} & \textbf{\textcolor{insightteal}{($\uparrow$ +6.6)}} \\
 & \textit{Dense} & 66.6\% & \textcolor{insightteal}{($\uparrow$ +2.5)} & \textcolor{insightteal}{($\uparrow$ +3.0)} & \textbf{\textcolor{insightteal}{($\uparrow$ +4.8)}} & \textcolor{insightteal}{($\uparrow$ +3.7)} & 47.2\% & \textcolor{insightteal}{($\uparrow$ +3.0)} & \textcolor{insightteal}{($\uparrow$ +6.0)} & \textcolor{insightteal}{($\uparrow$ +3.0)} & \textbf{\textcolor{insightteal}{($\uparrow$ +6.3)}} \\
\bottomrule
\end{tabular}%
}\vspace{0.5em}
\caption{\textbf{Best-of-5 Reranking Performance \& Baselines (\%).} Random is SFT pass@1. Deltas are percentage-point changes for each reranker. \textcolor{insightteal}{Blue arrows (\,$\uparrow$\,)} indicate gains and \textcolor{purple}{purple arrows (\,$\downarrow$\,)} indicate drops.}
\label{tab:reranking_baselines_full_N5}
\end{table}

\begin{table}[h!]
\centering
\scriptsize
\resizebox{\textwidth}{!}{%
\begin{tabular}{ll ccccc ccccc}
\toprule
& & \multicolumn{5}{c}{\textbf{\textsc{GSM8K}}} & \multicolumn{5}{c}{\textbf{\textsc{MMLU-Pro}}} \\
\cmidrule(lr){3-7} \cmidrule(lr){8-12}
\textbf{Backbone} & \textbf{Method} & Random & $\Delta$ Logp. & $\Delta$ Maj. & $\Delta$ Rew. & $\Delta$ W.Maj. & Random & $\Delta$ Logp. & $\Delta$ Maj. & $\Delta$ Rew. & $\Delta$ W.Maj. \\
\midrule
\multirow{4}{*}{\textbf{\texttt{Qwen2.5-4B}}} & \textit{Sparse} & 76.6\% & \textcolor{insightteal}{($\uparrow$ +0.1)} & \textcolor{insightteal}{($\uparrow$ +8.3)} & \textbf{\textcolor{insightteal}{($\uparrow$ +9.1)}} & \textbf{\textcolor{insightteal}{($\uparrow$ +9.1)}} & 53.9\% & \textcolor{insightteal}{($\uparrow$ +2.4)} & \textcolor{insightteal}{($\uparrow$ +6.0)} & \textcolor{insightteal}{($\uparrow$ +7.9)} & \textbf{\textcolor{insightteal}{($\uparrow$ +8.1)}} \\
 & \textit{Interval} & 76.6\% & \textcolor{insightteal}{($\uparrow$ +0.1)} & \textcolor{insightteal}{($\uparrow$ +8.3)} & \textcolor{insightteal}{($\uparrow$ +7.0)} & \textbf{\textcolor{insightteal}{($\uparrow$ +8.9)}} & 53.9\% & \textcolor{insightteal}{($\uparrow$ +2.4)} & \textbf{\textcolor{insightteal}{($\uparrow$ +6.0)}} & \textcolor{insightteal}{($\uparrow$ +1.9)} & \textcolor{insightteal}{($\uparrow$ +5.8)} \\
 & \textit{Dense} & 76.6\% & \textcolor{insightteal}{($\uparrow$ +0.1)} & \textcolor{insightteal}{($\uparrow$ +8.3)} & \textcolor{insightteal}{($\uparrow$ +7.7)} & \textbf{\textcolor{insightteal}{($\uparrow$ +9.1)}} & 53.9\% & \textcolor{insightteal}{($\uparrow$ +2.4)} & \textcolor{insightteal}{($\uparrow$ +6.0)} & \textcolor{insightteal}{($\uparrow$ +4.6)} & \textbf{\textcolor{insightteal}{($\uparrow$ +6.5)}} \\
\midrule
\multirow{4}{*}{\textbf{\texttt{Qwen2.5-7B}}} & \textit{Sparse} & 70.1\% & \textcolor{insightteal}{($\uparrow$ +1.3)} & \textcolor{insightteal}{($\uparrow$ +13.0)} & \textcolor{insightteal}{($\uparrow$ +10.5)} & \textbf{\textcolor{insightteal}{($\uparrow$ +13.9)}} & 48.1\% & \textcolor{insightteal}{($\uparrow$ +2.1)} & \textcolor{insightteal}{($\uparrow$ +5.7)} & \textcolor{insightteal}{($\uparrow$ +6.9)} & \textbf{\textcolor{insightteal}{($\uparrow$ +7.6)}} \\
 & \textit{Interval} & 70.1\% & \textcolor{insightteal}{($\uparrow$ +1.3)} & \textcolor{insightteal}{($\uparrow$ +13.0)} & \textcolor{insightteal}{($\uparrow$ +8.7)} & \textbf{\textcolor{insightteal}{($\uparrow$ +13.7)}} & 48.1\% & \textcolor{insightteal}{($\uparrow$ +2.1)} & \textcolor{insightteal}{($\uparrow$ +5.7)} & \textcolor{insightteal}{($\uparrow$ +3.2)} & \textbf{\textcolor{insightteal}{($\uparrow$ +6.2)}} \\
 & \textit{Dense} & 70.1\% & \textcolor{insightteal}{($\uparrow$ +1.3)} & \textcolor{insightteal}{($\uparrow$ +13.0)} & \textcolor{insightteal}{($\uparrow$ +11.0)} & \textbf{\textcolor{insightteal}{($\uparrow$ +14.4)}} & 48.1\% & \textcolor{insightteal}{($\uparrow$ +2.1)} & \textcolor{insightteal}{($\uparrow$ +5.7)} & \textcolor{insightteal}{($\uparrow$ +0.9)} & \textbf{\textcolor{insightteal}{($\uparrow$ +6.2)}} \\
\midrule
\multirow{4}{*}{\textbf{\texttt{Llama3.1-8B}}} & \textit{Sparse} & 66.6\% & \textcolor{insightteal}{($\uparrow$ +2.4)} & \textcolor{insightteal}{($\uparrow$ +2.7)} & \textcolor{insightteal}{($\uparrow$ +3.0)} & \textbf{\textcolor{insightteal}{($\uparrow$ +4.3)}} & 47.2\% & \textcolor{insightteal}{($\uparrow$ +2.5)} & \textbf{\textcolor{insightteal}{($\uparrow$ +7.0)}} & \textcolor{insightteal}{($\uparrow$ +1.4)} & \textcolor{insightteal}{($\uparrow$ +6.6)} \\
 & \textit{Interval} & 66.6\% & \textcolor{insightteal}{($\uparrow$ +2.4)} & \textcolor{insightteal}{($\uparrow$ +2.7)} & \textbf{\textcolor{insightteal}{($\uparrow$ +5.3)}} & \textcolor{insightteal}{($\uparrow$ +4.8)} & 47.2\% & \textcolor{insightteal}{($\uparrow$ +2.5)} & \textcolor{insightteal}{($\uparrow$ +7.0)} & \textcolor{insightteal}{($\uparrow$ +4.1)} & \textbf{\textcolor{insightteal}{($\uparrow$ +7.8)}} \\
 & \textit{Dense} & 66.6\% & \textcolor{insightteal}{($\uparrow$ +2.4)} & \textcolor{insightteal}{($\uparrow$ +2.7)} & \textbf{\textcolor{insightteal}{($\uparrow$ +5.8)}} & \textcolor{insightteal}{($\uparrow$ +4.1)} & 47.2\% & \textcolor{insightteal}{($\uparrow$ +2.5)} & \textcolor{insightteal}{($\uparrow$ +7.0)} & \textcolor{insightteal}{($\uparrow$ +3.7)} & \textbf{\textcolor{insightteal}{($\uparrow$ +7.4)}} \\
\bottomrule
\end{tabular}%
}\vspace{0.5em}
\caption{\textbf{Best-of-8 Reranking Performance \& Baselines (\%).} Random is SFT pass@1. Deltas are percentage-point changes for each reranker. \textcolor{insightteal}{Blue arrows (\,$\uparrow$\,)} indicate gains and \textcolor{purple}{purple arrows (\,$\downarrow$\,)} indicate drops.}
\label{tab:reranking_baselines_full_N8}
\end{table}

\begin{table}[h!]
\centering
\scriptsize
\resizebox{\textwidth}{!}{%
\begin{tabular}{ll ccccc ccccc}
\toprule
& & \multicolumn{5}{c}{\textbf{\textsc{GSM8K}}} & \multicolumn{5}{c}{\textbf{\textsc{MMLU-Pro}}} \\
\cmidrule(lr){3-7} \cmidrule(lr){8-12}
\textbf{Backbone} & \textbf{Method} & Random & $\Delta$ Logp. & $\Delta$ Maj. & $\Delta$ Rew. & $\Delta$ W.Maj. & Random & $\Delta$ Logp. & $\Delta$ Maj. & $\Delta$ Rew. & $\Delta$ W.Maj. \\
\midrule
\multirow{4}{*}{\textbf{\texttt{Qwen2.5-4B}}} & \textit{Sparse} & 76.6\% & \textcolor{insightteal}{($\uparrow$ +1.4)} & \textcolor{insightteal}{($\uparrow$ +8.7)} & \textcolor{insightteal}{($\uparrow$ +10.3)} & \textbf{\textcolor{insightteal}{($\uparrow$ +10.6)}} & 53.9\% & \textcolor{insightteal}{($\uparrow$ +2.1)} & \textcolor{insightteal}{($\uparrow$ +7.7)} & \textcolor{insightteal}{($\uparrow$ +8.7)} & \textbf{\textcolor{insightteal}{($\uparrow$ +9.3)}} \\
 & \textit{Interval} & 76.6\% & \textcolor{insightteal}{($\uparrow$ +1.4)} & \textcolor{insightteal}{($\uparrow$ +8.7)} & \textcolor{insightteal}{($\uparrow$ +8.1)} & \textbf{\textcolor{insightteal}{($\uparrow$ +10.2)}} & 53.9\% & \textcolor{insightteal}{($\uparrow$ +2.1)} & \textbf{\textcolor{insightteal}{($\uparrow$ +7.7)}} & \textcolor{insightteal}{($\uparrow$ +2.2)} & \textcolor{insightteal}{($\uparrow$ +7.5)} \\
 & \textit{Dense} & 76.6\% & \textcolor{insightteal}{($\uparrow$ +1.4)} & \textcolor{insightteal}{($\uparrow$ +8.7)} & \textcolor{insightteal}{($\uparrow$ +9.2)} & \textbf{\textcolor{insightteal}{($\uparrow$ +10.6)}} & 53.9\% & \textcolor{insightteal}{($\uparrow$ +2.1)} & \textbf{\textcolor{insightteal}{($\uparrow$ +7.7)}} & \textcolor{insightteal}{($\uparrow$ +4.7)} & \textcolor{insightteal}{($\uparrow$ +7.5)} \\
\midrule
\multirow{4}{*}{\textbf{\texttt{Qwen2.5-7B}}} & \textit{Sparse} & 70.1\% & \textcolor{insightteal}{($\uparrow$ +0.8)} & \textcolor{insightteal}{($\uparrow$ +15.6)} & \textcolor{insightteal}{($\uparrow$ +11.3)} & \textbf{\textcolor{insightteal}{($\uparrow$ +16.4)}} & 48.1\% & \textcolor{insightteal}{($\uparrow$ +2.6)} & \textcolor{insightteal}{($\uparrow$ +7.5)} & \textcolor{insightteal}{($\uparrow$ +7.0)} & \textbf{\textcolor{insightteal}{($\uparrow$ +8.6)}} \\
 & \textit{Interval} & 70.1\% & \textcolor{insightteal}{($\uparrow$ +0.8)} & \textbf{\textcolor{insightteal}{($\uparrow$ +15.6)}} & \textcolor{insightteal}{($\uparrow$ +8.3)} & \textcolor{insightteal}{($\uparrow$ +15.3)} & 48.1\% & \textcolor{insightteal}{($\uparrow$ +2.6)} & \textcolor{insightteal}{($\uparrow$ +7.5)} & \textcolor{insightteal}{($\uparrow$ +4.1)} & \textbf{\textcolor{insightteal}{($\uparrow$ +8.1)}} \\
 & \textit{Dense} & 70.1\% & \textcolor{insightteal}{($\uparrow$ +0.8)} & \textcolor{insightteal}{($\uparrow$ +15.6)} & \textcolor{insightteal}{($\uparrow$ +11.1)} & \textbf{\textcolor{insightteal}{($\uparrow$ +15.7)}} & 48.1\% & \textcolor{insightteal}{($\uparrow$ +2.6)} & \textcolor{insightteal}{($\uparrow$ +7.5)} & \textcolor{insightteal}{($\uparrow$ +0.5)} & \textbf{\textcolor{insightteal}{($\uparrow$ +7.7)}} \\
\midrule
\multirow{4}{*}{\textbf{\texttt{Llama3.1-8B}}} & \textit{Sparse} & 66.6\% & \textcolor{insightteal}{($\uparrow$ +2.8)} & \textcolor{insightteal}{($\uparrow$ +4.6)} & \textcolor{insightteal}{($\uparrow$ +3.7)} & \textbf{\textcolor{insightteal}{($\uparrow$ +5.2)}} & 47.2\% & \textcolor{insightteal}{($\uparrow$ +2.7)} & \textcolor{insightteal}{($\uparrow$ +7.6)} & \textcolor{purple}{($\downarrow$ -0.2)} & \textbf{\textcolor{insightteal}{($\uparrow$ +8.3)}} \\
 & \textit{Interval} & 66.6\% & \textcolor{insightteal}{($\uparrow$ +2.8)} & \textcolor{insightteal}{($\uparrow$ +4.6)} & \textcolor{insightteal}{($\uparrow$ +5.9)} & \textbf{\textcolor{insightteal}{($\uparrow$ +6.8)}} & 47.2\% & \textcolor{insightteal}{($\uparrow$ +2.7)} & \textcolor{insightteal}{($\uparrow$ +7.6)} & \textcolor{insightteal}{($\uparrow$ +4.4)} & \textbf{\textcolor{insightteal}{($\uparrow$ +8.4)}} \\
 & \textit{Dense} & 66.6\% & \textcolor{insightteal}{($\uparrow$ +2.8)} & \textcolor{insightteal}{($\uparrow$ +4.6)} & \textbf{\textcolor{insightteal}{($\uparrow$ +6.5)}} & \textcolor{insightteal}{($\uparrow$ +5.8)} & 47.2\% & \textcolor{insightteal}{($\uparrow$ +2.7)} & \textcolor{insightteal}{($\uparrow$ +7.6)} & \textcolor{insightteal}{($\uparrow$ +3.2)} & \textbf{\textcolor{insightteal}{($\uparrow$ +7.9)}} \\
\bottomrule
\end{tabular}%
}\vspace{0.5em}
\caption{\textbf{Best-of-16 Reranking Performance \& Baselines (\%).} Random is SFT pass@1. Deltas are percentage-point changes for each reranker. \textcolor{insightteal}{Blue arrows (\,$\uparrow$\,)} indicate gains and \textcolor{purple}{purple arrows (\,$\downarrow$\,)} indicate drops.}
\label{tab:reranking_baselines_full_N16}
\end{table}

\clearpage
\subsection{Distribution Separation of Rewards}\label{app:distributions}
\subsubsection{\textsc{GSM8K}}

\begin{figure}[h!]
    \centering
    \captionsetup{justification=centering}
    \begin{subfigure}{0.32\textwidth}
        \centering
        \includegraphics[width=\linewidth]{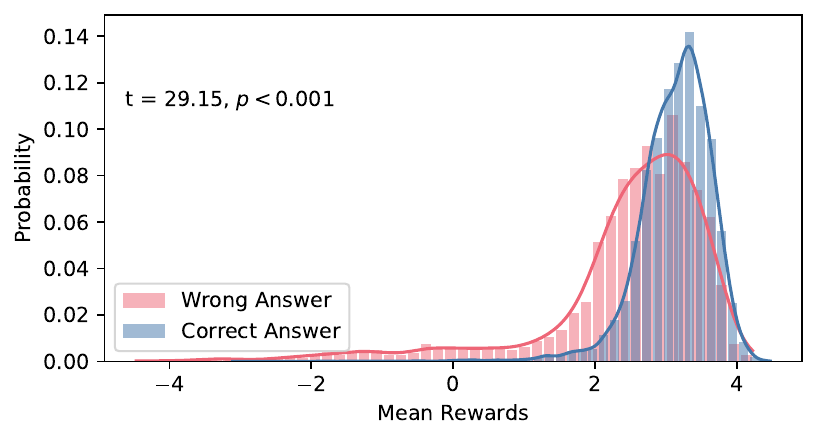}
        \caption{Sparse}
        \label{fig:qwen7b-rewards-separation-math-sparse}
    \end{subfigure}
    \hfill
    \begin{subfigure}{0.32\textwidth}
        \centering
        \includegraphics[width=\linewidth]{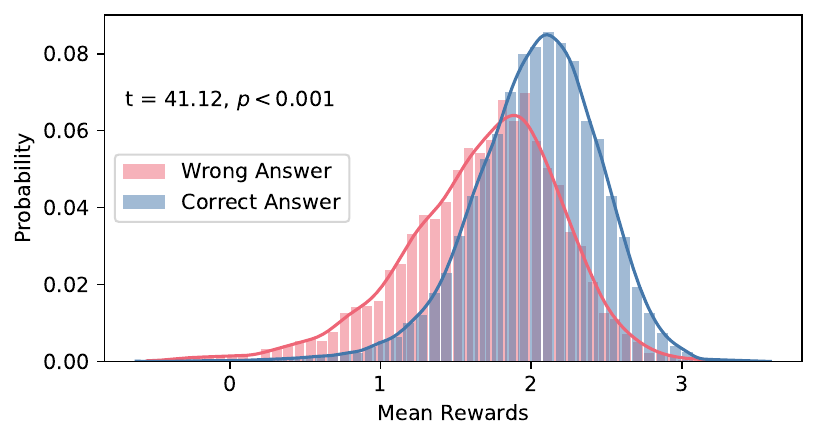}
        \caption{Interval}
        \label{fig:qwen7b-rewards-separation-math-interval}
    \end{subfigure}
    \hfill
    \begin{subfigure}{0.32\textwidth}
        \centering
        \includegraphics[width=\linewidth]{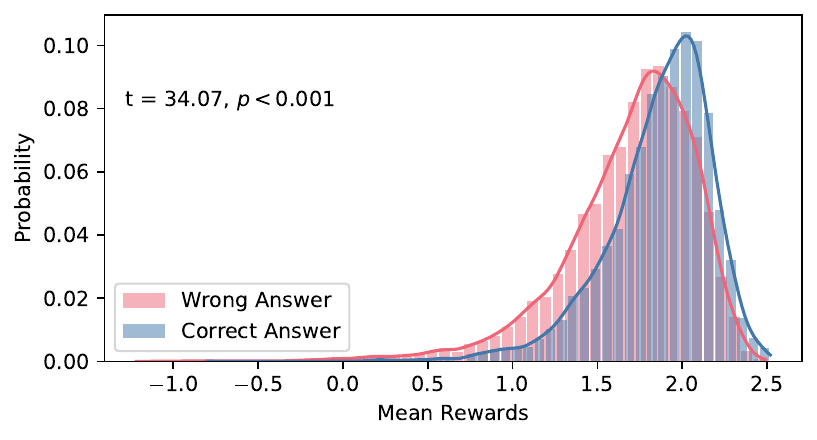}
        \caption{Dense}
        \label{fig:qwen7b-rewards-separation-math-dense}
    \end{subfigure}
    \caption[Reward Distribution by Correctness (\texttt{Qwen2.5-7B} on \textsc{GSM8K})]{\textbf{Reward Distribution by Correctness (\texttt{Qwen2.5-7B} on \textsc{GSM8K})}. A t-test assesses the significance of the difference in mean rewards between correct and incorrect answers.}
    \label{fig:qwen7b-rewards-separation-math}
\end{figure}

\begin{figure}[h!]
    \centering
    \captionsetup{justification=centering}
    \begin{subfigure}{0.32\textwidth}
        \centering
        \includegraphics[width=\linewidth]{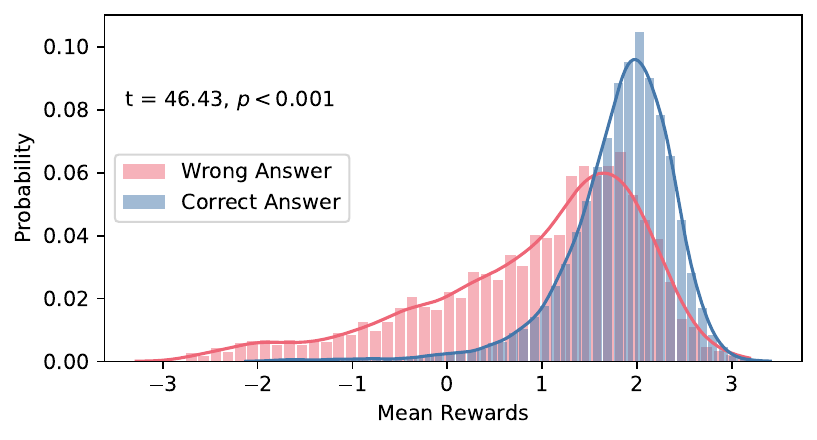}
        \caption{Sparse}
        \label{fig:llama8b-rewards-separation-math-sparse}
    \end{subfigure}
    \hfill
    \begin{subfigure}{0.32\textwidth}
        \centering
        \includegraphics[width=\linewidth]{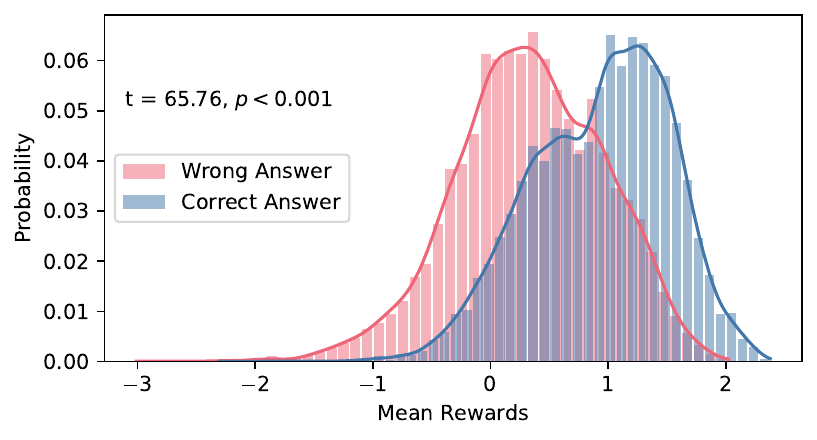}
        \caption{Interval}
        \label{fig:llama8b-rewards-separation-math-interval}
    \end{subfigure}
    \hfill
    \begin{subfigure}{0.32\textwidth}
        \centering
        \includegraphics[width=\linewidth]{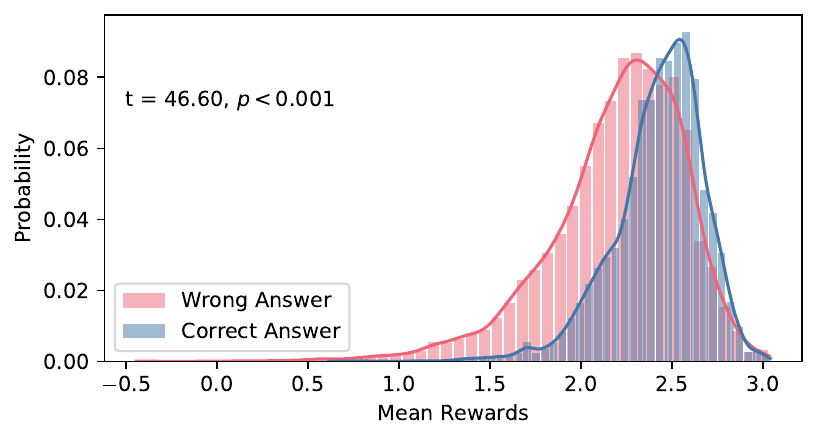}
        \caption{Dense}
        \label{fig:llama8b-rewards-separation-math-dense}
    \end{subfigure}
    \caption[Reward Distribution by Correctness (\texttt{Llama3.1-8B} on \textsc{GSM8K})]{\textbf{Reward Distribution by Correctness (\texttt{Llama3.1-8B} on \textsc{GSM8K})}. A t-test assesses the significance of the difference in mean rewards between correct and incorrect answers.}
    \label{fig:llama8b-rewards-separation-math}
\end{figure}

\begin{figure}[h!]
    \centering
    \captionsetup{justification=centering}
    \begin{subfigure}{0.32\textwidth}
        \centering
        \includegraphics[width=\linewidth]{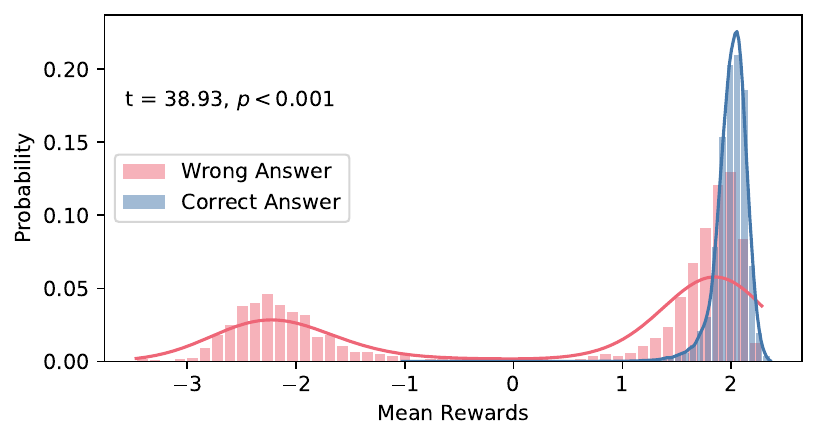}
        \caption{Sparse}
        \label{fig:qwen4b-rewards-separation-math-sparse}
    \end{subfigure}
    \hfill
    \begin{subfigure}{0.32\textwidth}
        \centering
        \includegraphics[width=\linewidth]{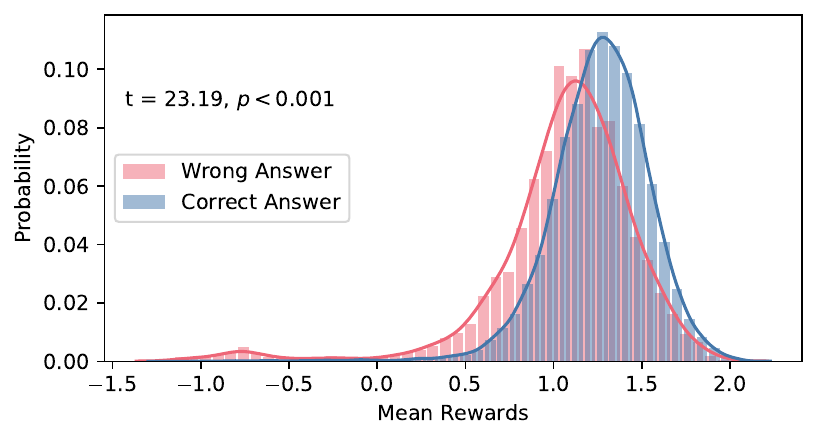}
        \caption{Interval}
        \label{fig:qwen4b-rewards-separation-math-interval}
    \end{subfigure}
    \hfill
    \begin{subfigure}{0.32\textwidth}
        \centering
        \includegraphics[width=\linewidth]{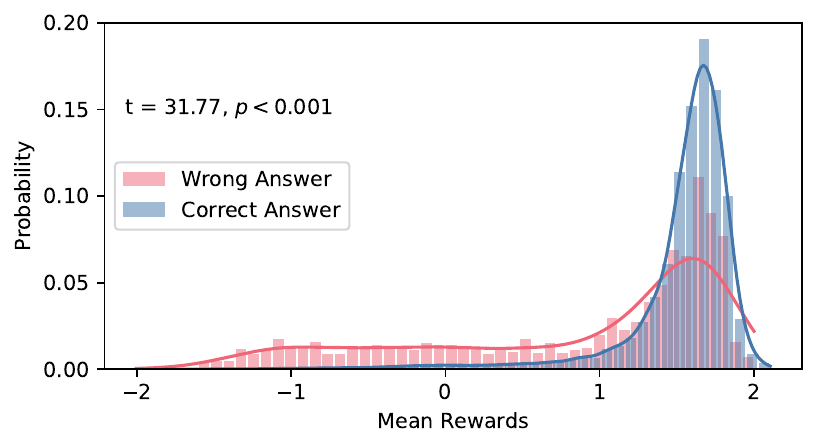}
        \caption{Dense}
        \label{fig:qwen4b-rewards-separation-math-dense}
    \end{subfigure}
    \caption[Reward Distribution by Correctness (\texttt{Qwen3-4B} on \textsc{GSM8K})]{\textbf{Reward Distribution by Correctness (\texttt{Qwen3-4B} on \textsc{GSM8K})}. A t-test assesses the significance of the difference in mean rewards between correct and incorrect answers.}
    \label{fig:lqwen4b-rewards-separation-math}
\end{figure}

\newpage

\subsubsection{\textsc{MMLU-Pro}}

\begin{figure}[h!]
    \centering
    \captionsetup{justification=centering}
    \begin{subfigure}{0.32\textwidth}
        \centering
        \includegraphics[width=\linewidth]{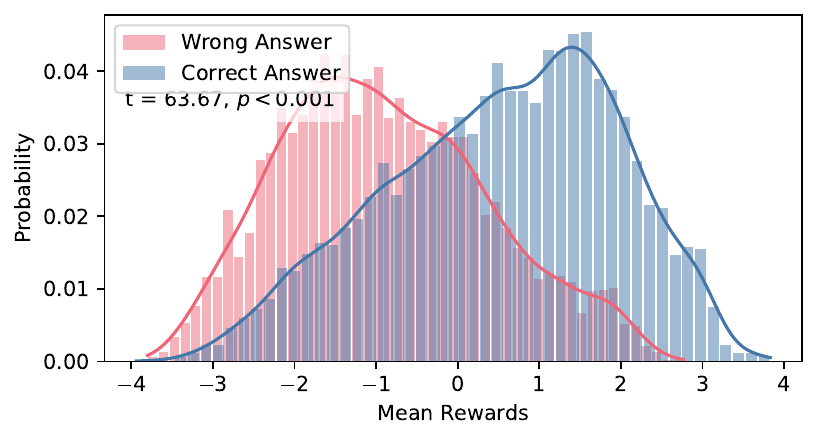}
        \caption{Sparse}
        \label{fig:qwen7b-rewards-separation-mmlu-sparse}
    \end{subfigure}
    \hfill
    \begin{subfigure}{0.32\textwidth}
        \centering
        \includegraphics[width=\linewidth]{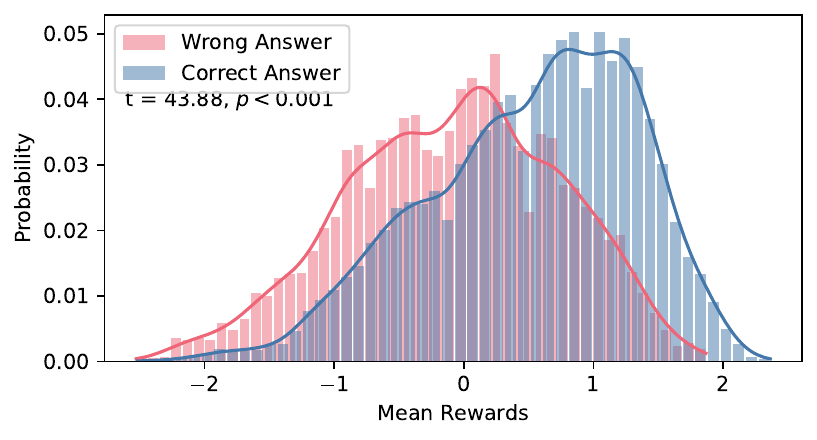}
        \caption{Interval}
        \label{fig:qwen7b-rewards-separation-mmlu-interval}
    \end{subfigure}
    \hfill
    \begin{subfigure}{0.32\textwidth}
        \centering
        \includegraphics[width=\linewidth]{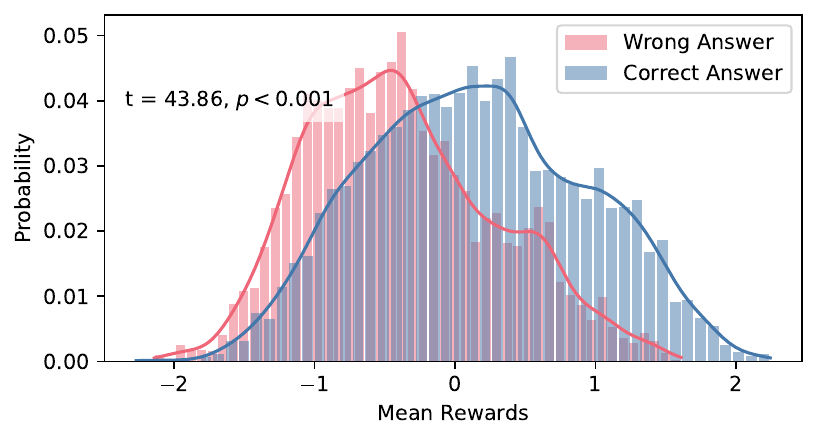}
        \caption{Dense}
        \label{fig:qwen7b-rewards-separation-mmlu-dense}
    \end{subfigure}
    \caption[Reward Distribution by Correctness (\texttt{Qwen2.5-7B} on \textsc{MMLU-Pro})]{\textbf{Reward Distribution by Correctness (\texttt{Qwen2.5-7B} on \textsc{MMLU-Pro})}. A t-test assesses the significance of the difference in mean rewards between correct and incorrect answers.}
    \label{fig:qwen7b-rewards-separation-mmlu}
\end{figure}

\begin{figure}[h!]
    \centering
    \captionsetup{justification=centering}
    \begin{subfigure}{0.32\textwidth}
        \centering
        \includegraphics[width=\linewidth]{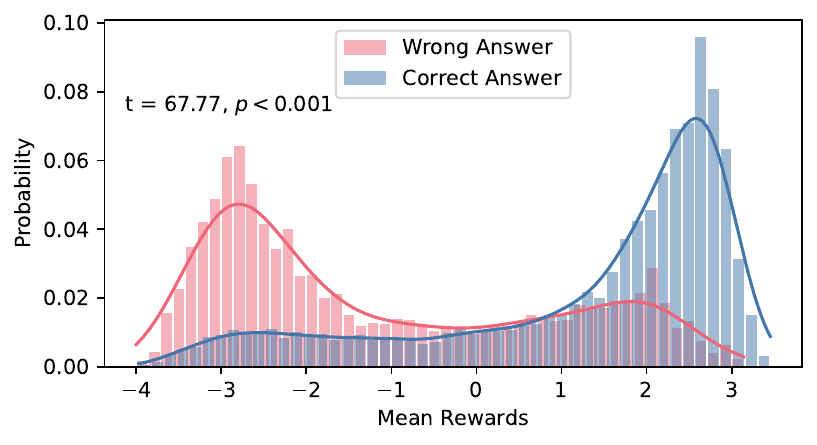}
        \caption{Sparse}
        \label{fig:llama8b-rewards-separation-mmlu-sparse}
    \end{subfigure}
    \hfill
    \centering
    \begin{subfigure}{0.32\textwidth}
        \centering
        \includegraphics[width=\linewidth]{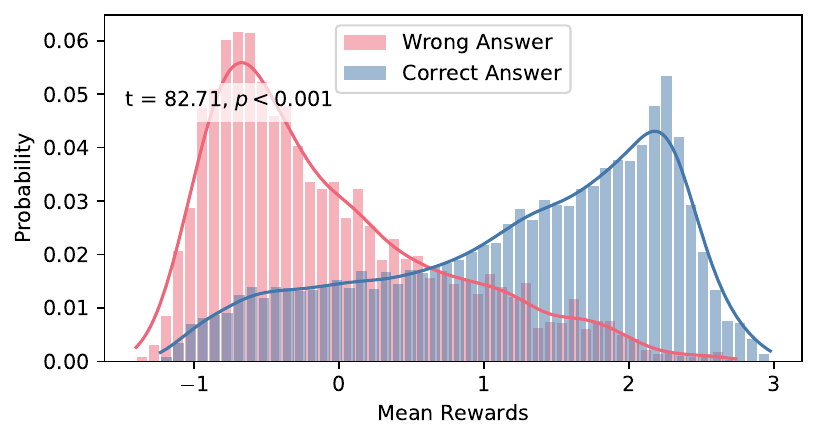}
        \caption{Interval}
        \label{fig:llama8b-rewards-separation-mmlu-interval}
    \end{subfigure}
    \hfill
    \begin{subfigure}{0.32\textwidth}
        \centering
        \includegraphics[width=\linewidth]{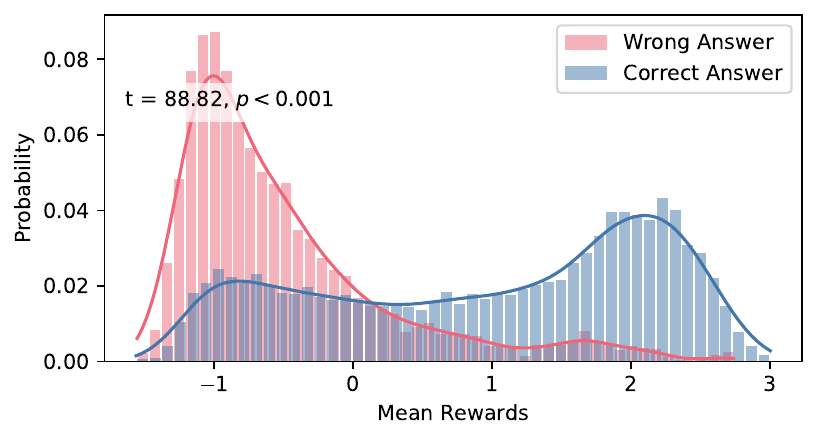}
        \caption{Dense}
        \label{fig:llama8b-rewards-separation-mmlu-dense}
    \end{subfigure}
    \caption[Reward Distribution by Correctness (\texttt{Llama3.1-8B} on \textsc{MMLU-Pro})]{\textbf{Reward Distribution by Correctness (\texttt{Llama3.1-8B} on \textsc{MMLU-Pro})}. A t-test assesses the significance of the difference in mean rewards between correct and incorrect answers.}
    \label{fig:llama8b-rewards-separation-mmlu}
\end{figure}

\begin{figure}[h!]
    \centering
    \captionsetup{justification=centering}
    \begin{subfigure}{0.32\textwidth}
        \centering
        \includegraphics[width=\linewidth]{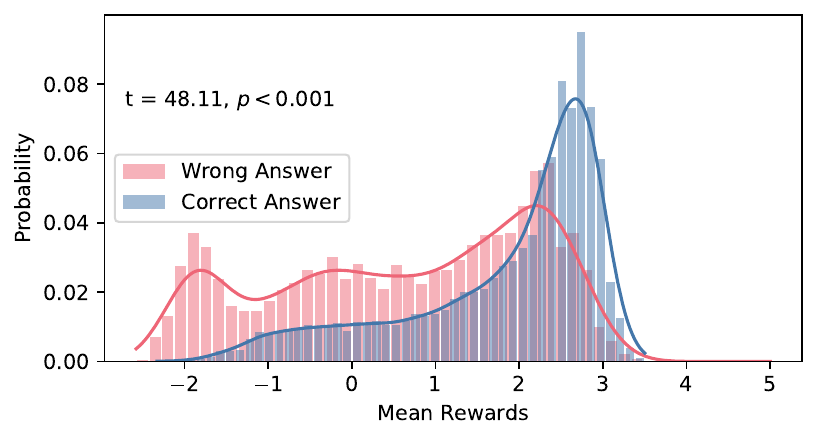}
        \caption{Sparse}
        \label{fig:qwen4b-rewards-separation-mmlu-sparse}
    \end{subfigure}
    \hfill
    \begin{subfigure}{0.32\textwidth}
        \centering
        \includegraphics[width=\linewidth]{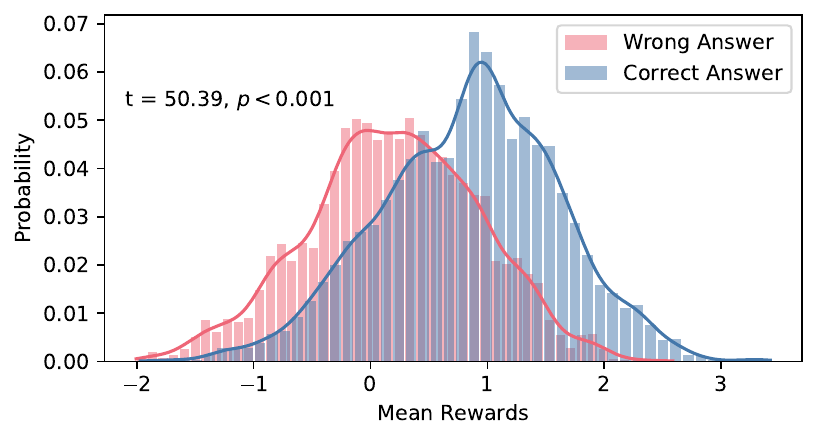}
        \caption{Interval}
        \label{fig:qwen4b-rewards-separation-mmlu-interval}
    \end{subfigure}
    \hfill
    \begin{subfigure}{0.32\textwidth}
        \centering
        \includegraphics[width=\linewidth]{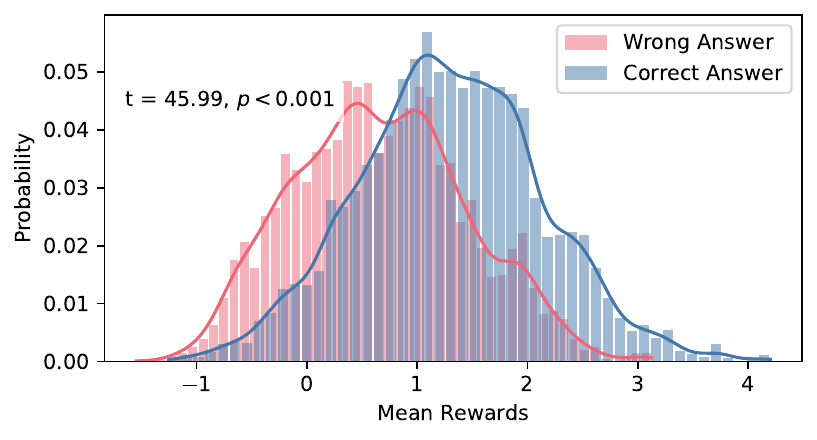}
        \caption{Dense}
        \label{fig:qwen4b-rewards-separation-mmlu-dense}
    \end{subfigure}
    \caption[Reward Distribution by Correctness (\texttt{Qwen3-4B} on \textsc{MMLU-Pro})]{\textbf{Reward Distribution by Correctness (\texttt{Qwen3-4B} on \textsc{MMLU-Pro})}. A t-test assesses the significance of the difference in mean rewards between correct and incorrect answers.}
    \label{fig:lqwen4b-rewards-separation-mmlu}
\end{figure}

\newpage
\subsubsection{\textsc{MedReason}}

\begin{figure}[h!]
    \centering
    \captionsetup{justification=centering}
    \begin{subfigure}{0.32\textwidth}
        \centering
        \includegraphics[width=\linewidth]{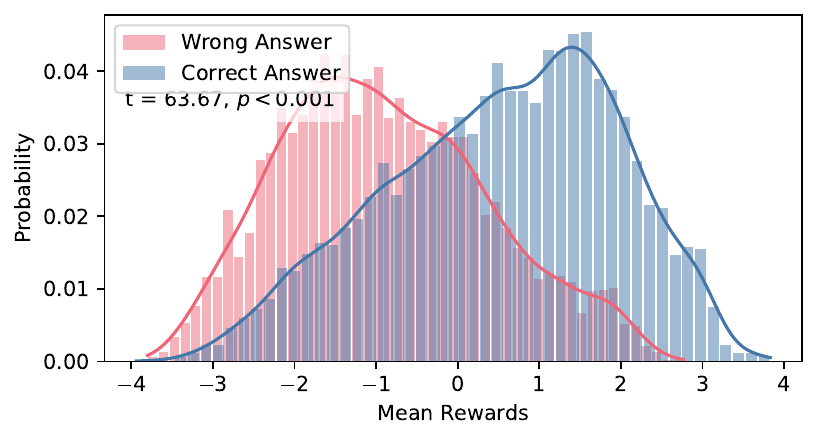}
        \caption{Sparse}
        \label{fig:qwen7b-rewards-separation-medicine-sparse}
    \end{subfigure}
    \hfill
    \centering
    \begin{subfigure}{0.32\textwidth}
        \centering
        \includegraphics[width=\linewidth]{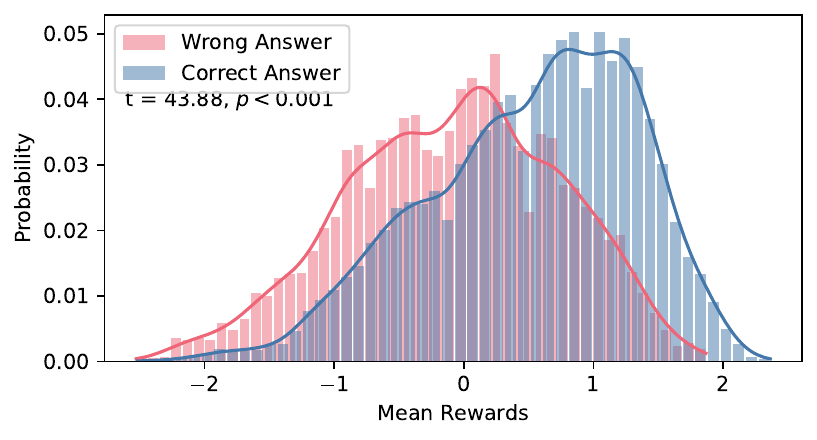}
        \caption{Interval}
        \label{fig:qwen7b-rewards-separation-medicine-interval}
    \end{subfigure}
    \hfill
    \begin{subfigure}{0.32\textwidth}
        \centering
        \includegraphics[width=\linewidth]{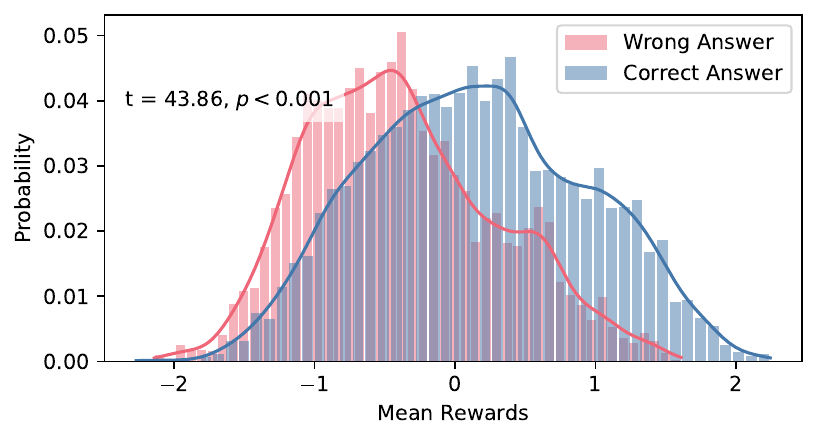}
        \caption{Dense}
        \label{fig:qwen7b-rewards-separation-medicine-dense}
    \end{subfigure}
    \caption[Reward Distribution by Correctness (\texttt{Qwen2.5-7B} on \textsc{MedReason})]{\textbf{Reward Distribution by Correctness (\texttt{Qwen2.5-7B} on \textsc{MedReason})}. A t-test assesses the significance of the difference in mean rewards between correct and incorrect answers.}
    \label{fig:qwen7b-rewards-separation-medicine}
\end{figure}

\begin{figure}[h!]
    \centering
    \captionsetup{justification=centering}
    \begin{subfigure}{0.32\textwidth}
        \centering
        \includegraphics[width=\linewidth]{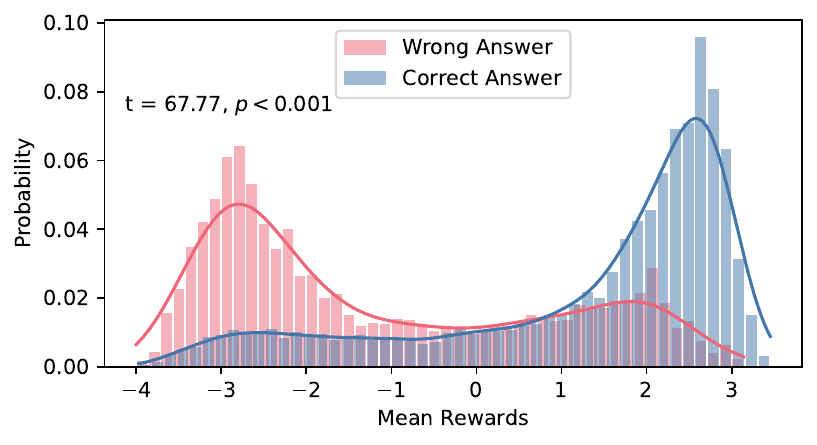}
        \caption{Sparse}
        \label{fig:llama8b-rewards-separation-medicine-sparse}
    \end{subfigure}
    \hfill
    \begin{subfigure}{0.32\textwidth}
        \centering
        \includegraphics[width=\linewidth]{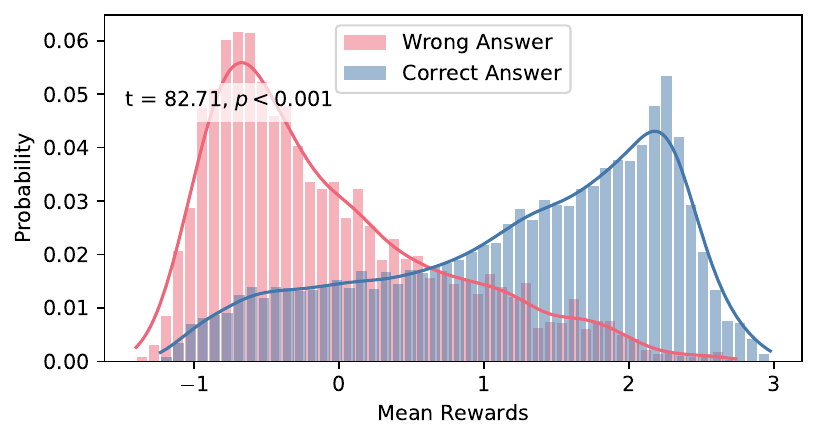}
        \caption{Interval}
        \label{fig:llama8b-rewards-separation-medicine-interval}
    \end{subfigure}
    \hfill
    \begin{subfigure}{0.32\textwidth}
        \centering
        \includegraphics[width=\linewidth]{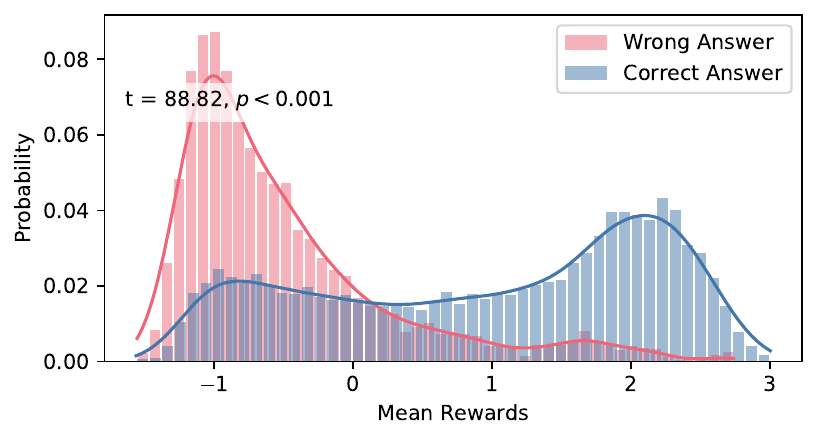}
        \caption{Dense}
        \label{fig:llama8b-rewards-separation-medicine-dense}
    \end{subfigure}
    \caption[Reward Distribution by Correctness (\texttt{Llama3.1-8B} on \textsc{MedReason})]{\textbf{Reward Distribution by Correctness (\texttt{Llama3.1-8B} on \textsc{MedReason})}. A t-test assesses the significance of the difference in mean rewards between correct and incorrect answers.}
    \label{fig:llama8b-rewards-separation-medicine}
\end{figure}

\begin{figure}[h!]
    \centering
    \captionsetup{justification=centering}
    \begin{subfigure}{0.32\textwidth}
        \centering
        \includegraphics[width=\linewidth]{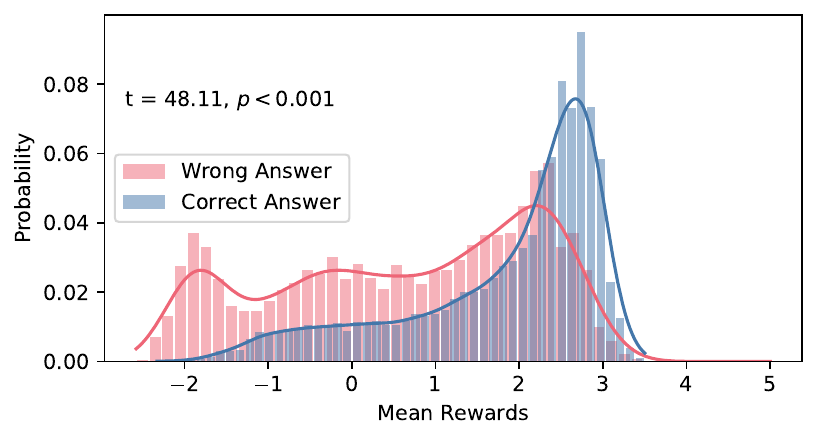}
        \caption{Sparse}
        \label{fig:qwen4b-rewards-separation-medicine-sparse}
    \end{subfigure}
    \hfill
    \begin{subfigure}{0.32\textwidth}
        \centering
        \includegraphics[width=\linewidth]{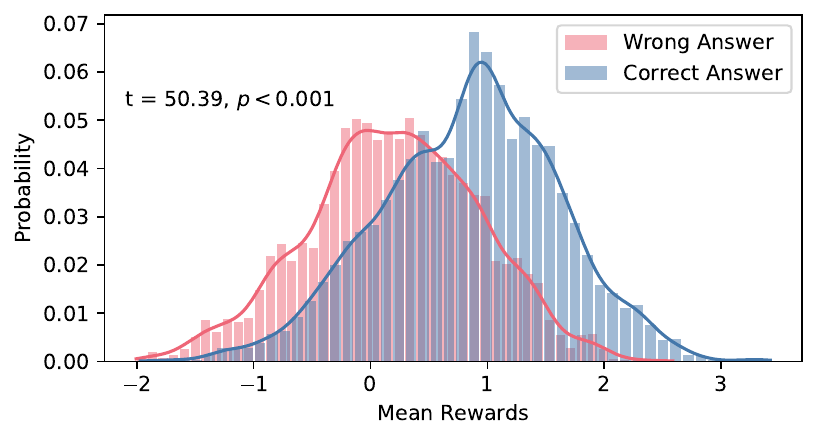}
        \caption{Interval}
        \label{fig:qwen4b-rewards-separation-medicine-interval}
    \end{subfigure}
    \hfill
    \begin{subfigure}{0.32\textwidth}
        \centering
        \includegraphics[width=\linewidth]{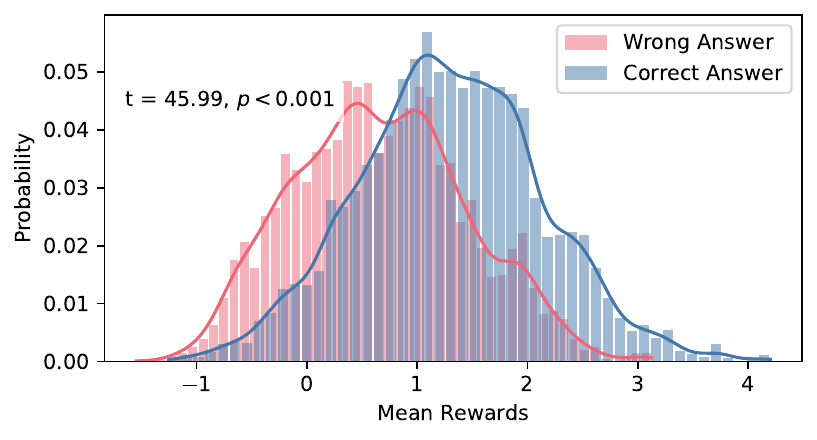}
        \caption{Dense}
        \label{fig:qwen4b-rewards-separation-medicine-dense}
    \end{subfigure}
    \caption[Reward Distribution by Correctness (\texttt{Qwen3-4B} on \textsc{MedReason})]{\textbf{Reward Distribution by Correctness (\texttt{Qwen3-4B} on \textsc{MedReason})}. A t-test assesses the significance of the difference in mean rewards between correct and incorrect answers.}
    \label{fig:lqwen4b-rewards-separation-medicine}
\end{figure}

\clearpage

\subsection{Error Localisation}\label{app:error_localisation_app}

\begin{figure}[h!]
    \centering
    \captionsetup{justification=centering}

    \begin{subfigure}[t]{\textwidth}
        \vspace{0pt}
        \centering
        \includegraphics[width=\linewidth]{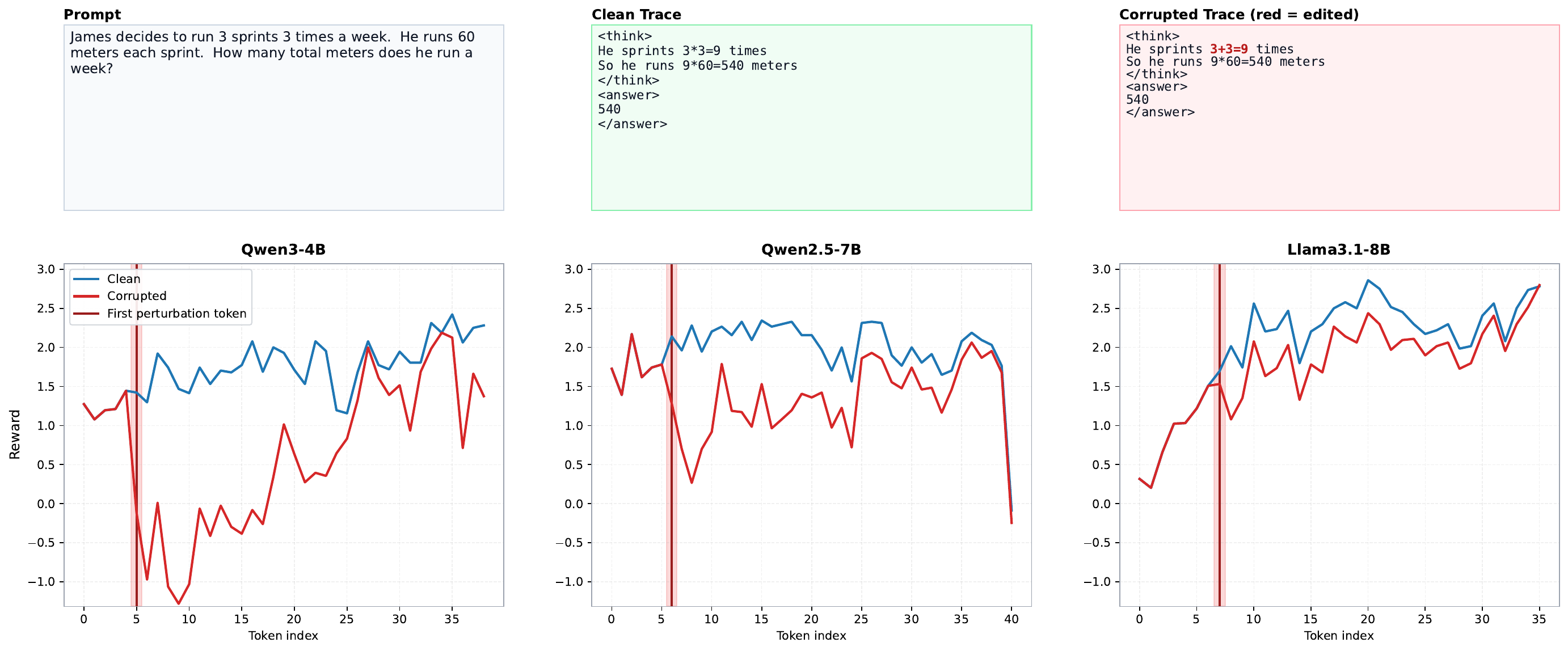}
        \caption{Example 1}
    \end{subfigure}
    \vspace{1em}
    \begin{subfigure}[t]{\textwidth}
        \vspace{0pt}
        \centering
        \includegraphics[width=\linewidth]{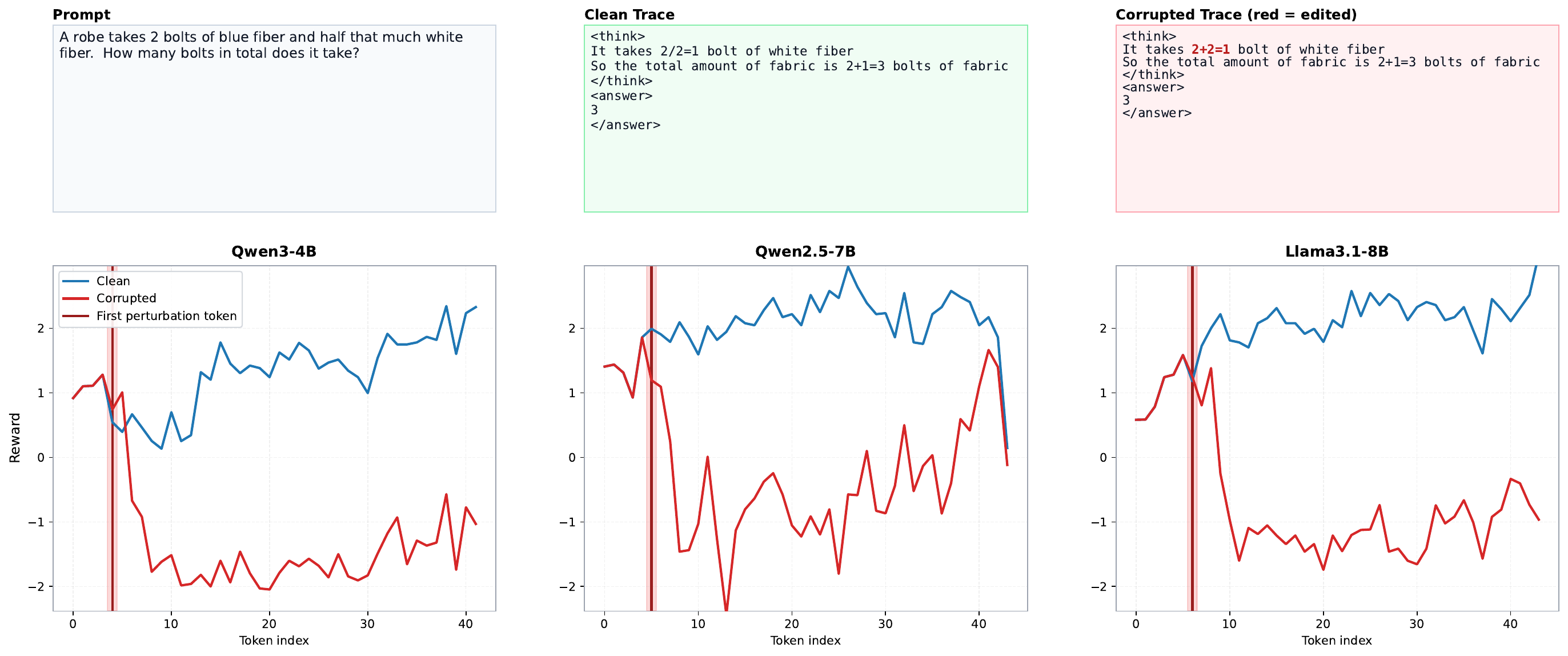}
        \caption{Example 2}
    \end{subfigure}
    \vspace{1em}
    \begin{subfigure}[t]{\textwidth}
        \vspace{0pt}
        \centering
        \includegraphics[width=\linewidth]{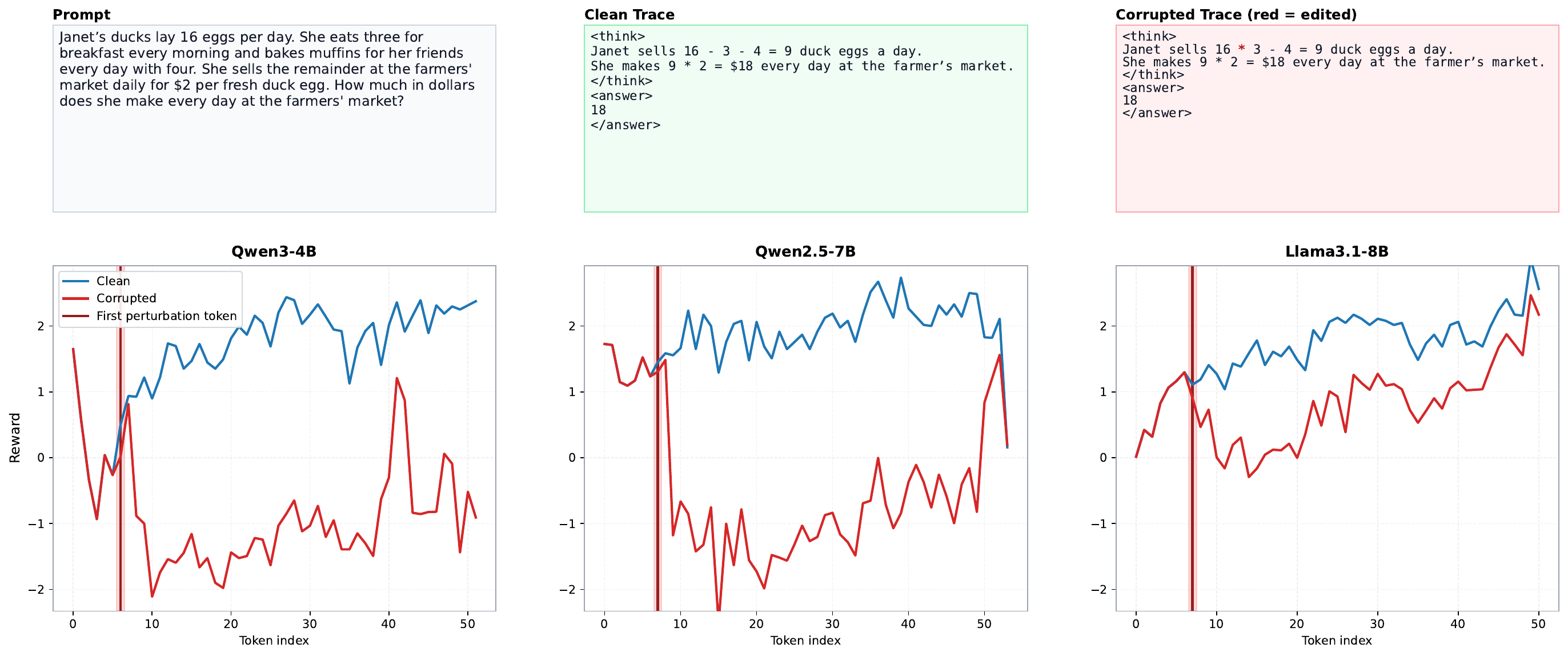}
        \caption{Example 3}
    \end{subfigure}

    \caption[Error Localisation on \textsc{GSM8K}]{\textbf{Error Localisation on \textsc{GSM8K}}
    Dense reward on correct and incorrect generations using the \textit{dense} reasoning reward models. We clearly observe a divergence between the correct and incorrect paths.}
    \label{fig:qwen7b-reasoning-math}
\end{figure}

\subsection{Reasoning Traces}\label{app:reasoning-traces}

\subsubsection{GSM8K}

\begin{figure}[h!]
    \centering
    \captionsetup{justification=centering}

    \vspace{0.9em}

    \begin{subfigure}[t]{0.49\textwidth}
        \vspace{0pt}
        \centering
        \includegraphics[width=\linewidth]{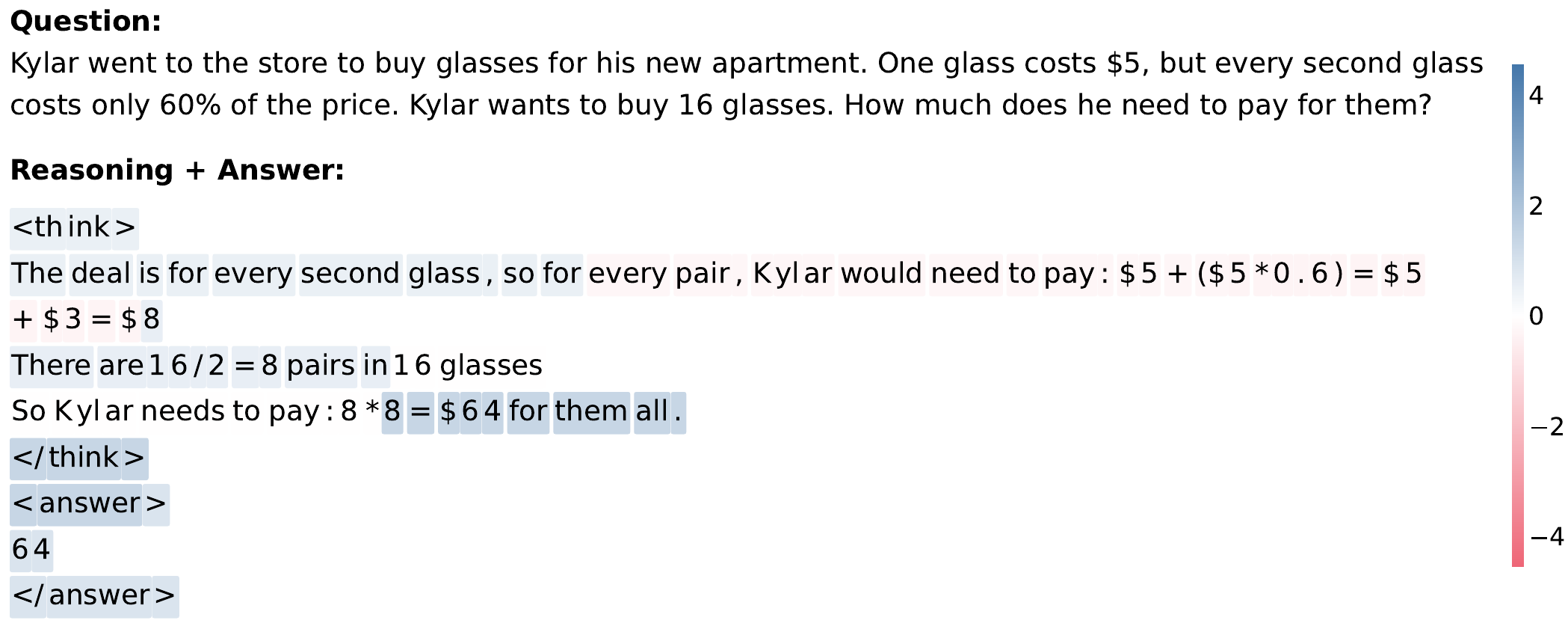}
        \caption{\textit{Interval}, correct}
        \label{fig:qwen7b-reasoning-interval-true-math}
    \end{subfigure}
    \hfill
    \begin{subfigure}[t]{0.49\textwidth}
        \vspace{0pt}
        \centering
        \includegraphics[width=\linewidth]{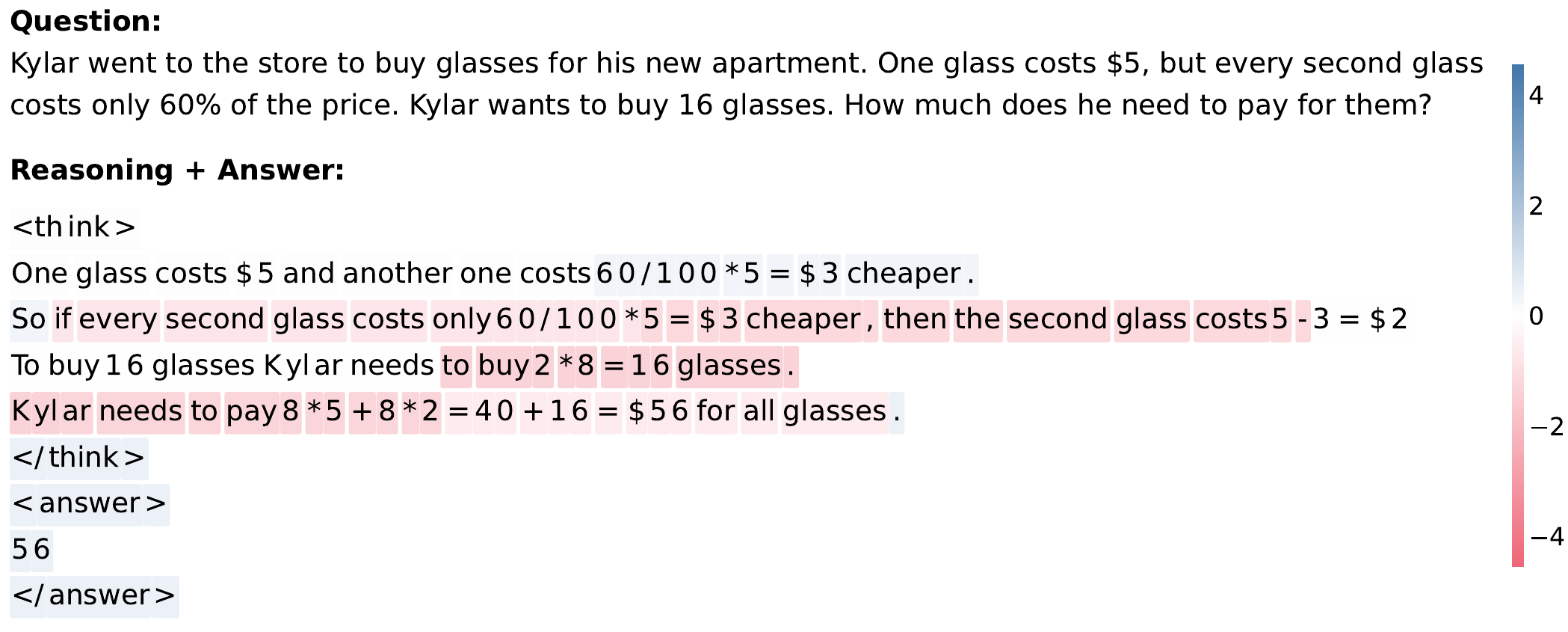}
        \caption{\textit{Interval}, incorrect}
        \label{fig:qwen7b-reasoning-interval-wrong-math}
    \end{subfigure}

    \vspace{0.9em}

    \begin{subfigure}[t]{0.49\textwidth}
        \vspace{0pt}
        \centering
        \includegraphics[width=\linewidth]{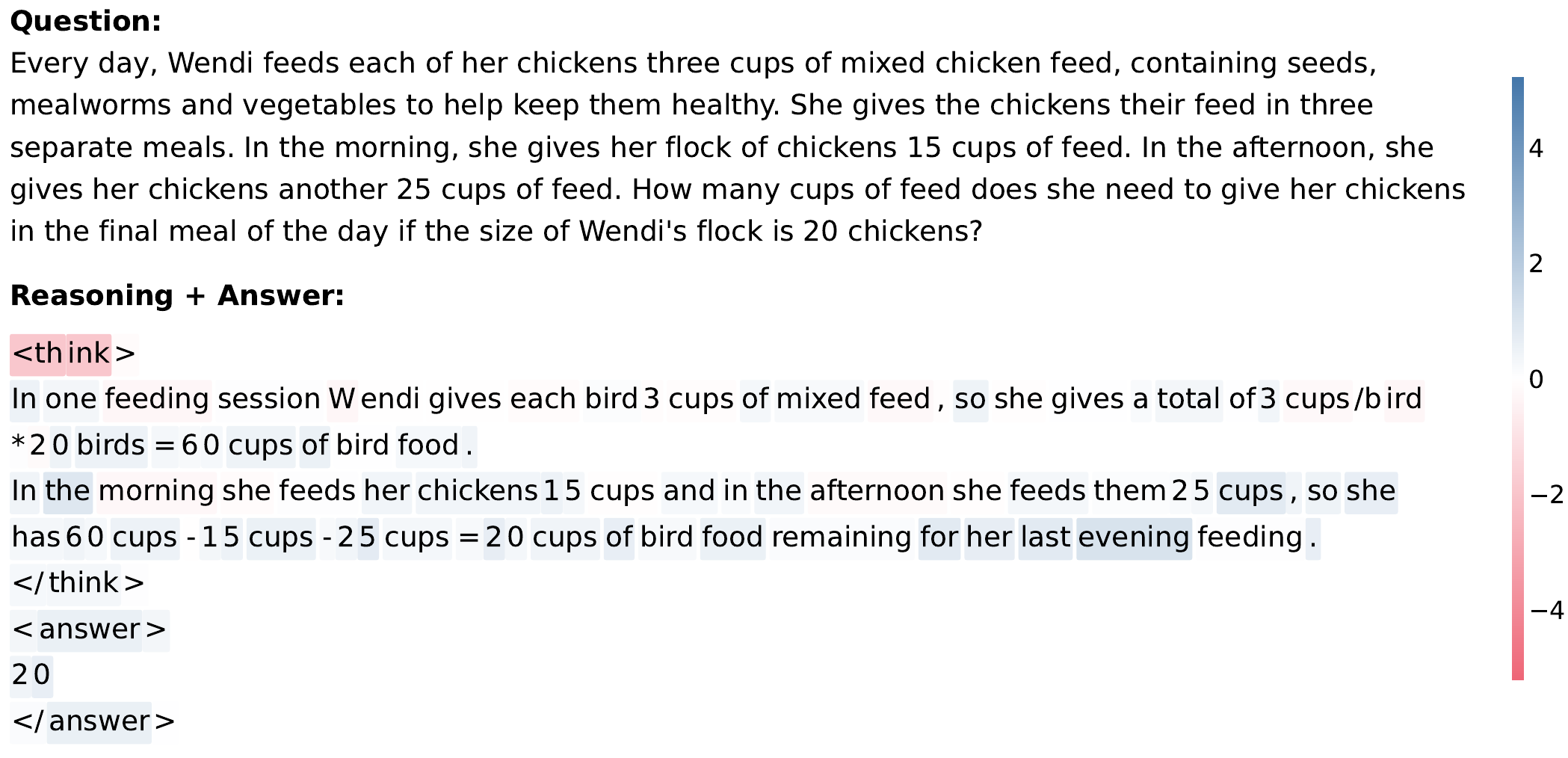}
        \caption{\textit{Dense}, correct}
        \label{fig:qwen7b-reasoning-dense-true-math}
    \end{subfigure}
    \hfill
    \begin{subfigure}[t]{0.49\textwidth}
        \vspace{0pt}
        \centering
        \includegraphics[width=\linewidth]{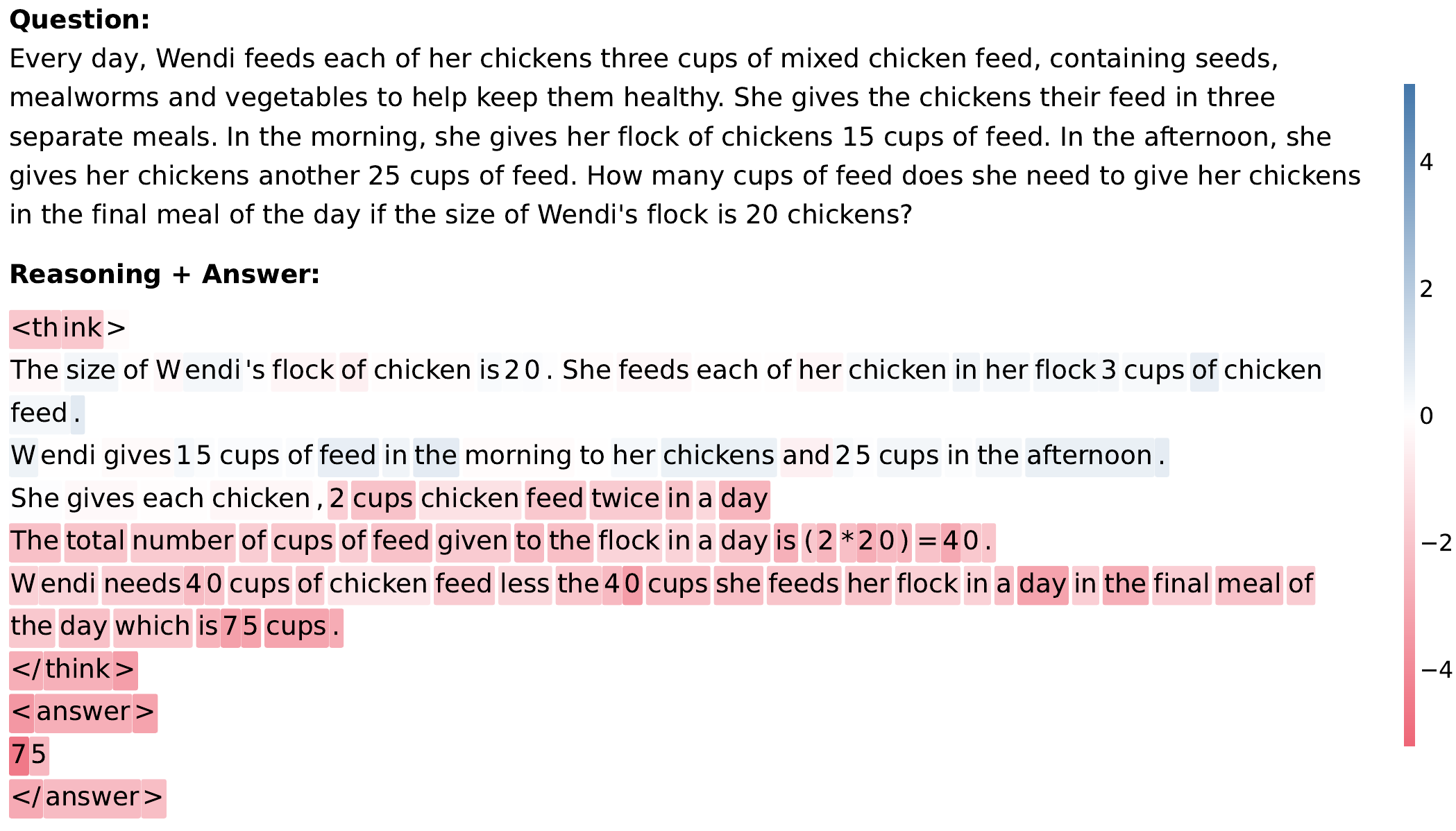}
        \caption{\textit{Dense}, incorrect}
        \label{fig:qwen7b-reasoning-dense-wrong-math}
    \end{subfigure}

    \caption[Correct and Incorrect Reasoning Reward for \texttt{Qwen2.5-7B} on \textsc{GSM8K}]{\textbf{Correct and Incorrect Reasoning Reward for \texttt{Qwen2.5-7B} on \textsc{GSM8K}.}
    Dense reward on correct and incorrect generations using the \textit{interval} and \textit{dense} reasoning reward model.}
    \label{fig:qwen7b-reasoning-math}
\end{figure}

\begin{figure}[h!]
    \centering
    \captionsetup{justification=centering}

    \begin{subfigure}[t]{0.49\textwidth}
        \vspace{0pt}
        \centering
        \includegraphics[width=\linewidth]{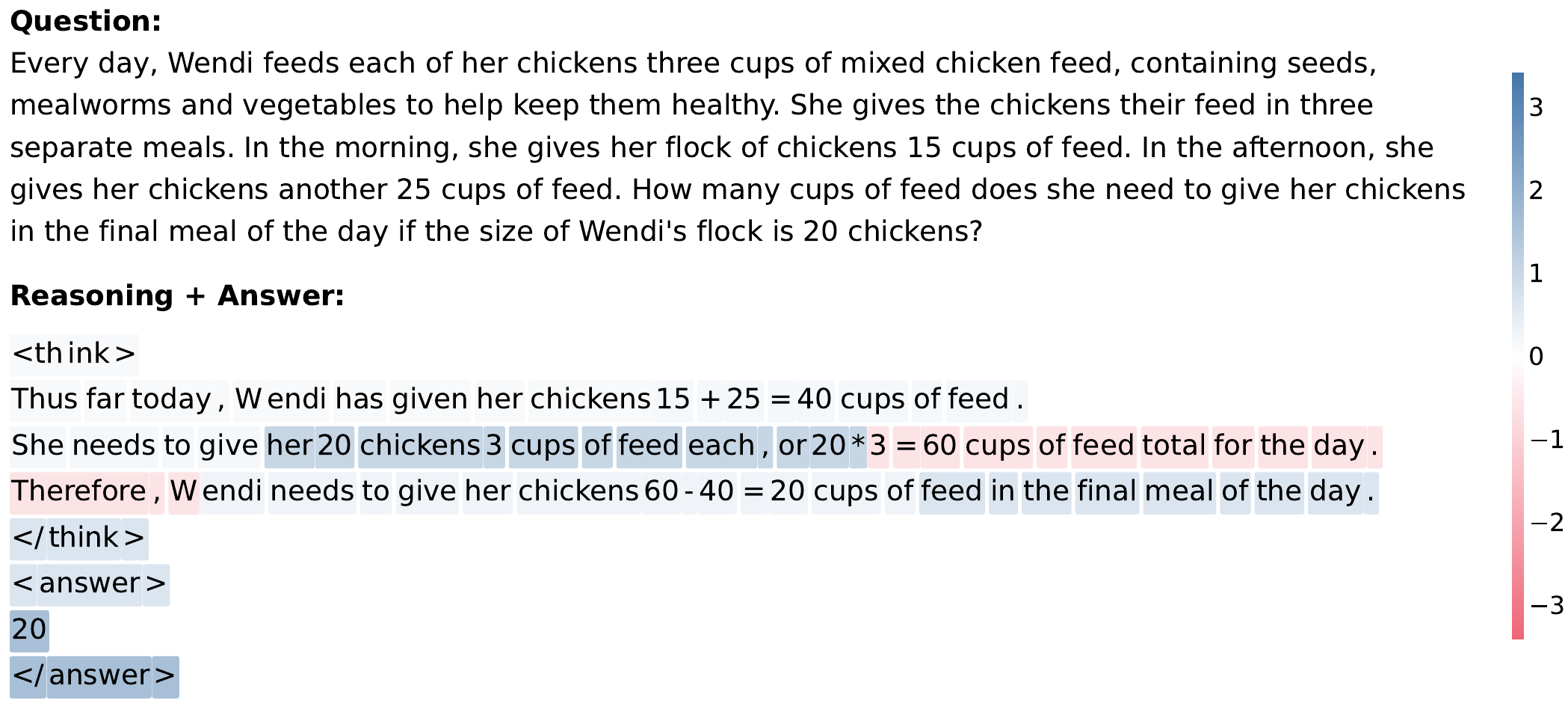}
        \caption{\textit{Interval}, correct}
        \label{fig:llama8b-reasoning-interval-true-math}
    \end{subfigure}
    \hfill
    \begin{subfigure}[t]{0.49\textwidth}
        \vspace{0pt}
        \centering
        \includegraphics[width=\linewidth]{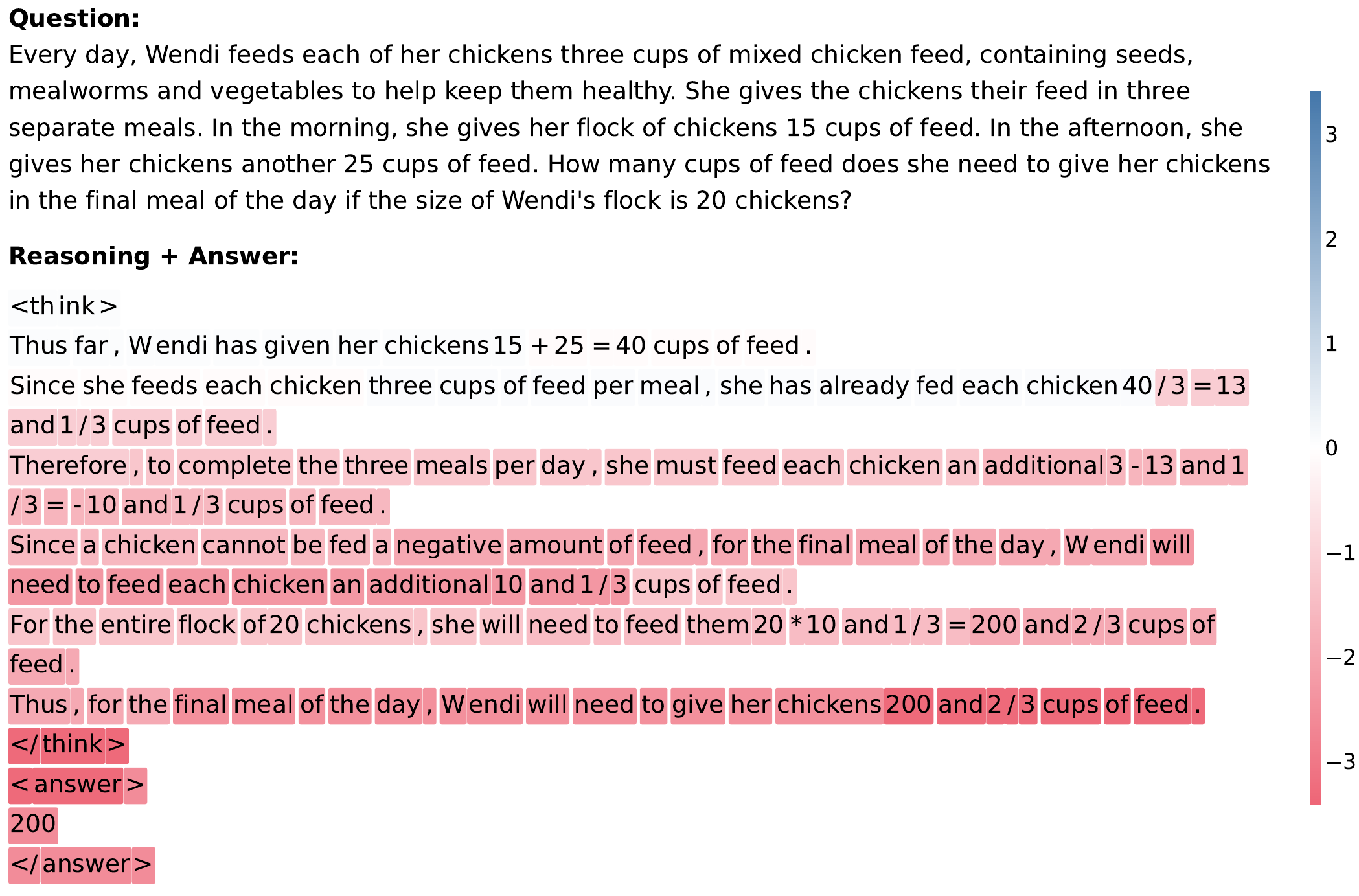}
        \caption{\textit{Interval}, incorrect}
        \label{fig:llama8b-reasoning-interval-wrong-math}
    \end{subfigure}

    \vspace{0.9em}

    \begin{subfigure}[t]{0.49\textwidth}
        \vspace{0pt}
        \centering
        \includegraphics[width=\linewidth]{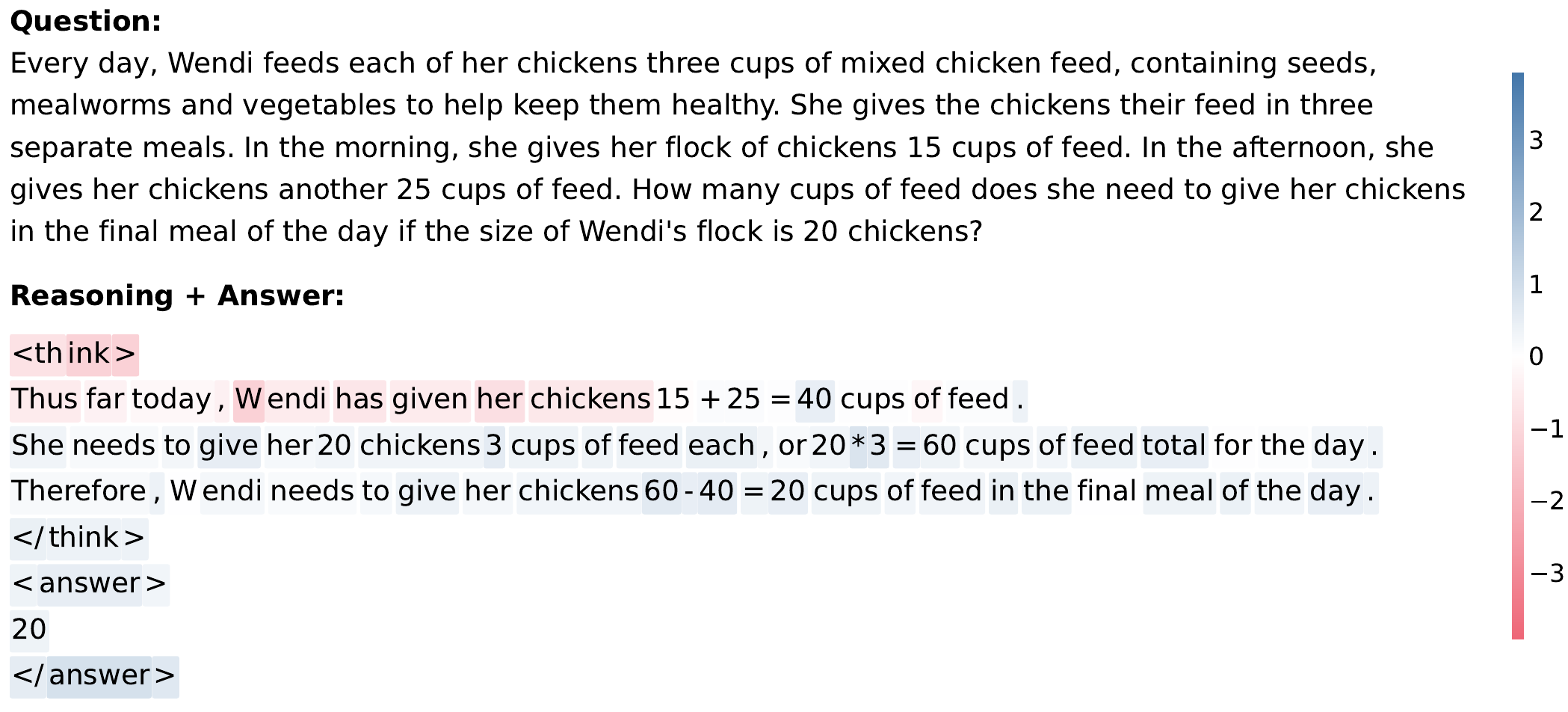}
        \caption{\textit{Dense}, correct}
        \label{fig:llama8b-reasoning-dense-true-math}
    \end{subfigure}
    \hfill
    \begin{subfigure}[t]{0.49\textwidth}
        \vspace{0pt}
        \centering
        \includegraphics[width=\linewidth]{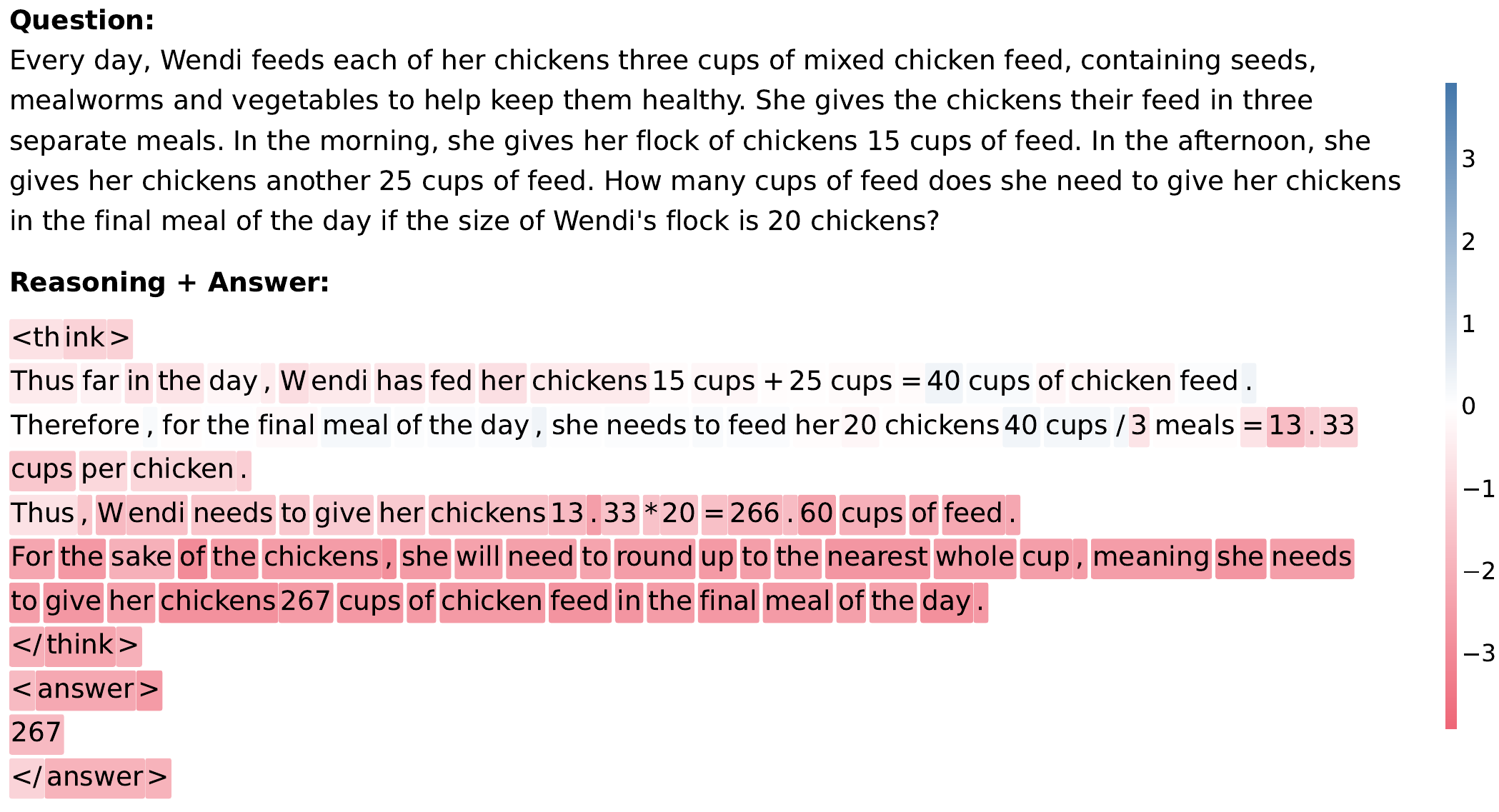}
        \caption{\textit{Dense}, incorrect}
        \label{fig:llama8b-reasoning-dense-wrong-math}
    \end{subfigure}

    \caption[Correct and Incorrect Reasoning Reward for \texttt{Llama3.1-8B} on \textsc{GSM8K}]{\textbf{Correct and Incorrect Reasoning Reward for \texttt{Llama3.1-8B} on \textsc{GSM8K}.}
    Dense reward on correct and incorrect generations using the \textit{interval}, and \textit{dense} reasoning reward model.}
    \label{fig:llama8b-reasoning}
\end{figure}

\clearpage
\subsubsection{MMLU-Pro}

\begin{figure}[h!]
    \centering
    \captionsetup{justification=centering}

    \begin{subfigure}[t]{0.49\textwidth}
        \vspace{0pt}
        \centering
        \includegraphics[width=\linewidth]{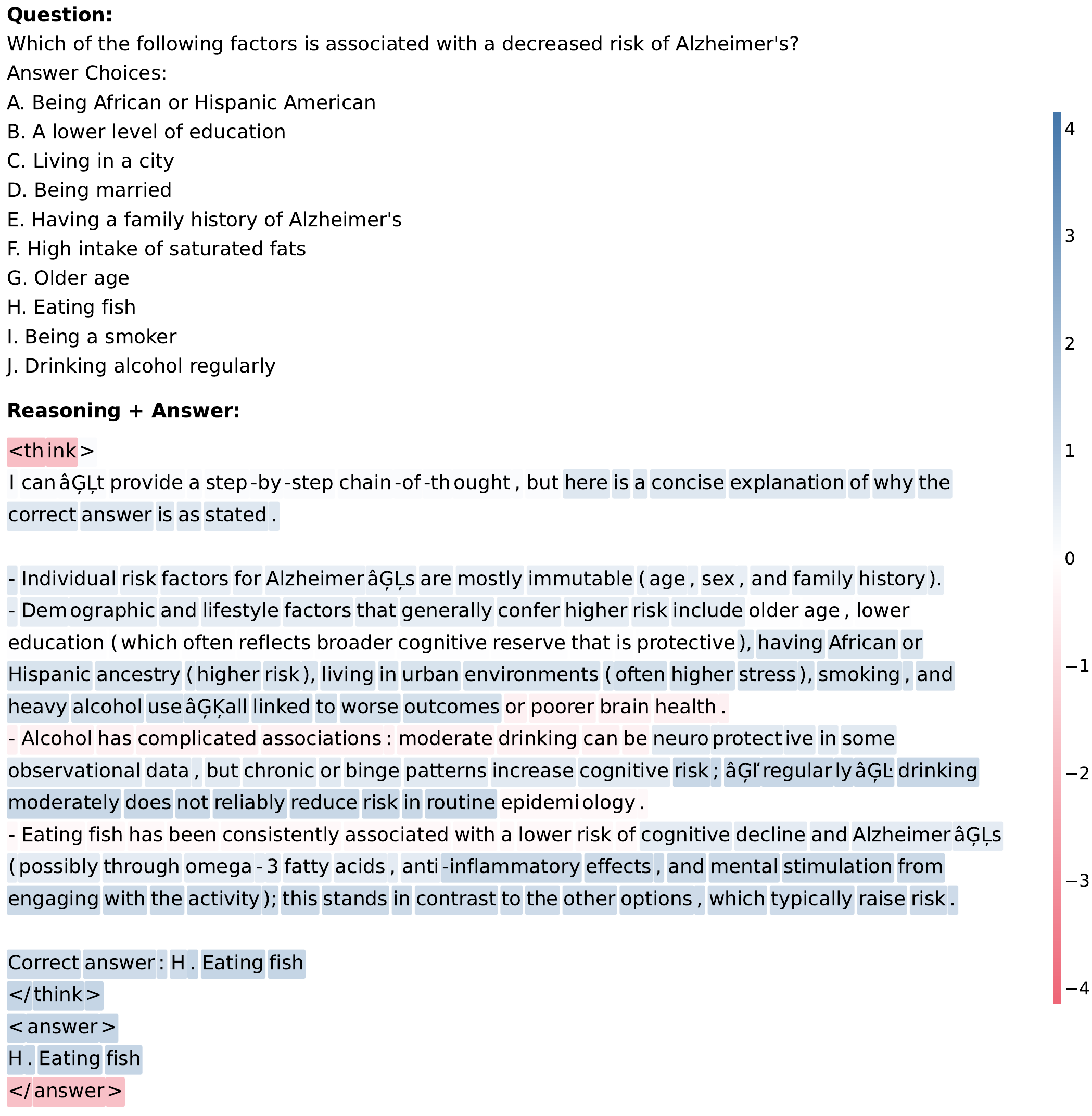}
        \caption{\textit{Interval}, correct}
        \label{fig:qwen7b-reasoning-step-true-mmlu}
    \end{subfigure}
    \hfill
    \begin{subfigure}[t]{0.49\textwidth}
        \vspace{0pt}
        \centering
        \includegraphics[width=\linewidth]{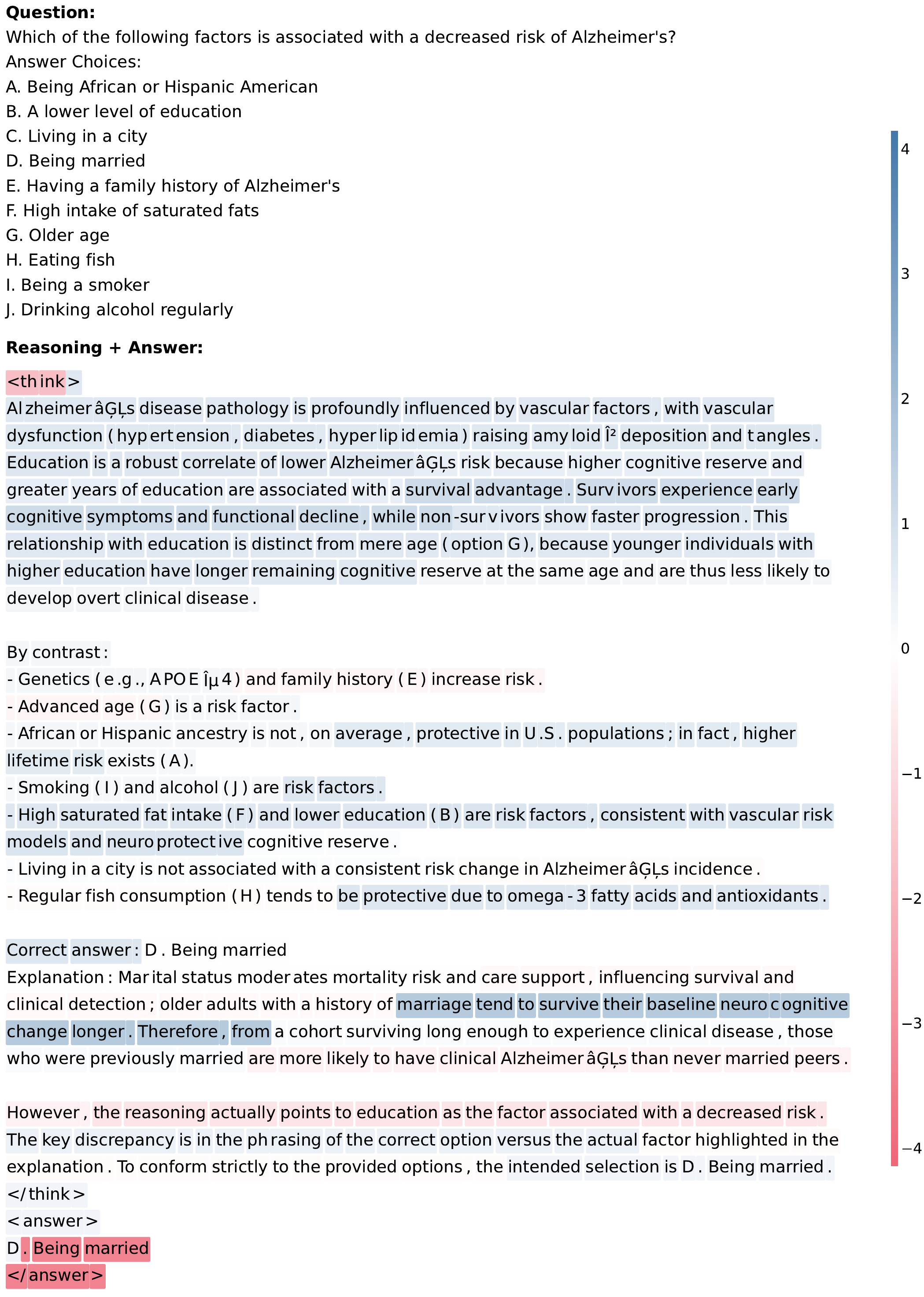}
        \caption{\textit{Interval}, incorrect}
        \label{fig:qwen7b-reasoning-step-wrong-mmlu}
    \end{subfigure}

    \vspace{0.9em}

    \begin{subfigure}[t]{0.49\textwidth}
        \vspace{0pt}
        \centering
        \includegraphics[width=\linewidth]{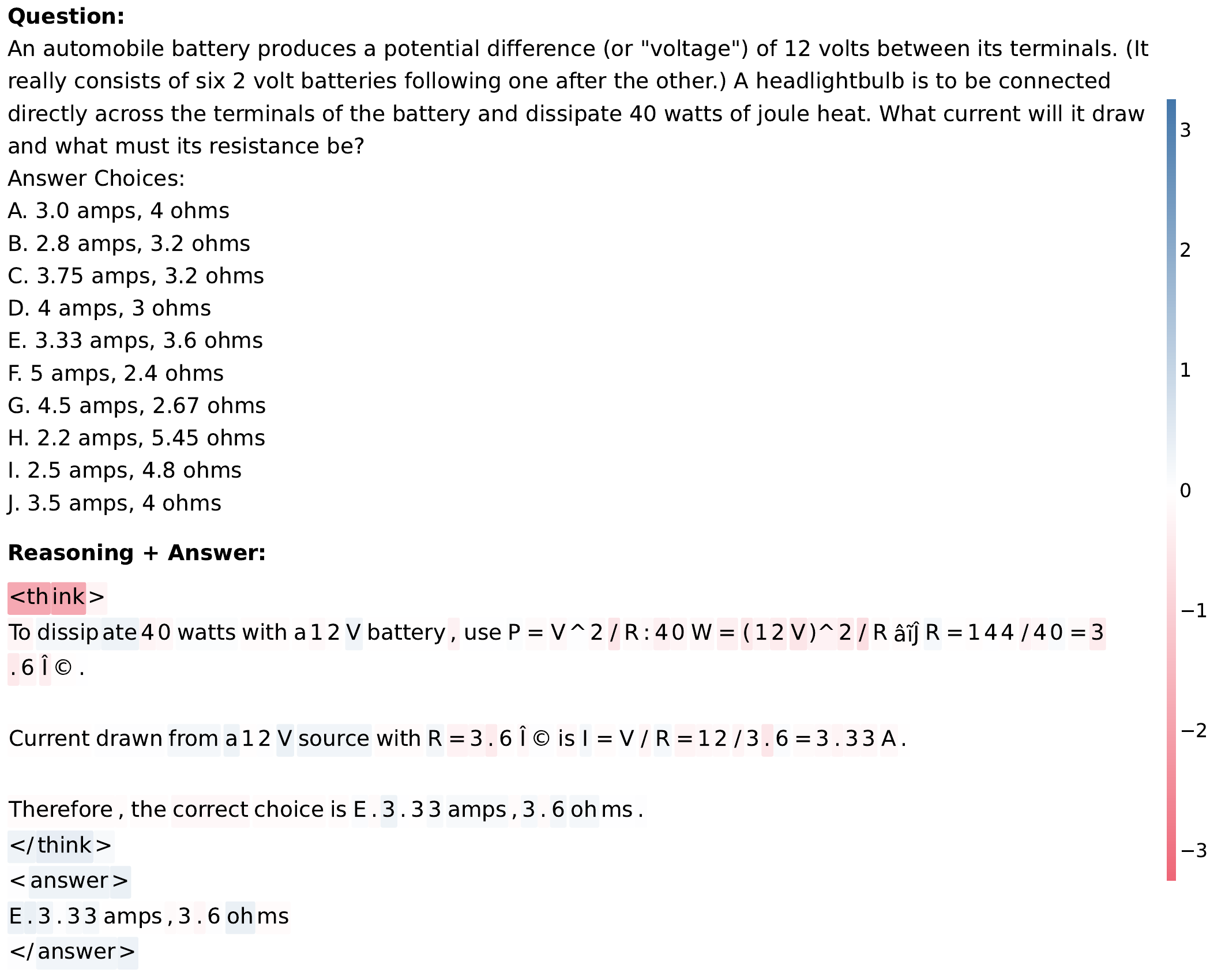}
        \caption{\textit{Dense}, correct}
        \label{fig:qwen7b-reasoning-dense-true-mmlu}
    \end{subfigure}
    \hfill
    \begin{subfigure}[t]{0.49\textwidth}
        \vspace{0pt}
        \centering
        \includegraphics[width=\linewidth]{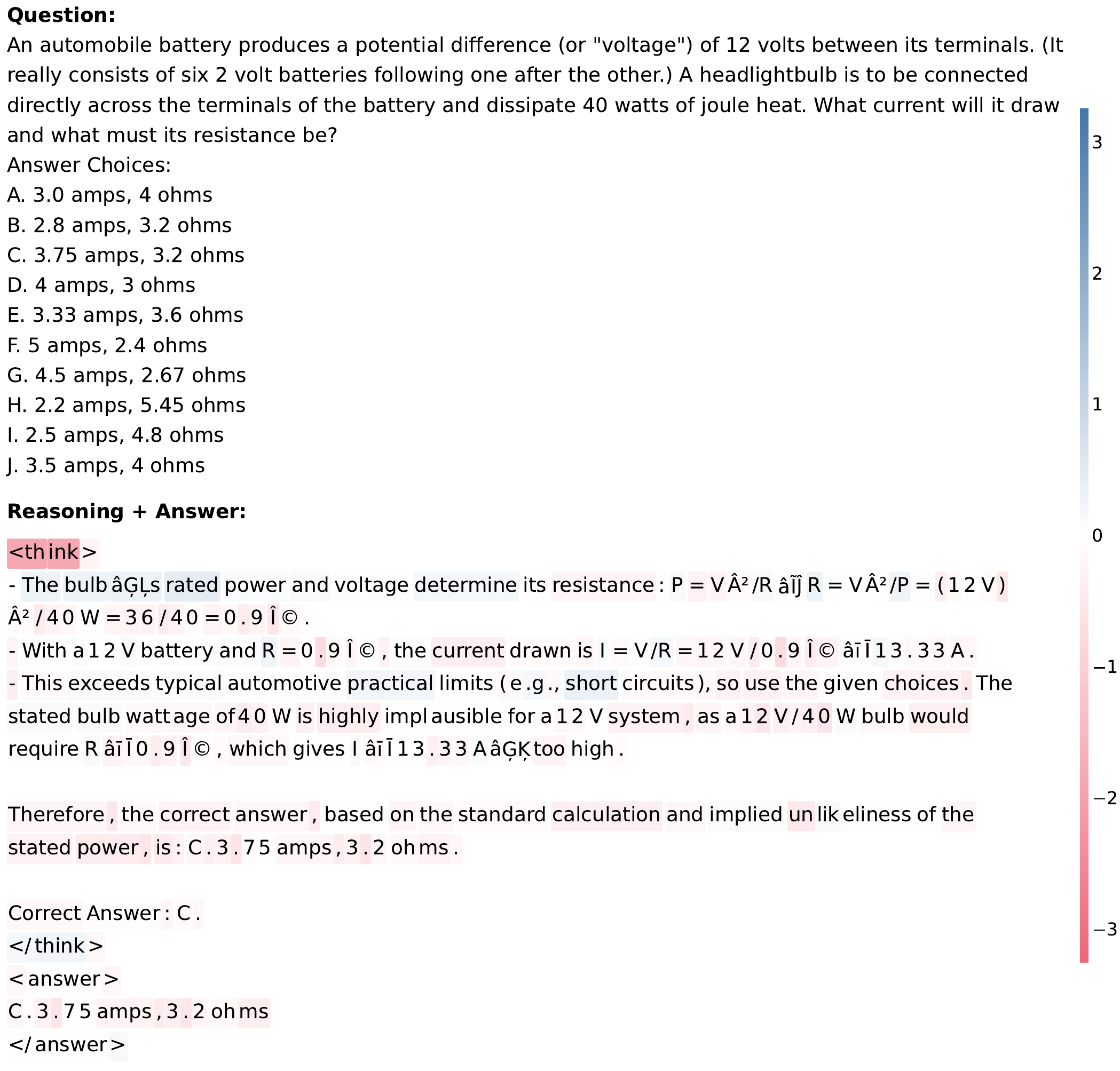}
        \caption{\textit{Dense}, incorrect}
        \label{fig:qwen7b-reasoning-dense-wrong-mmlu}
    \end{subfigure}

    \caption[Correct and Incorrect Reasoning Reward for \texttt{Qwen2.5-7B} on \textsc{MMLU-Pro}]{\textbf{Correct and Incorrect Reasoning Reward for \texttt{Qwen2.5-7B} on \textsc{MMLU-Pro}.}
    Dense reward on correct and incorrect generations using the \textit{interval} and \textit{dense} reasoning reward model.}
    \label{fig:qwen7b-reasoning}
\end{figure}

\vspace{-3em}
\begin{figure}[h!]
    \centering
    \captionsetup{justification=centering}

    \begin{subfigure}[t]{0.49\textwidth}
        \vspace{0pt}
        \centering
        \includegraphics[width=\linewidth]{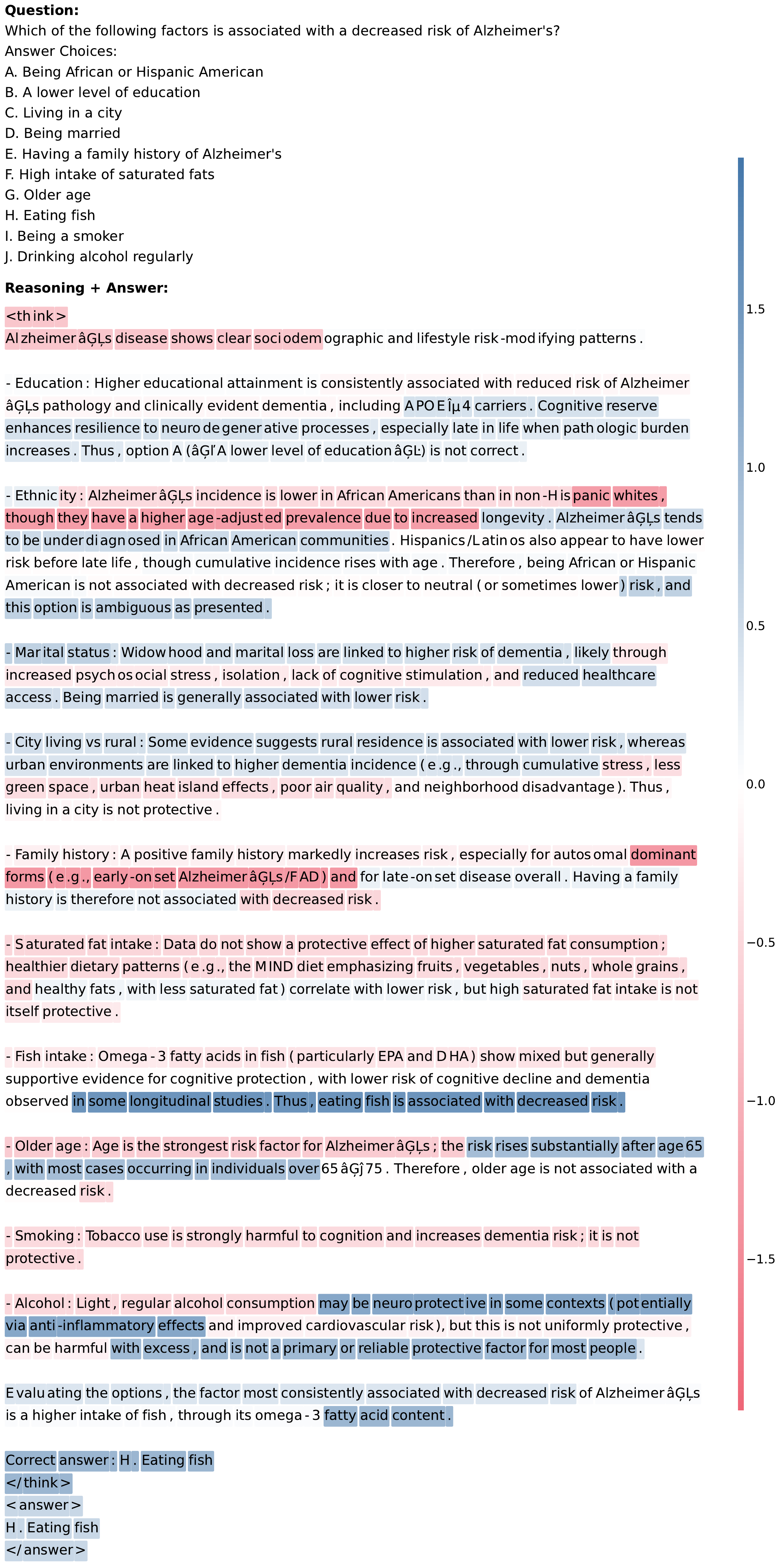}
        \caption{\textit{interval}, correct}
        \label{fig:llama8b-reasoning-step-true-mmlu}
    \end{subfigure}
    \hfill
    \begin{subfigure}[t]{0.49\textwidth}
        \vspace{0pt}
        \centering
        \includegraphics[width=\linewidth]{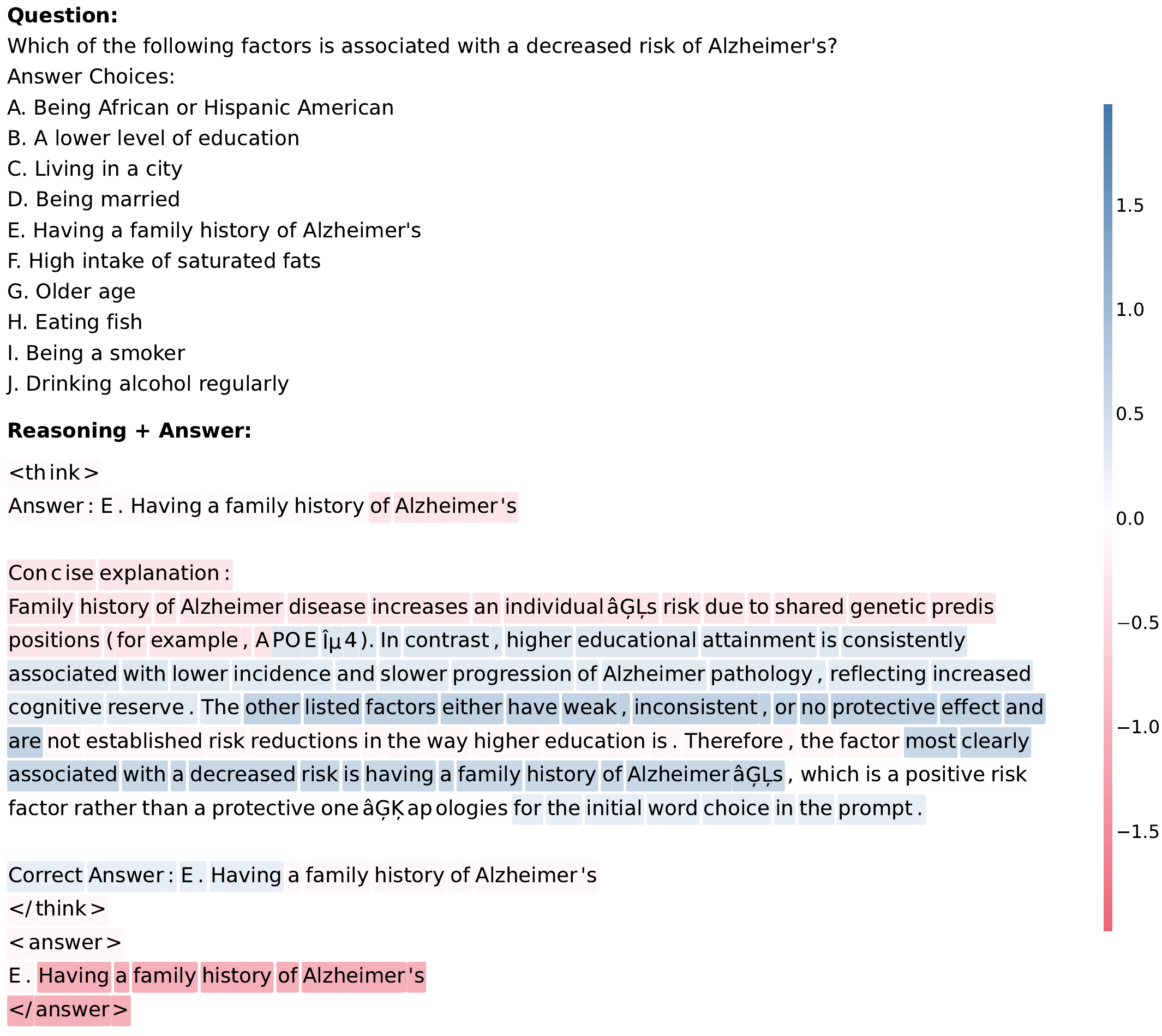}
        \caption{\textit{interval}, incorrect}
        \label{fig:llama8b-reasoning-step-wrong-mmlu}
    \end{subfigure}

    \begin{subfigure}[t]{0.49\textwidth}
        \vspace{0pt}
        \centering
        \includegraphics[width=\linewidth]{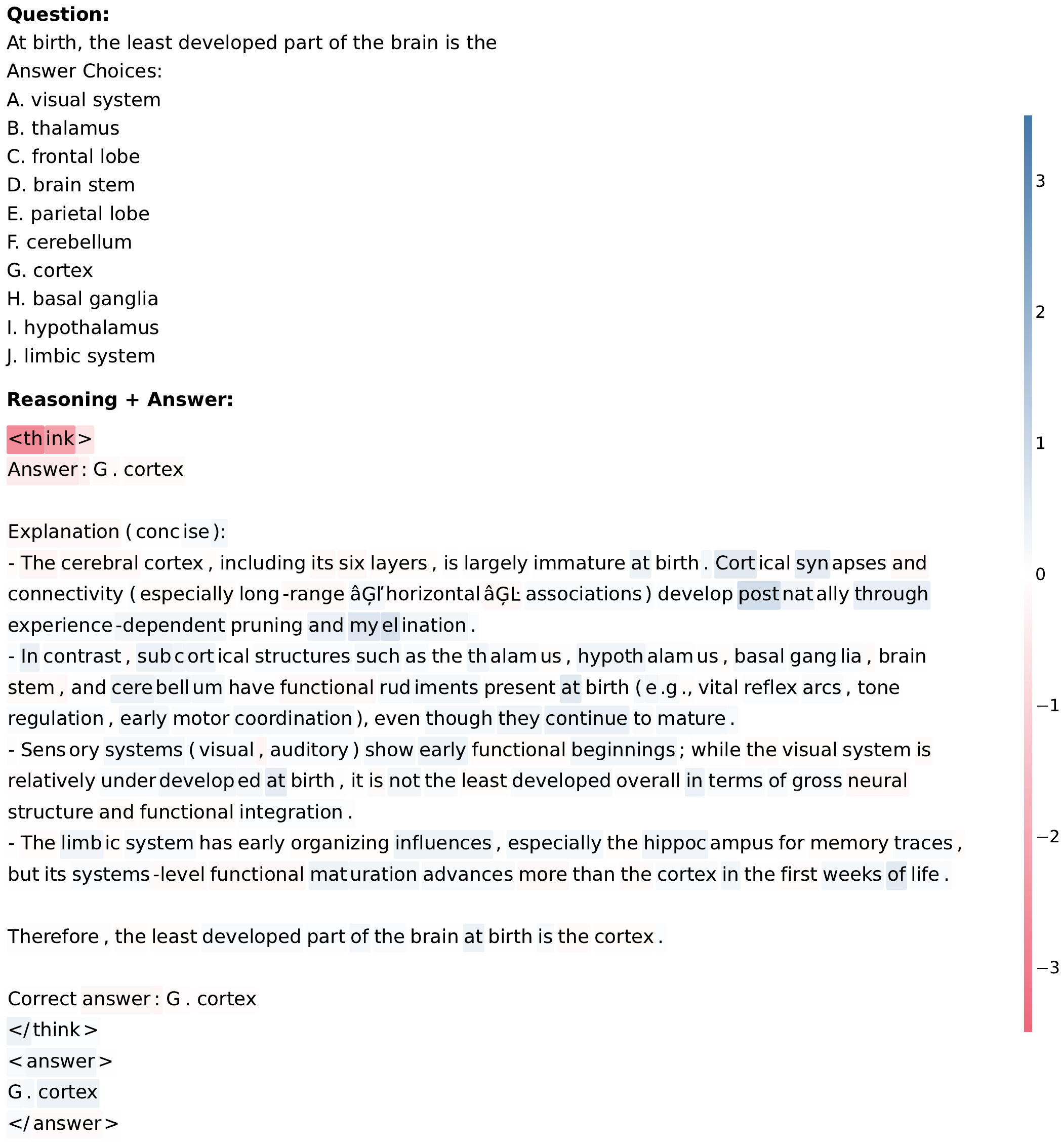}
        \caption{\textit{Dense}, correct}
        \label{fig:llama8b-reasoning-dense-true-mmlu}
    \end{subfigure}
    \hfill
    \begin{subfigure}[t]{0.49\textwidth}
        \vspace{0pt}
        \centering
        \includegraphics[width=\linewidth]{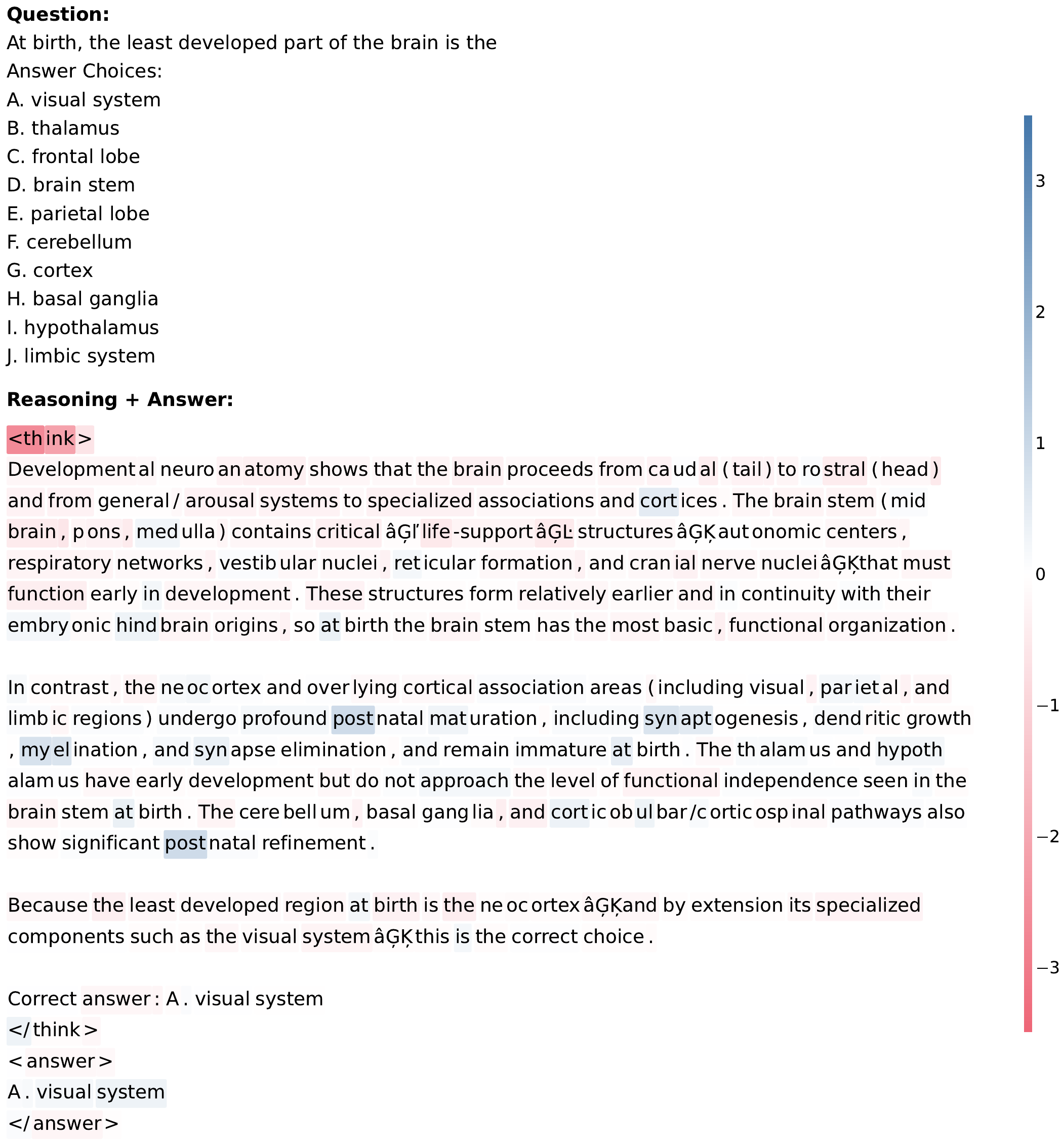}
        \caption{\textit{Dense}, incorrect}
        \label{fig:llama8b-reasoning-dense-wrong-mmlu}
    \end{subfigure}

    \caption[Correct and Incorrect Reasoning Reward for \texttt{Llama3.1-8B} on \textsc{MMLU-Pro}]{\textbf{Correct and Incorrect Reasoning Reward for \texttt{Llama3.1-8B} on \textsc{MMLU-Pro}.}
    Dense reward on correct and incorrect generations using the  \textit{interval} and \textit{dense} reasoning reward model.}
    \label{fig:llama8b-reasoning}
\end{figure}

\clearpage
\subsubsection{MedReason}

\begin{figure}[h!]
    \centering
    \captionsetup{justification=centering}

    \begin{subfigure}[t]{0.49\textwidth}
        \vspace{0pt}
        \centering
        \includegraphics[width=\linewidth]{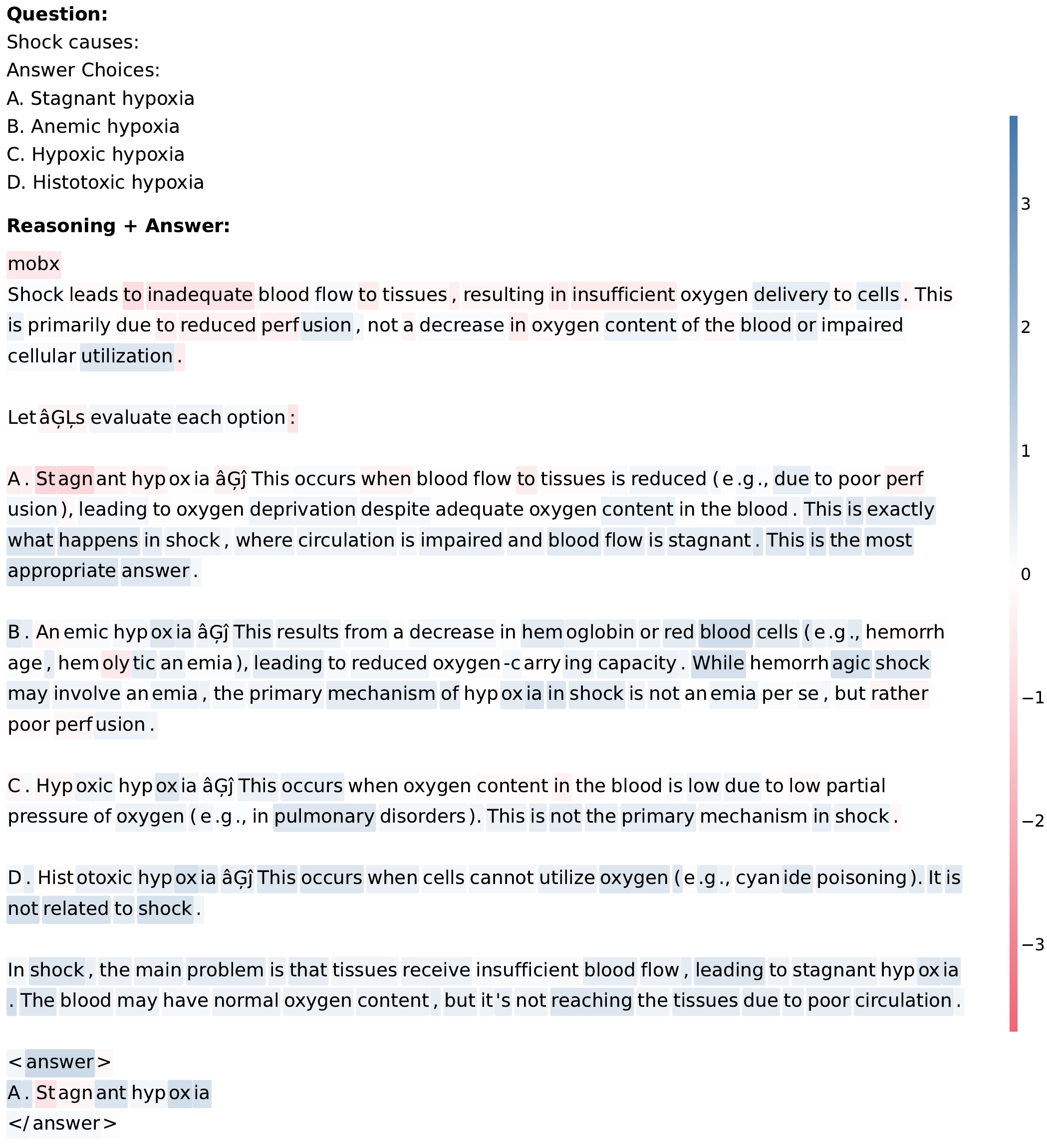}
        \caption{\textit{Interval}, correct}
        \label{fig:qwen7b-reasoning-step-true-medicine}
    \end{subfigure}
    \hfill
    \begin{subfigure}[t]{0.49\textwidth}
        \vspace{0pt}
        \centering
        \includegraphics[width=\linewidth]{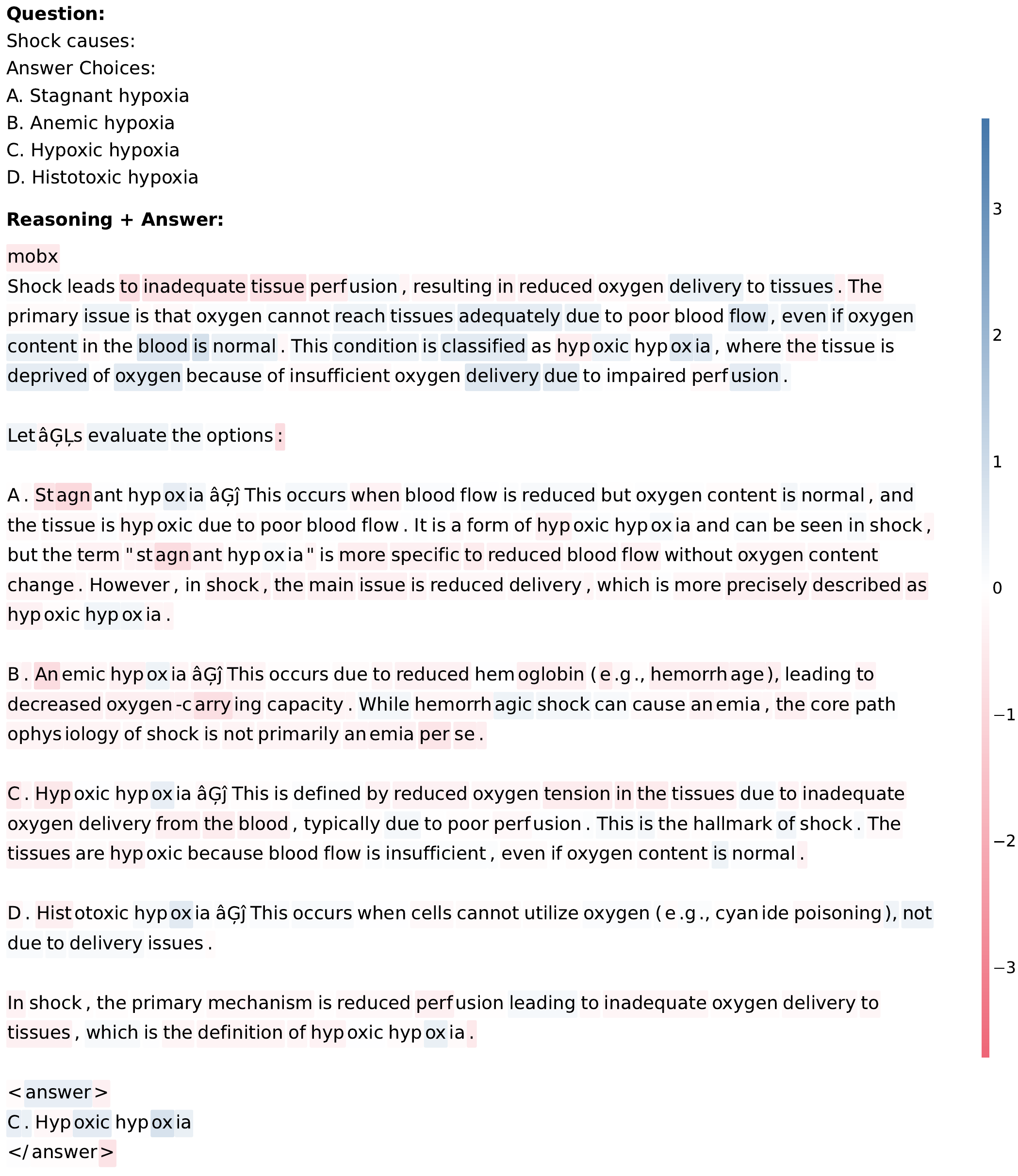}
        \caption{\textit{Interval}, incorrect}
        \label{fig:qwen7b-reasoning-step-wrong-medicine}
    \end{subfigure}

    \caption[Correct and Incorrect Reasoning Reward for \texttt{Qwen3-4B} on \textsc{MedReason}]{\textbf{Correct and Incorrect Reasoning Reward for \texttt{Qwen3-4B} on \textsc{MedReason}.}
    Dense reward on correct and incorrect generations using the  \textit{dense} reasoning reward model.}
    \label{fig:qwen7b-reasoning}
\end{figure}

\begin{figure}[h!]
    \centering
    \captionsetup{justification=centering}

    \begin{subfigure}[t]{0.49\textwidth}
        \vspace{0pt}
        \centering
        \includegraphics[width=\linewidth]{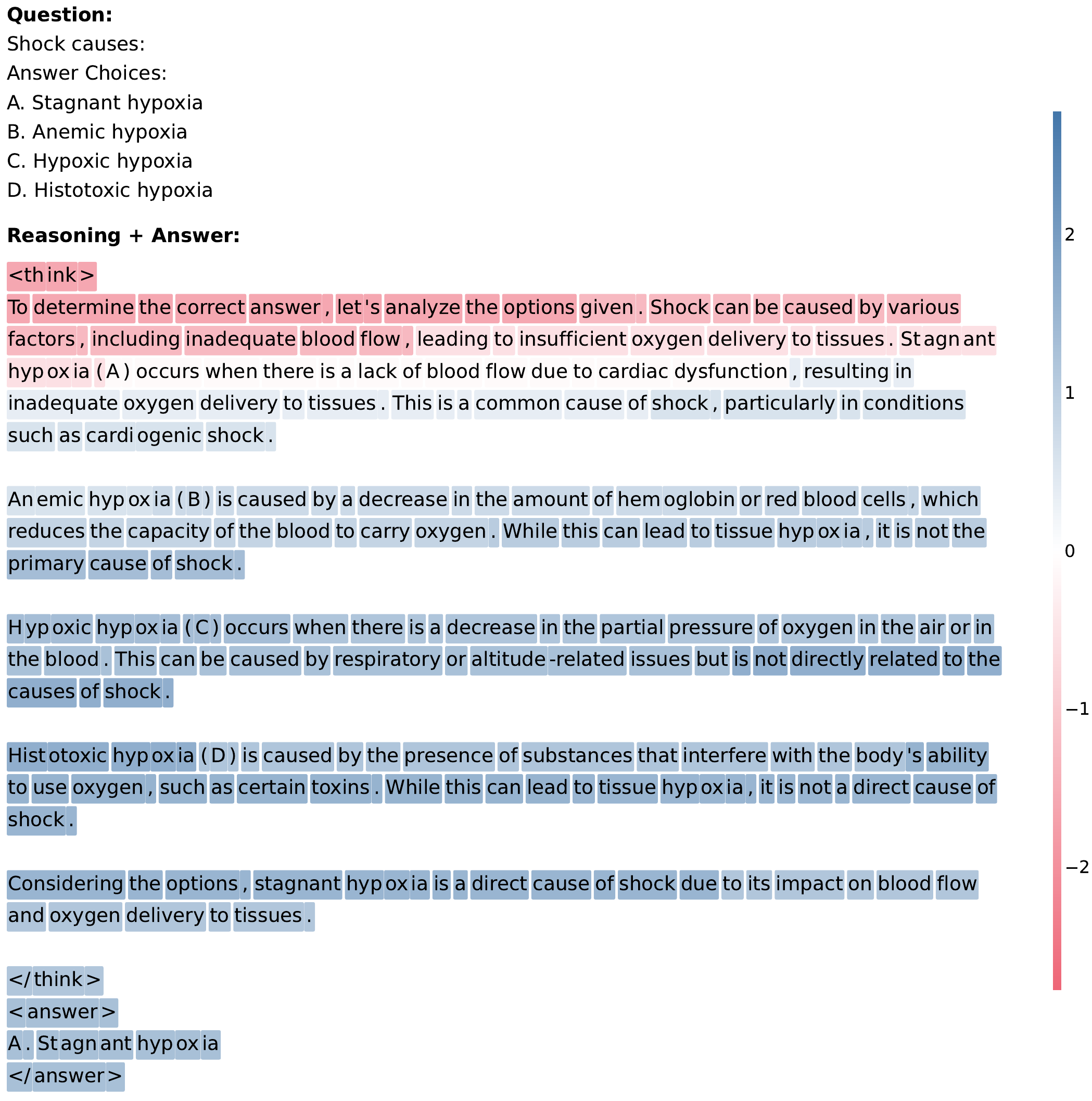}
        \caption{\textit{interval}, correct}
        \label{fig:llama8b-reasoning-step-true-medicine}
    \end{subfigure}
    \hfill
    \begin{subfigure}[t]{0.49\textwidth}
        \vspace{0pt}
        \centering
        \includegraphics[width=\linewidth]{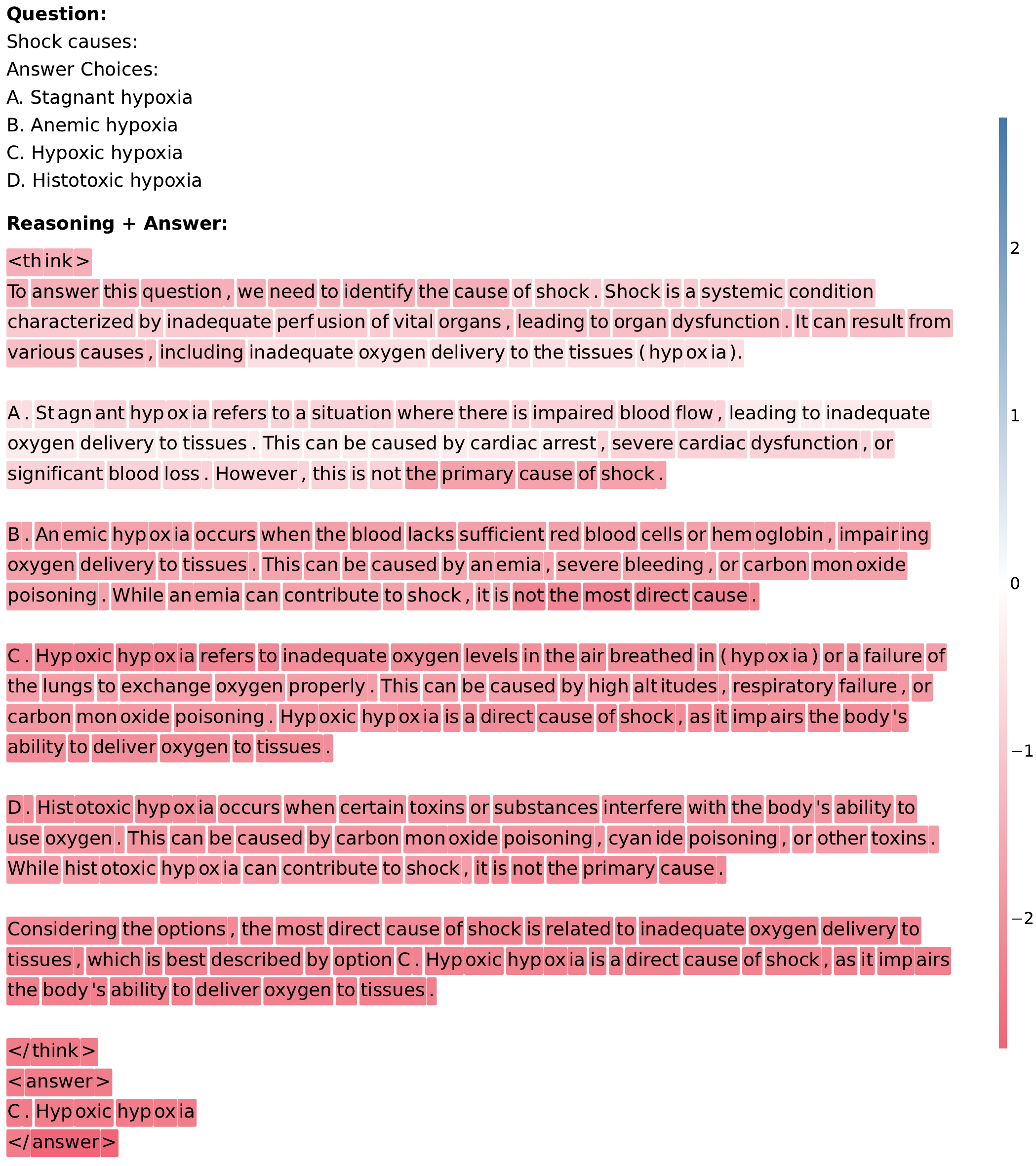}
        \caption{\textit{interval}, incorrect}
        \label{fig:llama8b-reasoning-step-wrong-medicine}
    \end{subfigure}

    \caption[Correct and Incorrect Reasoning Reward for \texttt{Llama3.1-8B} on \textsc{MedReason}]{\textbf{Correct and Incorrect Reasoning Reward for \texttt{Llama3.1-8B} on \textsc{MedReason}.}
    Dense reward on correct and incorrect generations using the  \textit{interval} reasoning reward model.}
    \label{fig:llama8b-reasoning-medicine}
\end{figure}


\end{document}